\documentclass{article}
\usepackage[preprint]{neurips_2024}
\usepackage[utf8]{inputenc} % allow utf-8 input
\usepackage[T1]{fontenc}    % use 8-bit T1 fonts
\usepackage{hyperref}       % hyperlinks
\usepackage{url}            % simple URL typesetting
\usepackage{booktabs}       % professional-quality tables
\usepackage{amsfonts}       % blackboard math symbols
\usepackage{nicefrac}       % compact symbols for 1/2, etc.
\usepackage{microtype}      % microtypography
\usepackage{kotex}
\usepackage{amsmath}
\usepackage{tabularx}
\usepackage[table]{xcolor}
\usepackage{tcolorbox}
\usepackage{adjustbox} % 표가 길 때 크기 조정용
\usepackage{multirow} 
\usepackage{subcaption}
\usepackage{graphicx}
\usepackage{float}
\usepackage{algorithm}
\usepackage{algpseudocode}
\usepackage{tablefootnote}
\usepackage{arydshln}
\usepackage{amssymb}% http://ctan.org/pkg/amssymb
\usepackage{pifont}% http://ctan.org/pkg/pifont
\newcommand{\cmark}{\ding{51}}%
\newcommand{\xmark}{\ding{55}}%

\newcommand{\dashedmidrule}{\hdashline[1pt/2pt]}

\newcommand{\kormo}{KORMo}

\title{KORMo: Korean Open Reasoning Model for Everyone}

\author{
  Minjun Kim$^{1,4}$\thanks{marks core contributors.} \quad 
  Hyeonseok Lim$^{1,4*}$ \quad 
  Hangyeol Yoo$^{1,4*}$ \quad 
  Inho Won$^{1*}$ \\
  \textbf{Seungwoo Song}$^{1,4}$ \quad 
  \textbf{Minkyung Cho}$^{1}$  \quad 
  \textbf{Junghun Yuk}$^{1}$ \quad 
  \textbf{Changsu Choi}$^{1,4}$  \\
  \textbf{Dongjae Shin}$^{1,4}$  \quad 
  \textbf{Huije Lee}$^{2}$ \quad 
  \textbf{Hoyun Song}$^{1}$ \\
  \textbf{Alice Oh}$^{3}$ \quad 
  \textbf{KyungTae Lim}$^{1}$ \\
  \\
  $^{1}$KAIST MLP Lab \quad 
  $^{2}$KAIST NLPCL Lab \quad 
  $^{3}$KAIST U\&I Lab \quad 
  $^{4}$SeoulTech \\
  \texttt{Corresponding Author: ktlim@kaist.ac.kr}
}

\begin{document}

\maketitle

\begin{abstract}
This work presents the first large-scale investigation into constructing a \emph{fully open} bilingual large language model (LLM) for a non-English language, specifically Korean, trained predominantly on synthetic data. We introduce \textbf{KORMo-10B}, a 10.8B-parameter model trained from scratch on a Korean–English corpus in which 68.74\% of the Korean portion is synthetic. Through systematic experimentation, we demonstrate that synthetic data, when carefully curated with balanced linguistic coverage and diverse instruction styles, does not cause instability or degradation during large-scale pretraining. Furthermore, the model achieves performance comparable to that of contemporary open-weight multilingual baselines across a wide range of reasoning, knowledge, and instruction-following benchmarks. Our experiments reveal two key findings: (1) synthetic data can reliably sustain long-horizon pretraining without model collapse, and (2) bilingual instruction tuning enables near-native reasoning and discourse coherence in Korean. By fully releasing all components including data, code, training recipes, and logs, this work establishes a transparent framework for developing \emph{synthetic data–driven fully open models (FOMs)} in low-resource settings and sets a reproducible precedent for future multilingual LLM research. All model checkpoints, datasets, and source codes are publicly available at 
\href{https://huggingface.co/kormo-lm}{\texttt{huggingface.co/kormo-lm}}.
\end{abstract}

% (총괄: 호윤, 민준서폿)
\section{Introduction}

% 오픈소스 대형 언어모델(LLM)은 최근 상용 모델에 근접한 성능과 활용성을 보이며 학술과 산업 전반에서 빠르게 확산되고 있다 \citep{llama3,qwen25,deepseekr1}. 그러나 이들 중 상당수는 \emph{open-weight model (OWM)}에 해당하여 최종 파라미터만 공개하고, 데이터·전처리·코드·하이퍼파라미터·학습 로그 등 핵심 레시피는 비공개로 남는 경우가 많다. 이러한 공개 범위의 제약은 재현성, 공정한 비교, 책임 있는 배포에 필요한 \emph{증거사슬(chain of custody)}을 약화시키며, 결과적으로 후속 연구의 확장성을 저해한다.

Open-source large language models (LLMs) have recently demonstrated performance and utility approaching that of proprietary models, which has led to their rapid adoption in both academia and industry~\citep{llama3,qwen25,deepseekr1}. However, a significant number of these are \emph{open-weight model (OWM)}, where only the final parameters are released, while critical components of the training recipe, such as data, preprocessing methods, code, hyperparameters, and training logs, often remain undisclosed. This limited scope of disclosure weakens the \emph{chain of custody} necessary for reproducibility, fair comparison, and responsible deployment, ultimately hindering the scalability of subsequent research.

% 이에 대한 대안으로 학습 전 과정을 투명하게 공개하는 \emph{fully open model (FOM)} 시도가 부상하고 있다. \citet{olmo2,olmoe}는 데이터 소스와 정제 규칙, 학습 스크립트, 하이퍼파라미터, 로그, 체크포인트를 포함한 전 과정을 공개하여 FOM의 실효성을 입증했다. 다국어 공개 시도로는 \citet{bloom}이 대표적이며, 경량·장문·추론 강화 설정을 전과정 공개로 탐색한 시도 또한 이어지고 있다 \citep{smollm3}. 이러한 흐름은 재현 가능 연구와 파생 모델 생태계를 촉진한다는 점에서 학술적·사회적 파급력이 크다.

In response, \emph{fully open models (FOMs)}, which transparently release the entire training pipeline, have emerged as a compelling alternative. The efficacy of the FOM approach was shown by~\citet{olmo2,olmoe}, who revealed their full methodology, including data sources, cleaning protocols, training scripts, hyperparameters, logs, and model checkpoints. While \citet{bloom} stands as a prominent multilingual example, the trend towards full disclosure has also extended to explorations of compact, long-context, and reasoning-focused models~\citep{smollm3}. This trend has significant academic and societal impacts by enabling reproducible research and fostering an ecosystem for derivative models.

%그럼에도 현재의 FOM 성과는 영어 중심에 치우쳐 있다. 비영어권에서는 (i) 대규모 크롤 코퍼스의 부재 및 저작권/라이선스 복잡성, (ii) 품질 정제·중복 제거·오염(contamination) 통제의 높은 비용, (iii) 토크나이저와 언어 혼합비 설계의 불확실성 등으로 FOM 구축 난도가 높다. 반면 일단 최초 FOM이 공개되면 해당 언어권의 연구 장벽이 급감하고 생태계가 빠르게 활성화되는 경향이 보고되어 왔다 \citep{bloom,olmo2}. 따라서 비영어권 \emph{최초} FOM을 실증하는 작업은 학술적·실무적 의의가 크다.

Despite these advancements, FOM development remains predominantly focused on English. In non-English settings, creating an FOM is made more difficult by several challenges: (i) a lack of large-scale web-crawled corpora, further complicated by copyright and licensing issues; (ii) the significant cost of data curation, including quality refinement, deduplication, and contamination control; and (iii) the ambiguities related to tokenizer architecture and language composition design. Nevertheless, it has been observed that releasing a foundational FOM for a specific language dramatically lowers research barriers and accelerates ecosystem development~\citep{bloom,olmo2}. Therefore, demonstrating the successful implementation of a \emph{first} non-English FOM is a task of profound academic and practical significance.

% 최근에는 이러한 격차를 해소하기 위한 수단으로 합성/증강 데이터(augmentation)가 주목받고 있다. \citet{phi1,phi15}는 \emph{textbook-style} 합성 데이터가 소형/중형 모델 효율 향상에 기여함을 보였고, 산업 배포 사례에서도 대규모 합성 파이프라인이 사용되고 있다 \citep{nemotron4tech}. 동시에 장기·대규모 사전학습에 적합하도록 정제된 웹 말뭉치가 공개되며 데이터 생태계가 빠르게 개선되는 추세다 \citep{dolma,nemotroncc,ultrafineweb_2025,fineweb_2024}. 합성 데이터는 사전학습뿐 아니라 SFT와 RLHF 단계에서도 활발히 쓰이고 있으며 \citep{selfinstruct,instructgpt,dpo,orpo,kto,grpo}, 프롬프트 설계와 데이터 커리큘럼이 성능 안정성에 중요한 역할을 한다. 그러나 생성 데이터의 자기섭식(self-consuming)으로 인한 성능 붕괴 위험도 거듭 지적되어 왔기에 \citep{modelcollapse,syntheticrisk}, 비영어권에서 합성 데이터를 \emph{주력 자원}으로 삼으려면 안정성과 편향에 대한 정량적 검증이 필수적이다.

More recently, the use of synthetic and augmented data has emerged as a prominent method for addressing this disparity. Research by \citet{phi1,phi15} has demonstrated that textbook-style synthetic data contributes to enhanced efficiency in small- to medium-scale models. At the same time, large-scale synthesis pipelines are also being deployed in industrial applications \citep{nemotron4tech}. Simultaneously, the data ecosystem is undergoing rapid enhancement through the public release of refined web corpora suitable for extensive, long-term pre-training \citep{dolma,nemotroncc,ultrafineweb_2025,fineweb_2024}. The application of synthetic data is prevalent not only in pre-training but also during the SFT and RLHF phases \citep{selfinstruct,instructgpt,dpo,orpo,kto,grpo}, with prompt engineering and data curricula being critical factors for model stability. However, the risk of model collapse due to the self-consuming nature of synthetic data has been a persistent concern \citep{modelcollapse,syntheticrisk}. Consequently, before synthetic data can be leveraged as a \emph{primary resource} in non-English domains, rigorous quantitative validation of its stability and potential biases is imperative.

% 토크나이저와 언어 혼합비 설계 또한 비영어권 FOM의 성패를 좌우하는 핵심 요인이다. 서브워드 체계(BPE/Unigram/byte-level), 어휘 크기, 언어별 비율은 압축 효율과 학습 비용, 다운스트림 일반화에 직접적인 영향을 준다 \citep{bpe,sentencepiece,radford2019gpt2,subwordcurse}. 특히 비라틴 문자권·형태 풍부 언어에서는 표층 단위 선택과 어휘 경계가 다르게 작동할 수 있으므로, FOM 맥락에서는 초기 설계 단계에서 이를 명시적으로 탐색해야 한다.

The design of the tokenizer and the language mixture ratio are also critical factors that determine the success or failure of a non-English FOM. The subword tokenization scheme (e.g., BPE, Unigram, byte-level), vocabulary size, and the proportion of each language directly impact compression efficiency, training cost, and downstream generalization \citep{bpe,sentencepiece,radford2019gpt2,subwordcurse}. This consideration is especially critical for non-Latin script and morphologically rich languages, where surface-form unit selection and vocabulary boundaries can operate differently. Therefore, in the context of an FOM, these aspects must be explicitly explored during the initial design phase.

%본 논문은 한국어를 대상으로 \emph{합성 데이터 주도} FOM 구축의 타당성과 한계를 체계적으로 검증한다. 우리는 한국어–영어 이중언어 모델을 \emph{from scratch}로 학습하고, 전체 한국어 데이터의 76.4\%를 합성으로 구성했으며, 사전학습–SFT–강화학습(RL) 전주기에 걸쳐 총 210B 토큰 규모의 커리큘럼을 설계하였다. 합성 데이터는 Qwen과 공개 OSS 계열 모델(GPT-OSS 등)을 조합해 다양한 스타일·주제 분포를 확보하였다. 이러한 설정을 기반으로 다음의 연구질문을 다룬다:
%\begin{enumerate}
%  \item \textbf{RQ1. 안전성}: 합성 데이터가 정규화/어텐션/안정화 기법 등 핵심 구성요소에 장기적 부정효과를 유발하지 않는가? \citep{modelcollapse,syntheticrisk}
%  \item \textbf{RQ2. 토크나이저}: 합성 비중이 높은 환경에서 어떤 토크나이저/어휘 크기/언어 혼합비가 압축 효율과 일반화 간 균형을 최적화하는가? \citep{bpe,sentencepiece,subwordcurse}
%  \item \textbf{RQ3. 편향}: 합성 데이터를 생성한 상위 모델의 언어·문화 편향이 목표 언어의 미묘함을 훼손하지 않는가? \citep{selfinstruct,instructgpt}
%\end{enumerate}

We conduct a systematic investigation into the feasibility and limitations of constructing a \emph{synthetic data-driven} FOM for Korean. We developed a Korean-English bilingual model \emph{from scratch}, wherein synthetic data comprises 68.73\% of the Korean corpus. We curated a training curriculum totaling 150 billion tokens. This curriculum was applied across all major stages of the training pipeline: pre-training, supervised fine-tuning (SFT), and preference learning. To achieve a varied distribution of styles and topics, the synthetic data was produced using a combination of Qwen and models from open-source families (e.g., GPT-OSS). This framework serves as the foundation for addressing the following research questions:
\begin{enumerate}
  \item \textbf{RQ1. Stability}: Does synthetic data introduce long-term adverse effects on core components such as normalization, attention, and stabilization techniques? \citep{modelcollapse,syntheticrisk}
  \item \textbf{RQ2. Tokenizer}: When using a high proportion of synthetic data, what is the optimal configuration of tokenizer, vocabulary size, and language mixture ratio for balancing compression efficiency against generalization? \citep{bpe,sentencepiece,subwordcurse}
  \item \textbf{RQ3. Bias}: When generating synthetic data, are the linguistic and cultural biases of the source model transferred to the new data, damaging or erasing the subtle nuances of the target language? \citep{selfinstruct,instructgpt}
\end{enumerate}

%연구 설계는 두 단계로 이루어진다. 먼저 비용 효율적인 \emph{proxy} 설정(1B 모델, 60B 토큰)에서 합성 100\%와 비합성 100\% 조건을 직접 대조하고, RMSNorm/Pre-LN, 학습률·배치·길이 커리큘럼 등 안정화 기법과 토크나이저/언어 혼합비를 교차 분석한다. 안전성과 효율이 확인되면, 10.4B 규모의 한국어–영어 \emph{fully open} 모델을 구축·공개하며, 데이터 소스, 정제 규칙, 스크립트, 하이퍼파라미터, 로그, 체크포인트까지 포함해 FOM 원칙을 준수한다 \citep{olmo2,bloom}. 평가는 지식·추론·독해·상식·수학 등 표준 벤치마크 전반을 포괄한다(예: MMLU \citep{mmlu}, MMLU-Pro \citep{mmlupro}, ARC \citep{arc}, HellaSwag \citep{hellaswag} 등).

Our research design consists of two stages. First, in a cost-effective \emph{proxy} setting (1B model, 60B tokens), we directly compare a 100\% synthetic data condition with a 100\% non-synthetic condition. In this stage, we cross-analyze stabilization techniques such as RMSNorm/Pre-LN and curricula for learning rate, batch size, and sequence length, along with different tokenizers and language mixture ratios. Once stability and efficiency are confirmed, then we proceed to build and release a 10.8B-scale Korean–English \emph{fully open} model. We adhere to FOM principles by disclosing the entire pipeline, including data sources, filtering rules, scripts, hyperparameters, logs, and checkpoints \citep{olmo2,bloom}. The evaluation covers a wide range of standard benchmarks in knowledge, reasoning, reading comprehension, common sense, and mathematics (e.g., MMLU \citep{mmlu}, MMLU-Pro \citep{mmlupro}, ARC \citep{arc}, HellaSwag \citep{hellaswag}, etc.).

%본 논문의 기여는 다음과 같다. (1) 비영어권에서 \emph{합성 데이터 다수 비중}으로도 FOM 구축이 가능함을 최초로 체계적으로 실증한다. (2) 토크나이저/언어 혼합비/학습 커리큘럼 관점에서 \emph{안정성–효율–일반화}의 상충관계를 분석하고 실무 가이드를 제시한다. (3) 한국어–영어 10.4B \emph{fully-open} 모델(데이터·코드·레시피·로그 포함)을 공개하여 다언어 FOM 연구의 재현성과 접근성을 실질적으로 확장한다.

This paper presents the following contributions: (1) We are the first to systematically demonstrate that it is feasible to build an FOM in a non-English language, even with a \emph{majority proportion of synthetic data}. (2) We examine how various tokenizer settings, language mixture ratios, and training curricula affect the trade-offs between \emph{stability, efficiency, and generalization}, offering practical guidelines based on our results. and (3) By releasing a 10.8B parameter Korean-English \emph{fully-open} model, including data, code, recipes, and logs, we significantly improve the reproducibility and accessibility of multilingual FOM research.

\section{Exploring Training Design Choices}\label{sec:proxy-expr}

% 대규모 언어모델(LLM)의 학습 과정은 모델 아키텍처, 규모(e.g., depth, width, number of layers), learning rate 및 스케줄과 같은 하이퍼파라미터 설정, 그리고 학습 데이터의 구성 및 전처리 전략에 이르기까지 수많은 설계 결정을 포함한다. 기존 연구에서는 주로 소규모의 \textit{proxy} 모델을 활용하여 다양한 설정을 탐색한 뒤 그 결과를 기반으로 최종 모델의 아키텍처를 결정하는 접근이 효과적인 것으로 보고되었다 \citep{kaplan2020scaling,hoffmann2022training}. 이러한 절차는 비용 효율성과 탐색 속도의 균형을 확보할 수 있는 실질적 방법론으로 자리 잡았으며, 최근에는 \emph{compute-optimal} 관점에서의 스케일링 우선순위 또한 정교화되고 있다 \citep{hoffmann2022training}.
The training process of a large language model (LLM) involves numerous design decisions, spanning from the model architecture and scale (e.g., depth, width, number of layers), to hyperparameter settings like learning rate and its schedule, and strategies for the composition and preprocessing of training data. Previous research has reported the effectiveness of an approach where various configurations are first explored using smaller-scale \textit{proxy} models, with the final model's architecture then being determined based on these initial results \citep{kaplan2020scaling,hoffmann2022training}. This approach has proven to be a practical strategy for balancing cost-effectiveness with exploration speed. More recently, scaling priorities from a compute-optimal perspective are also being refined \citep{hoffmann2022training}.

%본 연구 역시 이와 같은 접근을 따르되, \emph{증강(합성) 데이터} 활용에 특화된 상황을 별도로 고려하였다. 대규모 합성 데이터를 핵심 자원으로 사용하는 만큼, 합성 데이터 사용이 모델 학습에 미칠 잠재적 위험성(예: \emph{self-consuming}에 따른 성능 붕괴)을 정량적으로 검증하는 것이 필요하다 \citep{modelcollapse,syntheticrisk}. 이를 위해 1B 규모의 \textit{proxy} 모델을 60B 토큰까지 학습하며 다음 두 가지 질문을 탐구하였다: (1) 합성 데이터로만 학습한 모델과 비합성 데이터로 학습한 모델 간 성능 차이는 무엇인가? (2) 합성 데이터가 학습 안정성(예: loss spike 빈도)에 부정적 영향을 주지 않는가? 이러한 분석을 통해 합성 데이터 기반 학습의 \emph{유효성}과 \emph{안전성}을 실험적으로 검증하였다.
While our study also follows this approach, we give special consideration to a scenario focused on the use of \emph{augmented (synthetic) data}. Given that we use large-scale synthetic data as a primary resource, it is necessary to quantitatively verify the potential risks that this data might pose to model training, such as performance degradation from \emph{self-consuming} loops \citep{modelcollapse,syntheticrisk}. We therefore trained a 1B-scale \textit{proxy} model on 60B tokens to investigate two questions: (1) What are the performance differences between a model trained solely on synthetic data versus one trained on non-synthetic data? (2) Does synthetic data have a negative impact on training stability (e.g., the frequency of loss spikes)? Through this analysis, we experimentally validated the \emph{effectiveness} and \emph{stability} of a synthetic data-driven training approach.

%더불어 토크나이저 학습 과정에서의 데이터 mixture 구성, 토큰 압축률(tokenization compression ratio), 그리고 합성 데이터가 분포에 미치는 영향 등 핵심 설계 요소를 체계적으로 평가하였다. 토크나이저 선택(BPE/Unigram/byte-level), 어휘 크기, 언어 혼합비는 학습 효율과 일반화 성능을 좌우하는 요인으로 여러 연구에서 강조되어 왔다. 이는 특히 비영어권 환경에서 데이터 부족을 합성 데이터로 보완할 때 필수적으로 검토해야 하는 설계 축이다.
In addition, we systematically evaluated core design factors in the tokenizer training process, including the composition of the data mixture, the tokenization compression ratio, and the influence of synthetic data on the overall distribution. The choice of tokenizer (BPE/Unigram/byte-level), vocabulary size, and language mixture ratio have been emphasized in numerous studies as factors that govern training efficiency and generalization performance. These are critical design considerations that must be examined, especially when compensating for data scarcity with synthetic data in non-English contexts.

\begin{table}[h]
\centering
\label{tab:model_configs}
\begin{tabular}{ll}
\toprule
\multicolumn{2}{l}{\textbf{Architecture Details}} \\
\midrule
Number of Total Parameters & 1.33B\\
Number of Embedding Parameters & 525M\\
Number of Non-Embedding Parameters & 805M\\
Vocabulary Size & 128256\\
Hidden Size & 2048 \\
Intermediate Size & 6144 \\
Number of Hidden Layers & 16 \\
Number of Attention Heads & 16 \\
Number of Key/Value Heads & 8 \\
Head Dimension & 128 \\
Attention Dropout & 0.0 \\
Attention Bias & 0.0 \\
Weight tying & False \\
Hidden Activation & \texttt{SwiGLU} \\
Normalizer & RMSNorm\\
RMS Norm Epsilon & $1e-05$ \\
RoPE Theta & $5e+5$ \\
Data type & \texttt{bfloat16} \\
\bottomrule
\end{tabular}
\caption{Proxy Model Default Configurations}
\label{tab:proxy_model}
\end{table}

%Table~\ref{tab:proxy_model}의 설정에 따라 다양한 탐색 실험을 수행하였고, 그 결과를 종합해 \kormo 최종 모델(10.4B)의 학습 디자인을 확정하였다. 최종 설계에는 안정적 학습을 위한 \emph{Pre-LayerNorm} \citep{xiong2020layernorm}과 \emph{RMSNorm} \citep{zhang2019rmsnorm}, 추론 효율을 위한 \emph{Grouped-Query Attention(GQA)} \citep{ainslie2023gqa} 및 \emph{Multi-Query Attention(MQA)} \citep{shazeer2019mqa}, \emph{SwiGLU} 활성화 \citep{shazeer2020glu}와 \emph{RoPE} 위치임베딩 \citep{su2021rope} 등이 반영되었다. 또한 대량 말뭉치 학습을 위한 문서 \emph{packing} 기법을 사용해 시퀀스 길이에 맞춰 데이터를 효율적으로 채워 학습 낭비를 줄였다 \citep{chowdhery2022palm,llama3}. 이 일련의 결정은 scratch 학습을 통한 재현 가능하고 공개 가능한 비영어권 대규모 언어모델 구축에 대한 실질적 설계 지침을 제공한다.
We conducted a series of exploratory experiments as specified in Table~\ref{tab:proxy_model}. The synthesis of these results guided the final design of the \kormo (10.8B) model. The final design incorporates Pre-LN\citep{xiong2020layernorm} and RMSNorm \citep{zhang2019rmsnorm} for stable training; Grouped-Query Attention (GQA) \citep{ainslie2023gqa} and Multi-Query Attention (MQA) \citep{shazeer2019mqa} for inference efficiency; as well as SwiGLU activation \citep{shazeer2020glu} and RoPE positional embeddings \citep{su2021rope}. Additionally, we utilized a document packing technique for large-scale corpus training, which efficiently fills sequences to their maximum length to reduce training waste \citep{chowdhery2022palm,llama3}. This series of decisions provides practical design guidelines for building reproducible and open large-scale language models for non-English languages from scratch.

\subsection{Experiment Settings}
To explore the design choices for model architecture and training methods, a baseline training corpus, evaluation benchmarks, a \textit{proxy} model, and a default training configuration are necessary. This study configures the experiments as follows to make scale-up decisions based on patterns observable even in small-scale models (a \emph{proxy-to-target} transfer). Given the ongoing debate about the \emph{emergent} abilities reported in large-scale models \citep{wei2022emergent,schaeffer2024emergent}, we interpret the performance trends observed at the \emph{proxy} stage with a focus on \emph{relative comparison}.

% 모델 아키텍처 및 학습 방법의 설계 선택을 탐구하기 위해서는 기준이 되는 학습 데이터, 평가 벤치마크, \textit{proxy} 모델, 그리고 기본 학습 설정이 필요하다. 본 연구는 작은 규모의 모델에서도 관찰 가능한 패턴을 기반으로 스케일업 결정을 내리기 위해(이른바 \emph{proxy-to-target} 전이) 다음과 같이 실험을 구성하였다. 대규모 모델에서 보고된 \emph{emergent} 현상에 대한 논쟁이 존재하므로 \citep{wei2022emergent,schaeffer2024emergent}, 우리는 \emph{proxy} 단계에서 나타나는 성능 추세를 \emph{상대 비교} 중심으로 해석한다.

\begin{enumerate}
    \item \textbf{Language:} In the proxy stage, to ensure rich benchmark resources and experimental reliability, we limited our scope to English data to compare synthetic versus non-synthetic setups.
    
    \item \textbf{Train Corpus:} For non-synthetic data, we used 60B tokens randomly sampled from \emph{UltraFineWeb}, which has undergone multi-dimensional filtering \citep{ultrafineweb_2025,fineweb_2024}. During training, document \emph{packing} was applied to match the maximum sequence length, thereby minimizing data waste \citep{chowdhery2022palm}. For the synthetic data experiments, we used 60B tokens from the \emph{high-quality synthetic} split of \emph{Nemotron-CC} \citep{nemotroncc}. (While a broader comparison of various synthetic sources would be ideal, there were constraints in terms of data construction and training costs.)

    \item \textbf{Evaluation Suite:} For the proxy model evaluation, we used MCQA benchmarks with balanced answer distributions and varied difficulty levels. General language understanding and reading comprehension were assessed with RACE \citep{race}, BoolQ \citep{boolq}, and TruthfulQA (TFQA) \citep{truthfulqa}. Science and reasoning were evaluated with ARC-Easy (ARC-E) \citep{arc} and OpenBookQA (OBQA) \citep{obqa}. Commonsense reasoning was evaluated using HellaSwag (HSWG) \citep{hellaswag}, Winogrande (WGRD) \citep{winogrande}, and PIQA \citep{piqa}. For consistent comparison, a default 5-shot setting was used, but due to the nature of the task, TFQA was evaluated in a 0-shot setting (to prevent the model from fabricating confident answers).

    \item \textbf{Model Architecture:} The base structure followed the Llama-3 series architecture, which includes \emph{Pre-LN}, \emph{GQA} \citep{ainslie2023gqa}, \emph{SwiGLU} \citep{shazeer2020glu}, and \emph{RoPE} \citep{su2021rope} \citep{llama3}.

    \item \textbf{Tokenizer:} Since the proxy experiments were conducted solely on English data, we used the well-established BPE-based tokenizer from Llama-3 \citep{llama3}. In Chapter 3, a tokenizer trained on English-Korean \emph{bilingual} data is evaluated separately to reflect the actual target environment.

\end{enumerate}

% \begin{enumerate}
%     \item \textbf{Language: } Proxy 단계에서는 풍부한 벤치마크 리소스와 실험 신뢰성을 고려해 영어 데이터로 한정하여 합성/비합성 설정을 비교하였다.
    
%     \item \textbf{Train Corpus: } 비합성 데이터는 다차원 필터링이 적용된 \emph{UltraFineWeb}에서 무작위로 추출한 60B 토큰을 사용하였다 \citep{ultrafineweb_2025,fineweb_2024}. 학습 시에는 최대 시퀀스 길이에 맞춘 문서 \emph{packing}을 적용해 데이터 낭비를 최소화하였다 \citep{chowdhery2022palm}. 합성 데이터 실험에는 \emph{Nemotron-CC}의 \emph{high-quality synthetic} 분할에서 60B를 사용하였다 \citep{nemotroncc}. (다양한 합성 소스를 더 넓게 비교하는 것이 이상적이지만, 데이터 구축·학습 비용 측면의 제약이 있었다.)
    
%     \item \textbf{Evaluation Suite: } Proxy 모델 평가에는 정답 분포가 균형적이고 난이도가 다양한 MCQA 벤치마크를 사용하였다: 일반 언어 이해/독해는 RACE \citep{race}, BoolQ \citep{boolq}, TruthfulQA(TFQA) \citep{truthfulqa}; 과학 및 추론은 ARC-Easy(ARC-E) \citep{arc}와 OpenBookQA(OBQA) \citep{obqa}; 상식 추론은 HellaSwag(HSWG) \citep{hellaswag}, Winogrande(WGRD) \citep{winogrande}, PIQA \citep{piqa}로 평가하였다. 일관성 있는 비교를 위해 기본 5-shot을 사용하되, 태스크 특성상 TFQA는 0-shot으로 평가하였다(모델의 허구적 확신을 방지하기 위함).
    
%     \item \textbf{Model Architecture: } 기본 구조는 \emph{Pre-LayerNorm}, \emph{GQA} \citep{ainslie2023gqa}, \emph{SwiGLU} \citep{shazeer2020glu}, \emph{RoPE} \citep{su2021rope} 등을 포함한 Llama-3 계열 아키텍처를 따랐다 \citep{llama3}. 
    
%     \item \textbf{Tokenizer: } Proxy 실험은 영어 데이터만을 대상으로 하므로 범용성이 검증된 Llama-3의 BPE 기반 토크나이저를 사용하였다 \citep{llama3}. 3장에서는 실제 목표 환경을 반영하기 위해 영어–한국어 \emph{bilingual} 데이터로 학습된 토크나이저를 별도로 평가한다.
% \end{enumerate}

\subsection{Normalization Methods}
Normalization plays a key role in both training stability and performance improvement, and it is broadly categorized into \textit{post-LN} and \textit{Pre-LN} depending on its placement. \citet{xiong2020layernorm} theoretically and empirically analyzed the differences in initialization and gradient behavior between the two methods, showing that \emph{Pre-LN} converges more stably in large-scale pre-training. Additionally, \emph{RMSNorm} was proposed as an alternative to reduce the computational cost of LayerNorm \citep{zhang2019rmsnorm}, and stabilization techniques for very deep transformers (hundreds to thousands of layers), such as DeepNorm, have also been reported \citep{wang2022deepnet}. More recently, hybrid approaches (such as \textit{MixLN}) that mix different normalization methods in the initial and later layers have been discussed in this context, aiming to mitigate the gradient vanishing problem in very deep networks.

In this study, we hypothesized that the choice of normalization is directly related to \textbf{RQ1: Does augmented data negatively affect training stability?} This is because if augmented data exacerbates instability under a specific normalization method, it could become a critical risk factor in large-scale training.

% Normalization은 학습 안정성과 성능 향상 모두에 핵심적인 역할을 하며, 적용 위치에 따라 크게 \textit{post-LN}과 \textit{pre-LN}으로 구분된다. \citet{xiong2020layernorm}은 두 방식의 초기화·그래디언트 거동 차이를 이론·실험적으로 분석하며, 대규모 사전학습에서 \emph{pre-LN}이 더 안정적으로 수렴함을 보였다. 또한 LayerNorm의 연산비용을 줄이는 대안으로 \emph{RMSNorm}이 제안되었고 \citep{zhang2019rmsnorm}, 초심층(수백~천 레이어) 트랜스포머를 위한 안정화 기법(예: DeepNorm)도 보고되었다 \citep{wang2022deepnet}. 최근에는 이러한 맥락에서 초기/후반 레이어에 서로 다른 정규화를 혼합하는 하이브리드 접근(\textit{MixLN} 등)이 논의되고 있으며, 매우 깊은 네트워크에서의 gradient vanishing을 완화하는 것을 목표로 한다.

% 본 연구에서는 normalization 선택이 \textbf{RQ1: 증강 데이터가 학습 안정성에 부정적 영향을 미치지 않는가?}와 직접적으로 연관된다고 가정하였다. 증강 데이터가 특정 normalization 하에서 불안정성을 가중시킨다면 대규모 학습에서 치명적 위험 요인이 될 수 있기 때문이다. 

\begin{table}[!h]
\begin{adjustbox}{max width=\textwidth}
\centering
\begin{tabular}{llccccccccccc}
\toprule
Norm. type &      data    & arc\_e & boolq & hswg   & obqa  & piqa   & race   & tfqa\_mc1 & tfqa\_mc2  & wgrd    & AVG \\
\midrule
Pre-LN     & Web 	& 65.24	& 56.54	 & 37.86  & 23.0  & 70.24  & 31.77  & 19.95  & 32.84  & 52.96  & 43.38\\
MixLN      & Web    & 60.52	& 52.54	 & 35.18  & 20.6  & 68.23  & 29.19  & 20.44  & 36.13  & 48.70  & 41.28\\
\midrule
Pre-LN & Synthetic & 68.81 & 58.81 & 38.10 & 23.8 & 72.58 & 32.54 & 25.70 & 35.67 & 54.62 & \textbf{45.63}\\

\bottomrule
\end{tabular}
\end{adjustbox}
\caption{Performance comparison across normalization methods (Pre-LN, MixLN) and data types. Pre-LN consistently outperforms MixLN, and the use of synthetic data introduces no performance degradation.}
% \caption{LayerNorm}
\label{table:layernorm}
\end{table}

In the experiment, MixLN was configured by applying Post-LN to the first two layers, which constitute 12.5\% of the total layers, and Pre-LN to the remainder. As shown in Table~\ref{table:layernorm}, Pre-LN was consistently superior to MixLN in average performance (43.38\% vs. 41.28\%). Therefore, Pre-LN was adopted as the default normalization method in the final architecture. A more critical observation is that when comparing the model trained with synthetic (augmented) data against the non-synthetic data model under the same Pre-LN setup, no performance degradation or increase in \emph{loss spikes} was observed. This suggests that, from a normalization standpoint, synthetic data does not introduce additional instability, providing \textbf{positive evidence for RQ1.}

However, these results were obtained from a 1B parameter \textit{proxy} model. Techniques like the MixLN family or stabilization methods for very deep networks such as DeepNorm may show more prominent potential at a larger model scale \citep{wang2022deepnet}. Therefore, a re-evaluation is necessary for models with tens to hundreds of billions of parameters.

% 실험에서는 전체 레이어 중 12.5\%에 해당하는 앞 2개 레이어에 Post-LN을 적용하고, 나머지에는 Pre-LN을 적용하여 MixLN을 구성하였다. 표~\ref{table:layernorm}에서 보이듯, Pre-LN이 평균 성능(43.38\%)에서 MixLN(41.28\%)보다 일관되게 우수하였다. 따라서 최종 아키텍처에서는 Pre-LN을 기본 normalization 방식으로 채택하였다. 더 중요한 관찰은 동일한 Pre-LN 설정 하에서 합성(증강) 데이터로 학습한 모델과 비합성 데이터 모델을 비교했을 때 성능 저하나 \emph{loss spike} 증가가 관찰되지 않았다는 점이다. 이는 normalization 관점에서 합성 데이터가 추가적인 불안정성을 야기하지 않음을 시사하며, \textbf{RQ1에 대한 긍정적 근거}를 제공한다.

% 다만 본 결과는 1B 파라미터의 \textit{proxy} 모델에서 얻어진 것이다. MixLN 계열이나 DeepNorm과 같은 초심층 안정화 기법은 거대 모델 규모에서 잠재력이 더 두드러질 수 있으므로 \citep{wang2022deepnet}, 수십억~수천억 파라미터 모델에서는 재평가가 필요하다.

%(민준)
\subsection{Attention Masking Method}
Attention masking has a significant impact on both training efficiency and performance because it directly dictates how context is formed in long sequence \textit{packing}. We compared four masking strategies using a \textit{proxy} model, and improved implementation efficiency by leveraging flash-style kernels \citep{dao2022flashattention,dao2023flashattention2}.

\begin{itemize}
    \item \textbf{Causal masking:} The standard method in which every token attends to all previous tokens.
    \item \textbf{Sliding causal masking:} Tokens attend only to other tokens within a predefined window (1024 in this experiment). Windowed attention for long contexts is widely used in models like Longformer \citep{beltagy2020longformer}.
    \item \textbf{Intra-document causal masking (Intra-doc):} Blocks cross-document attention, allowing tokens to attend only to other tokens within the same document. A recent comparative study reports that \emph{intra-doc} can improve downstream performance by reducing the noise that arises between documents during packing \citep{zhao2024packing}.
    \item \textbf{Sliding intra-doc masking:} Combines intra-doc masking with a sliding window.
    \end{itemize}
\begin{table}[h!]
\begin{adjustbox}{max width=\textwidth}
\centering
\begin{tabular}{lcccccccccccc}
\toprule
Masking type  & data      & arc\_e & boolq & hswg   & obqa  & piqa   & race   & tfqa\_mc1 & tfqa\_mc2  & wgrd    & AVG \\
\midrule
Causal    & Web       	& 65.24	& 56.54	 & 37.86  & 23.0  & 70.24  & 31.77  & 19.95  & 32.84  & 52.96  & 43.38\\
Sliding causal	  & Web    & 67.13	& 54.43	 & 37.75  & 23.0  & 71.22  & 32.54  & 19.22  & 32.61  & 52.09  & 43.33\\
Intra-doc	 & Web     & 67.68	& 56.88	 & 38.10  & 23.8  & 71.71  & 31.87  & 20.20  & 35.67  & 54.38  & 44.48\\
Sliding Intra-doc & Web 	& 67.80	& 53.33	 & 37.92  & 22.4  & 70.89  & 31.96  & 20.69  & 34.64  & 52.72  & 43.59\\

\midrule
Intra-doc & Synthetic               & 68.81	& 61.10	 & 39.40  & 21.8  & 72.58  & 34.35  & 22.40  & 36.98  & 53.51  & \textbf{45.66}\\

\bottomrule
\end{tabular}
\end{adjustbox}
\caption{Performance evaluation across different attention masking strategies and data types. Intra-doc masking was the most effective, and the model trained on synthetic data achieved higher performance under the same conditions.}
% \caption{Attention masking}
\label{table:attention mask}
\end{table}

According to Table~\ref{table:attention mask}, Intra-doc masking achieved the highest performance with an average of 44.48\%, surpassing standard causal masking (43.38\%). This suggests that the strategy of strengthening the independent context of each document by eliminating unnecessary contextual connections between them is effective \citep{zhao2024packing}. In contrast, the sliding-based methods may degrade performance by cutting off necessary long-range dependencies in long documents (similar observations are reported in the literature on windowed attention \citealp{beltagy2020longformer}).

Based on these results, Intra-doc masking was adopted for the final model. Under the same Intra-doc setting, the model trained on synthetic data performed better than its non-synthetic counterpart, with almost no difference in the learning curve and \emph{loss} volatility. This serves as additional evidence that synthetic data does not impair stability from an attention masking perspective, which further supports our findings for RQ1. However, as this experiment was conducted in a monolingual English setting, the control strategy for document boundaries and cross-references needs further exploration in a multilingual context. For instance, we did not explore methods like Cross-Lingual Document Attention (XLDA, \citealp{han2025trillion}), which could be a promising direction for future work.

\subsection{Multi-Token Prediction}
Currently, large language models (LLMs) are primarily pre-trained and perform inference using the Next-Token Prediction (NTP) objective. However, the inherently sequential nature of NTP introduces significant latency, especially during inference, which constrains the real-time application of these models.~\citep{gloeckle2024betterfasterlarge}
Furthermore, its focus on predicting only a single token leaves room for optimization in terms of potential training efficiency.~\citep{gloeckle2024betterfasterlarge} To overcome these limitations of NTP, Multi-Token Prediction (MTP) has emerged as a noteworthy methodology.~\citep{deepseekai2025deepseekv3technicalreport, qwen3technicalreport} MTP trains the model to predict several subsequent tokens simultaneously at each prediction step, offering two primary advantages: improved pre-training efficiency and accelerated inference speed.

In this context, we investigate the impact of the Multi-Token Prediction (MTP) objective, originally proposed in the DeepSeek-V3 architecture, on pre-training efficiency and downstream performance in small-scale language models.

Specifically, we set our primary goal to conduct an in-depth analysis of the applicability and effectiveness of MTP on a model with approximately 1 billion parameters. To integrate MTP into the base architecture, it is necessary to extend the conventional Next-Token Prediction (NTP; \(n=1\)) loss to calculate a loss based on multi-token predictions. %The MTP loss calculation used in this study consists of two stages. First, for a prediction depth of \(k\), the individual loss \(\mathcal{L}_{\text{MTP}}^{k}\) is defined as in Equation~\ref{eq:mtp_k}. Here, the prediction depth \(k\) indicates how many future tokens the model predicts ahead, and the loss is measured by the cross-entropy between the model's output distribution (\(p\)) and the ground-truth sequence (\(t\)) at that time step.

% \begin{equation} \label{eq:mtp_k}
%     \mathcal{L}_{\text{MTP}}^{k} = \text{CrossEntropy}(p_{2+k:T+1}^{k}, t_{2+k:T+1}) = -\frac{1}{T} \sum_{i=2+k}^{T+1} \log p_{i}^{k}[t_{i}]
% \end{equation}

% Then, the final MTP loss \(\mathcal{L}_{\text{MTP}}\) is calculated by averaging the individual losses for all depths and multiplying by a weight \(\lambda\) (Equation~\ref{eq:mtp_final}).

% \begin{equation} \label{eq:mtp_final}
%     \mathcal{L}_{\text{MTP}} = \frac{\lambda}{D} \sum_{k=1}^{D} \mathcal{L}_{\text{MTP}}^{k}
% \end{equation}

\begin{table}[h!]
\begin{adjustbox}{max width=\textwidth}
\centering
\begin{tabular}{lccccccccccc}
\toprule
Objective type        & arc\_e & boolq & hswg   & obqa  & piqa   & race   & tfqa\_mc1 & tfqa\_mc2  & wgrd    & AVG \\
\midrule
NTP              	& 65.24	& 56.54	 & 37.86  & 23.0  & 70.24  & 31.77  & 19.95  & 32.84  & 52.96  & \textbf{43.38}\\
MTP	                & 49.12	& 58.81	 & 29.57  & 27.6  & 61.53  & 26.51  & 25.70  & 41.72  & 51.62  & 41.35\\

\bottomrule
\end{tabular}
\end{adjustbox}
\caption{Performance comparison of NTP and MTP training objectives. On the 1B-scale model, NTP showed higher overall average performance than MTP.}
% \caption{MTP}
\label{table:mtp}
\end{table}

Table~\ref{table:mtp} presents a direct comparison between Next Token Prediction (NTP) and Multi-Token Prediction (MTP) under identical pre-training conditions. Overall, MTP showed slightly lower average performance compared to NTP (41.35 vs. 43.38, approx. –2.0\%p), suggesting that NTP remains a more effective general-purpose training objective for small-scale models.

However, a different pattern was observed on a task-by-task basis. MTP outperformed NTP on Question Answering (QA) benchmarks such as BoolQ (+2.3\%p), OBQA (+4.6\%p), TFQA\_mc1 (+5.8\%p), and TFQA\_mc2 (+8.9\%p). This suggests that the training objective of predicting multiple tokens simultaneously provided a stronger signal for tasks requiring multi-step reasoning or factual recall. In contrast, NTP showed a clear advantage on tasks that rely on precise next-token prediction, such as ARC-E, PIQA, and RACE.

These results are consistent with findings reported in previous research~\cite{aynetdinov2025pretrainingcurriculummultitokenprediction}. Small-scale language models may struggle to effectively handle the training complexity of MTP. Without sufficient capacity, they can show limitations in learning complex patterns and in generalization. In particular, the 1B parameter model used in this study, unlike models with tens of billions of parameters such as DeepSeek-V3~\cite{mehra2025multitokenpredictionefficientllm}, can be interpreted as having structural and capacity constraints that prevent it from realizing the full potential of MTP.

The scale of the data is also a critical factor. This study used a 60B token English web corpus, which is significantly smaller than the trillions of tokens used by recent very large-scale models. For example, DeepSeek-V3 was trained on 14.8 trillion tokens, and it is likely that a complex objective like MTP is more effective with such vast amounts of data. In contrast, in a limited data environment, MTP might lead to overfitting or fail to sufficiently learn complex relationships as intended by its design.

In summary, we confirmed that NTP is a more stable and efficient training objective under the conditions of a small-scale model and limited data. Therefore, this study ultimately adopted NTP as the training objective, and no additional MTP experiments were conducted on the augmented data-based models.

% Table~\ref{table:mtp}는 동일한 사전학습 조건에서 \textit{Next Token Prediction} (NTP)과 \textit{Multi-Token Prediction} (MTP)을 직접 비교한 결과를 제시한다. 전체적으로 MTP는 평균 성능에서 NTP 대비 다소 낮은 결과를 보였으며 (41.35 vs. 43.38, 약 –2.0\%p), 이는 소규모 모델에서는 여전히 NTP가 일반 목적 학습 목표로서 더 효과적임을 시사한다.  

% 그러나 태스크별로는 상이한 양상이 관찰되었다. MTP는 BoolQ(+2.3\%p), OBQA(+4.6\%p), TFQA\_mc1(+5.8\%p), TFQA\_mc2(+8.9\%p)와 같은 질의응답(QA) 벤치마크에서 NTP를 능가하였다. 이는 다중 토큰을 동시에 예측하는 학습 목표가 다단계 추론이나 사실적 회상을 요구하는 과제에서 보다 강력한 신호를 제공했음을 시사한다. 반면, ARC-E, PIQA, RACE와 같이 정밀한 다음 토큰 예측에 의존하는 과제에서는 NTP가 뚜렷한 우위를 보였다. 

% 이러한 결과는 기존 연구에서 보고된 바와 일치한다~\cite{aynetdinov2025pretrainingcurriculummultitokenprediction}. 소규모 언어모델(SLM)은 MTP의 학습 복잡성을 효과적으로 소화하기 어렵고, 충분한 용량을 갖추지 못할 경우 복잡한 패턴 학습과 일반화에 한계를 보일 수 있다. 특히 본 연구에서 사용한 1B 파라미터 규모 모델은 DeepSeek-V3와 같은 수백억 파라미터 규모의 모델~\cite{mehra2025multitokenpredictionefficientllm}과 달리, MTP의 잠재력을 발휘하기에 구조적·용량적 제약이 있었다고 해석할 수 있다.  

% 데이터 규모 또한 중요한 요인이다. 본 연구는 60B 토큰의 영어 웹 코퍼스를 사용하였는데, 이는 최근 초대규모 모델이 활용하는 수조 단위 토큰 대비 현저히 작은 규모이다. 예컨대 DeepSeek-V3는 14.8조 토큰으로 학습되었으며, 이와 같은 방대한 데이터에서야 MTP와 같은 복잡한 목표가 효과를 발휘할 가능성이 크다. 반면 제한된 데이터 환경에서는 MTP가 오버피팅을 유발하거나, 설계 의도대로 복잡한 관계를 충분히 학습하지 못할 수 있다.  

% 종합하면, 소규모 모델과 제한된 데이터 조건에서는 NTP가 더 안정적이고 효율적인 학습 목표임을 확인하였다. 따라서 본 연구에서는 최종적으로 NTP를 학습 목표로 채택하였으며, 증강 데이터 기반 모델에 대해서는 MTP 실험을 추가적으로 진행하지 않았다.  

%(인호, 한결)
\section{Proposed Tokenizer based on Mixture of Datasets}

% 토크나이저는 raw text를 token sequence로 변환하여 언어 모델에 입력할 수 있도록 하는 핵심 도구이다. 동일한 텍스트를 더 짧은 sequence로 변환할수록(즉, 더 나은 compression) 학습·추론 효율이 높아진다. 따라서 토크나이저 설계에서 compression 성능을 향상시키는 것은 중요 과제다. 본 연구는 bilingual 환경에서 compression을 개선하는 방법을 탐구하며, RQ2의 관점에서 \emph{synthetic 데이터가 토크나이저 성능에 미치는 영향}을 중점적으로 분석한다 \citep{subwordcurse}.
A tokenizer serves as a fundamental component that transforms raw text into a sequence of tokens interpretable by language models. Achieving shorter sequences for identical text inputs, indicating superior compression, directly enhances both training and inference efficiency. Consequently, improving compression performance constitutes a central challenge in tokenizer design. In this study, we investigate strategies to enhance compression within bilingual environments and, from the perspective of RQ2, conduct an in-depth analysis of \emph{the influence of synthetic data on tokenizer performance} \citep{subwordcurse}.

\subsection{Experiments Settings for Building Tokenizer from Scratch}
% 실험에 사용한 토크나이저는 최근 대형 언어모델에서 널리 쓰이는 \emph{byte-level BPE}로 학습하였다 \citep{radford2019gpt2}. byte-level BPE는 입력을 문자 대신 바이트 단위에서 시작하므로 모든 유니코드 텍스트를 손실 없이 처리할 수 있고 사실상 OOV 문제가 발생하지 않는다. 
The tokenizer employed in our experiments was trained using the \emph{byte-level Byte Pair Encoding} algorithm, which has been widely adopted in recent large language models \citep{radford2019gpt2}. Since byte-level BPE begins segmentation at the byte level rather than the character level, it can process any Unicode text without information loss and effectively eliminates the out-of-vocabulary (OOV) problem.

\paragraph{Experimental Setup} 
% byte-level BPE로 토크나이저를 학습할 때 \emph{synthetic 비중}이 compression에 미치는 영향을 분석했다. synthetic 데이터로는 Cosmopedia(영)와 ko-Cosmopedia(한, 자체 구축)를 각 20GB 샘플링했고, 대비 코퍼스로 UltraFineWeb(영)과 FineWeb2(한)를 각 20GB 샘플링했다 \citep{ultrafineweb_2025,fineweb_2024}. 검증은 \emph{non-synthetic} 측정을 위해 The Pile 일부를, \emph{synthetic} 측정을 위해 각 언어별 RLHF 데이터 일부를 사용했다 \citep{pile,hh-rlhf}. 선택 배경은 이들 세트가 일반 웹 도메인 외 다양한 분야를 포함해 압축 성능을 보다 종합적으로 확인하기에 적합하다고 판단했기 때문이다. Table~\ref{table:tokenizer-data}는 데이터 구성을 요약한다. 모델은 앞서 제안한 \textit{proxy} 모델에 \emph{토크나이저만} 교체하여 압축률과 다운스트림 성능을 비교했다.
We analyzed the effect of the \emph{synthetic data proportion} on compression when training tokenizers using the byte-level BPE algorithm. As synthetic data sources, we sampled 20 GB each from Cosmopedia (English) and ko-Cosmopedia (Korean, internally constructed) \citep{benallal2024cosmopedia}. As representative non-synthetic corpora, we sampled 20 GB each from UltraFineWeb (English) and FineWeb2 (Korean) \citep{ultrafineweb_2025,fineweb_2024}.
For evaluation, a subset of The Pile was used for non-synthetic measurements, while a portion of RLHF datasets for each language was used for synthetic measurements \citep{pile,hh-rlhf}. These datasets were selected because, beyond general web domains, they encompass diverse fields, allowing a more comprehensive assessment of compression performance.
Table~\ref{table:tokenizer-data} summarizes the data composition. The model follows the previously proposed configuration.

\paragraph{Evaluation Metrics}
% compression은 \emph{Bytes Per Token (BPT)}로 측정하였다. 동일 텍스트에서 더 짧은 토큰열을 생성하는 토크나이저일수록 BPT가 높아지며, 이는 더 효율적 압축을 의미한다. 단, 높은 압축이 항상 더 나은 다운스트림을 보장하지 않음이 보고되어 왔기에 \citep{tokenizerimpact}, 본 연구는 BPT와 과업 성능을 함께 평가했다. bilingual 시나리오 반영을 위해 한국어(Haerae, KMMLU, KoBEST)와 영어 벤치마크를 함께 사용했다 \citep{haerae,kmmlu,kobest}.
Compression was measured using \emph{Bytes Per Token} (BPT). A tokenizer that produces shorter token sequences for the same text yields a higher BPT, which indicates more efficient compression. However, since prior studies have shown that higher compression does not necessarily guarantee better downstream performance \citep{tokenizerimpact,zuo2025falcon}, this study jointly evaluates both BPT and model downstream performance.
To reflect the bilingual scenario, we evaluated downstream model performance using benchmark datasets in both Korean (Haerae, KMMLU, and KoBEST) \citep{haerae,kmmlu,kobest} and English.

\begin{table}[t]
\centering
\small
\setlength{\tabcolsep}{6pt}
\begin{tabular}{llcccc}
\toprule
\multicolumn{2}{c}{\textbf{Split / Corpus}} & \textbf{Lang} & \textbf{Category} & \textbf{Sampling} & \textbf{Size} \\
\midrule
\multirow{4}{*}{\textbf{Train}} 
& Cosmopedia & EN & Synthetic (encyclopedic) & Random & 20\,GB \\
& ko-Cosmopedia & KO & Synthetic (encyclopedic) & Random & 20\,GB \\
& UltraFineWeb & EN & Web (non-synthetic) & Random & 20\,GB \\
& FineWeb-2 & KO & Web (non-synthetic) & Random & 20\,GB \\
\midrule
\multirow{2}{*}{\textbf{Val. (Web mix)}} 
& The Pile (subset) & EN & Web mix & Random & 33\,MB \\
& culturaX (subset) & KO & Web mix & Random & 33\,MB \\
\multirow{2}{*}{\textbf{Val. (Synthetic)}} 
& RLHF (per-lang) & EN & Synthetic (RLHF) & Random & 33\,MB \\
& RLHF (per-lang) & KO & Synthetic (RLHF) & Random & 33\,MB \\
\bottomrule
\end{tabular}
\caption{Tokenizer compression study setup. We train a byte-level BPE tokenizer and analyze how \emph{synthetic} corpora (Cosmopedia, ko-Cosmopedia) versus \emph{non-synthetic web} corpora (UltraFineWeb, FineWeb-2) affect compression performance. Each training corpus is randomly sampled to 20\,GB. Validation uses (i) web-crawled subsets (\textit{The Pile}, \textit{cultura-ko-x}) and (ii) per-language synthetic RLHF data.}
\label{table:tokenizer-data}
% \vspace{-1mm}
\end{table}

\subsection{Impact of Synthetic Data on Tokenizer Training}

\begin{figure}[h]
    \centering
    \makebox[\textwidth][c]{%
        \includegraphics[width=\textwidth]{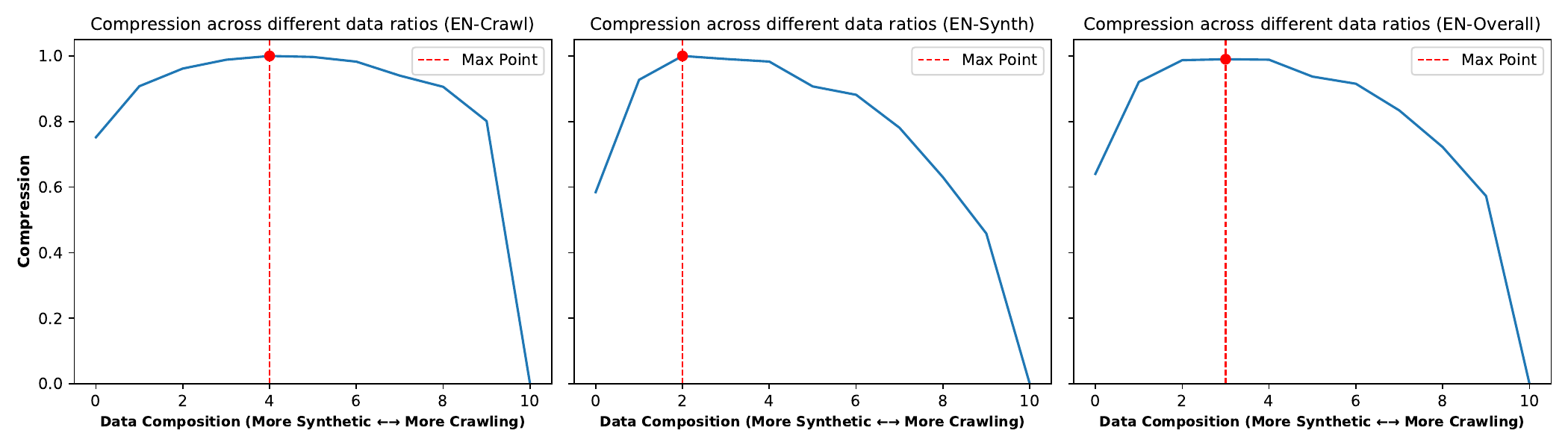}
    }
    \caption{Compression trends by data ratio in the English setting. The x-axis represents the synthetic–crawl ratio (left: synthetic-dominant, right: crawl-dominant), and the y-axis shows compression efficiency measured in bytes per token (BPT), where higher values indicate greater efficiency.}
    \label{fig:en_compression_wrt_synthetic}
\end{figure}

\begin{figure}[h]
    \centering
    \makebox[\textwidth][c]{%
        \includegraphics[width=\textwidth]{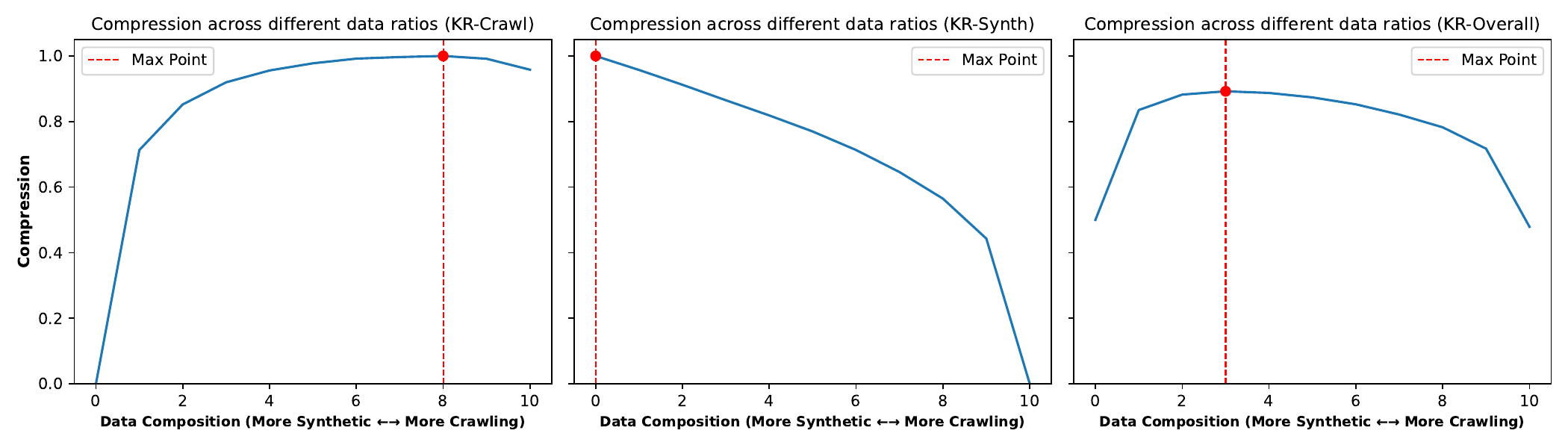}
    }
    \caption{Compression trends by data ratio in the Korean setting. The x-axis represents the synthetic–crawl ratio (left: synthetic-dominant, right: crawl-dominant), and the y-axis shows compression efficiency measured in bytes per token (BPT), where higher values indicate greater efficiency.}
    \label{fig:kr_compression_wrt_synthetic}
\end{figure}

% Figure~\ref{fig:en_compression_wrt_synthetic}, \ref{fig:kr_compression_wrt_synthetic}는 각각 영어·한국어 설정에서의 결과다. X축은 \emph{synthetic}과 \emph{crawling} 데이터 비율, Y축은 BPT 기반 압축률을 나타낸다. 전반적으로 두 언어에서 \emph{synthetic 비중을 높일 때 평균적인 compression 이득}이 관찰되었다. 세부 도메인 분석에서는 다음과 같은 차이가 있었다. 
Figure~\ref{fig:en_compression_wrt_synthetic} and ~\ref{fig:kr_compression_wrt_synthetic} present the results for English and Korean settings, respectively. The X-axis represents the ratio between synthetic and crawled data, while the Y-axis indicates compression performance measured in BPT. Overall, both languages exhibit an average compression gain as the proportion of synthetic data increases. A closer examination by domain, however, reveals several notable differences.

% 영어의 경우, 크롤링 도메인에서도 \(\sim\)60\% synthetic 비중에서 더 높은 compression을 보였다(압축·일반화 간 trade-off 가능성은 \citealp{tokenizerimpact} 참조). 반면 한국어에서는 크롤링 기반 도메인에서 \(\sim\)80\% 크롤링 비중이 필요했다. 이는 synthetic과 크롤링 데이터의 \emph{분포 차}가 영어보다 한국어에서 더 크게 나타났음을 시사하며, 한국어 특화 지식/문자 특성에 대한 반영 수준과도 관련이 있다 \citep{haerae,kmmlu,kobest}.
For English, even within crawled domains, higher compression was observed when the synthetic proportion reached approximately 60\% (see \citealp{tokenizerimpact} for discussion on the potential trade-off between compression and generalization).
In contrast, for Korean, a substantially higher crawled data proportion—around 80\%—was required to achieve comparable compression. This suggests that the distributional gap between synthetic and crawled data is more pronounced in Korean than in English, potentially reflecting differences in language-specific knowledge and character-level properties \citep{haerae,kmmlu,kobest}.

\begin{table*}[!h]
\centering
\small
\begin{tabular}{lccccccc}
    \toprule
    \multirow{2}{*}{\textbf{Alias}} & \multicolumn{2}{c}{\textbf{English}} & \multicolumn{2}{c}{\textbf{Korean}}  & \multirow{2}{*}{\textbf{Code}} & \multirow{2}{*}{\textbf{Data Size}} &\multirow{2}{*}{\textbf{Vocab Size}}\\
    \cmidrule(lr){2-3}\cmidrule(lr){4-5}
                                    & Crawling         & Synthetic         & Crawling         & Synthetic & & &\\

    \midrule
    \multirow{2}{*}{EK-Crawl}    & \multirow{2}{*}{25.5\%} & \multirow{2}{*}{59.5\%} & \multirow{2}{*}{4.5\%} & \multirow{2}{*}{10.5\%} & \multirow{2}{*}{-} & \multirow{2}{*}{20GB}& 125k \\
    &  &  &  &  &   & & 196k \\
    \midrule
    \multirow{2}{*}{EK}    & \multirow{2}{*}{-} & \multirow{2}{*}{85\%} & \multirow{2}{*}{-} & \multirow{2}{*}{15\%} & \multirow{2}{*}{-} & \multirow{2}{*}{20GB}& 125k \\
    &  &  &  &  &   & & 196k \\

    \midrule
    \multirow{2}{*}{EPK}    & \multirow{2}{*}{-} & \multirow{2}{*}{80\%} & \multirow{2}{*}{-} & \multirow{2}{*}{5\%} & \multirow{2}{*}{15\%} & \multirow{2}{*}{20GB} & 125k \\
    &  &  &  &  &   & & 196k \\
    \bottomrule
\end{tabular}
\caption{Data composition and configuration details of the proposed tokenizer candidates, showing the proportion of crawled, synthetic, and code data used for training under two vocabulary scales (125k and 196k).}
\label{tab:candidate_tokenizer}
\end{table*}

% 이 관찰을 바탕으로 세 가지 설계 원칙—(1) 언어별 최적 혼합 반영, (2) synthetic의 전반적 우수성 반영, (3) 도메인 취약점 보완—에 따라 토크나이저 후보군을 구성했다. 
% 첫째, \textbf{EK-Ratio}는 Figure~\ref{fig:en_compression_wrt_synthetic}, \ref{fig:kr_compression_wrt_synthetic}에서 관찰된 각 언어의 최적 비율을 토크나이저 학습 혼합에 그대로 반영한다. 
% 둘째, \textbf{EK}는 한–영 비율은 유지하되 두 언어 모두 \emph{synthetic 전용} 데이터로 학습해 synthetic의 평균 우위를 활용한다. 
% 셋째, \textbf{EPK}는 code 도메인 약점을 보완하기 위해 code 데이터를 추가 혼합한다. 
% 각 후보는 vocabulary 125k/196k 두 규모로 독립 학습했으며, 이후 상용 토크나이저와의 BPT 및 다운스트림 성능을 비교·분석한다.
Based on these observations, we constructed three tokenizer design candidates following three principles: (1) reflecting the optimal language-specific mixture, (2) leveraging the overall superiority of synthetic data, and (3) compensating for domain weaknesses.
First, \textbf{EK-Ratio} incorporates the optimal synthetic–crawled ratios for each language, as observed in Figures~\ref{fig:en_compression_wrt_synthetic} and \ref{fig:kr_compression_wrt_synthetic}, directly into the tokenizer training mixture.
Second, \textbf{EK} maintains the same English–Korean ratio but trains exclusively on synthetic data for both languages to exploit its overall advantage.
Third, \textbf{EPK} augments the mixture with additional code data to address weaknesses in code-related domains.
Each tokenizer candidate was independently trained with vocabulary sizes of 125K and 196K. Subsequent analyses compare their BPT and downstream performance against commercial tokenizers.

\subsection{Compression Comparison in Bilingual Settings with Commercial Tokenizers}
\begin{figure}[h]
    \centering
    \makebox[\textwidth][c]{%
        \includegraphics[width=\textwidth]{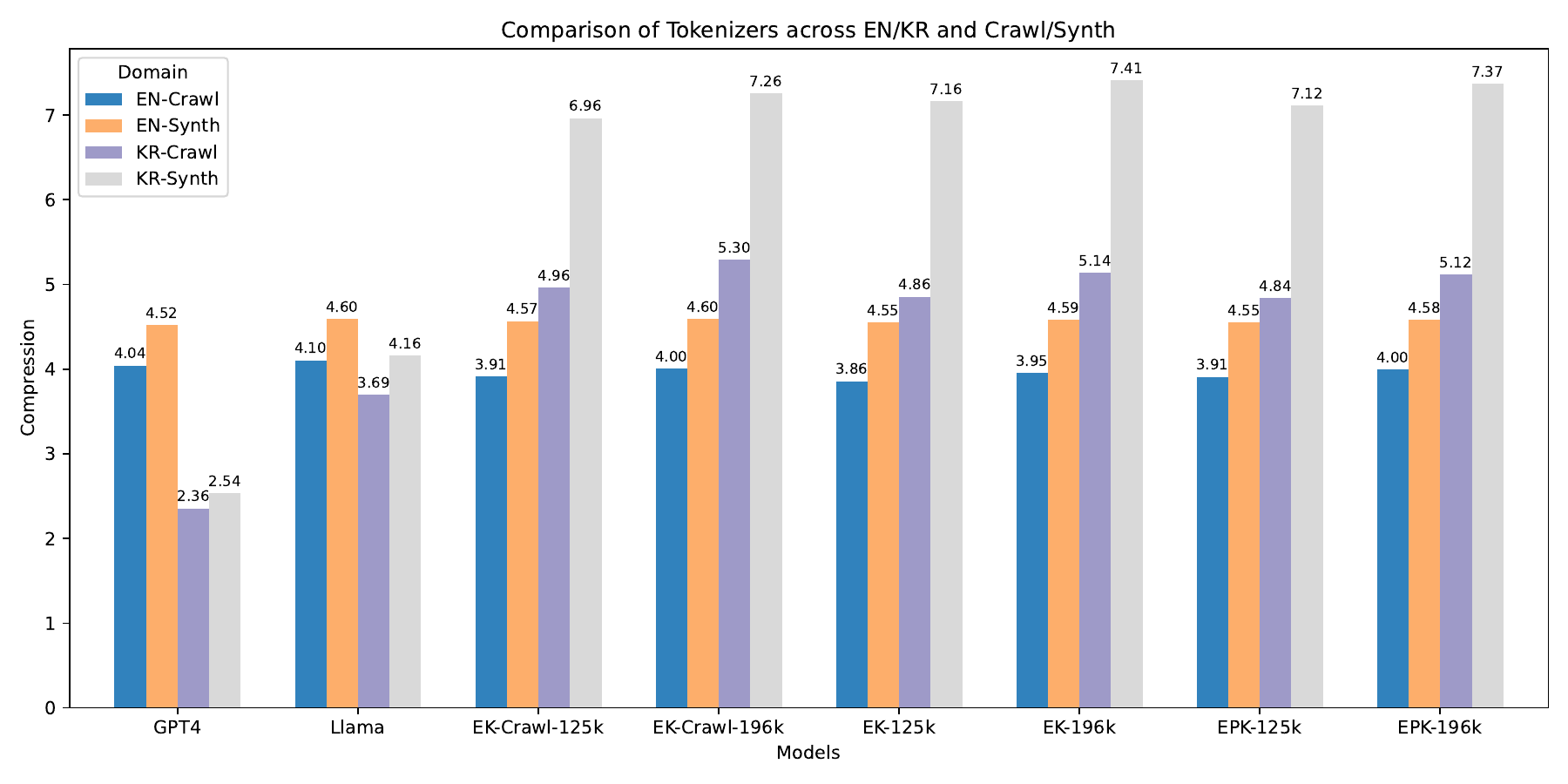}
    }
    \caption{Comparison of English and Korean compression performance between the tokenizer candidates defined in Table~\ref{tab:candidate_tokenizer} and commercial tokenizers (GPT-4, LLaMA).}
    \label{fig:compression_all}
\end{figure}

% 우리는 먼저 선정된 후보 토크나이저가 상용 모델(e.g., Llama, GPT-4)의 토크나이저와 견줄 만큼의 compression 성능을 보이는지 확인하였다. Figure~\ref{fig:compression_all}는 후보군과 상용 모델 토크나이저의 compression 결과를 요약한다. Llama 계열의 byte-level BPE 기반 토크나이저는 영어에서 강점을 보이는 반면 \citep{llama3,radford2019gpt2}, 한국어에서는 상대적으로 미흡했다. 반대로 본 연구의 후보 토크나이저는 영어 도메인에서 상용 수준과 유사한 성능을 유지하면서, 한국어에서는 상용 대비 현저한 우위를 보였다. 이는 \emph{bilingual} 혼합·도메인 조정이 비라틴 문자권에서의 압축 효율을 개선할 수 있음을 시사한다 \citep{tokenizerimpact,subwordcurse}.

We first examined whether the proposed tokenizer candidates achieve compression performance comparable to that of commercial models (e.g., LLaMA, GPT-4). Figure~\ref{fig:compression_all} summarizes the compression results for both our candidates and the commercial tokenizers. As expected, byte-level BPE–based tokenizers from the LLaMA family exhibit strong performance in English \citep{llama3,radford2019gpt2}, but their effectiveness in Korean is relatively limited.
In contrast, the proposed tokenizer candidates maintained performance on par with commercial models in English domains, while demonstrating a substantial advantage in Korean. This finding suggests that \emph{bilingual} mixture and domain adjustment can enhance compression efficiency for non-Latin scripts \citep{tokenizerimpact,subwordcurse}.
% 또한 vocabulary size가 compression에 미치는 영향을 점검했다. 동일한 학습 설정에서 vocabulary를 확장하면 전반적으로 약 2–7\% 수준의 compression 개선이 관찰되었다. 이는 어휘 확대가 압축 측면의 이득을 줄 수 있으나, 모델 파라미터/메모리·서빙 지연 등과의 \emph{trade-off}가 존재함을 의미한다 \citep{sentencepiece,tokenizerimpact}.
We also examined the effect of vocabulary size on compression. Under identical training configurations, expanding the vocabulary generally improved compression by approximately 2–7\%. This implies that while a larger vocabulary can yield compression benefits, it also introduces trade-offs related to model parameters and memory usage\citep{sentencepiece,tokenizerimpact}. % and serving latency 
% 앞서 언급했듯 compression은 토크나이저의 \emph{intrinsic} 지표 중 하나일 뿐이며, 실제 LLM 학습 후의 다운스트림 성능을 함께 봐야 한다 \citep{tokenizerimpact}. 이에 각 토크나이저로 한국어(FineWeb-2) 9B, 영어(UltraFineWeb) 51B 토큰을 학습한 모델을 다양한 다운스트림 과제로 평가했다. 토크나이저마다 분절 기준이 달라 동일한 “60B 토큰” 기준을 맞추기 위해 카운팅은 단일 기준 토크나이저(GPT-4 토크나이저)를 사용하였다 \citep{radford2019gpt2}.
As noted earlier, compression is merely one of the intrinsic metrics for tokenizer quality. Thus, downstream model performance must also be evaluated \citep{tokenizerimpact}. Accordingly, we trained models using each tokenizer on 9B Korean tokens (FineWeb2) and 51B English tokens (UltraFineWeb) and assessed their performance across a range of downstream tasks. Since tokenization granularity differs across tokenizers, token counts were normalized using a single reference tokenizer (GPT-4 tokenizer) to ensure comparability at the “60B-token” scale \citep{radford2019gpt2}.

\subsection{Downstream Model Performance with Diverse Tokenizers}
\begin{table}[h!]
\begin{adjustbox}{width=\textwidth}
\centering
\begin{tabular}{lcccccccccc}
\toprule
Model & arc\_e & boolq & hswg & obqa & piqa & race & tfqa\_mc1 & tfqa\_mc2 & wgrd & AVG \\
\midrule
LlamaTok & 66.79 & 51.56 & 37.27 & \textbf{24.80} & 69.48 & 31.58 & 20.69 & 35.65 & 50.75 & 43.17 \\
GPT Tok & 67.34 & 47.98 & 37.53 & 23.80 & 70.35 & 30.53 & 19.46 & 35.66 & 53.75 & 42.93 \\
Our EK 125k Tok & \underline{67.63} & \textbf{58.93} & \textbf{38.42} & 22.80 & \underline{71.06} & \textbf{32.92} & \underline{21.54} & 36.49 & \textbf{54.62} & \textbf{44.93} \\
Our EK 125k with Crawl Tok & 67.30 & 48.99 & \underline{37.94} & 22.20 & 70.84 & \underline{31.67} & 21.18 & \textbf{38.99} & 51.85 & 43.44 \\
Our EPK 125k Tok & \textbf{67.68} & \underline{57.06} & 37.75 & \underline{24.40} & \textbf{71.49} & 31.29 & \textbf{23.26} & \underline{37.36} & \underline{53.91} & \underline{44.91} \\
\dashedmidrule
Our EK 196k Tok & 67.55 & 48.38 & 37.32 & 23.20 & 70.95 & 31.10 & 20.81 & 38.71 & 52.64 & 43.41 \\
Our EPK 196k Tok & 68.39 & 46.57 & 38.00 & 24.40 & 70.89 & 32.54 & 21.30 & 38.23 & 52.88 & 43.69 \\
Our EK 196k with Crawl Tok & 67.97 & 55.87 & 37.89 & 25.40 & 70.18 & 31.87 & 22.28 & 36.90 & 52.64 & 44.56 \\
\bottomrule
\end{tabular}
\end{adjustbox}
\caption{Comparison of English downstream model performance across different tokenizers.}
\label{tab:tok-downstream-en}
\end{table}

\begin{table}[h!]
\begin{adjustbox}{width=\textwidth}
\centering
\begin{tabular}{lccccccccc}
\toprule
\multirow{2}{*}{Model} & \multirow{2}{*}{csatqa} & \multirow{2}{*}{haerae} & \multirow{2}{*}{kmmlu} & \multicolumn{5}{c}{KoBEST} & \multirow{2}{*}{AVG} \\
\cmidrule(lr){5-9}
 &  &  &  & boolq & copa & hellaswag & sentineg & wic &  \\
\midrule
Llama Tok & 17.65 & \underline{20.17} & 13.80 & \textbf{50.71} & \underline{54.90} & \textbf{35.00} & 51.39 & \underline{51.03} & 36.86 \\
GPT Tok & 16.04 & 19.25 & 27.24 & 48.93 & 53.50 & 33.00 & 54.16 & \textbf{51.91} & 38.03 \\
Our EK 125k Tok & \underline{24.06} & \textbf{21.08} & 24.24 & 49.72 & 53.60 & 31.00 & 55.92 & 50.64 & 38.83 \\
Our EK 125k with Crawl Tok & 19.79 & 18.42 & \underline{27.38} & 49.29 & \textbf{56.40} & \underline{34.00} & \underline{56.93} & 49.52 & \underline{38.97} \\
Our EPK 125k Tok & \textbf{24.60} & \underline{20.17} & \textbf{28.46} & \underline{50.21} & 53.50 & 31.00 & \textbf{57.18} & 49.84 & \textbf{39.42} \\
\dashedmidrule
Our EK 196k Tok & 16.58 & 18.24 & 19.57 & 49.86 & 55.60 & 30.00 & 56.42 & 51.11 & 37.22 \\
Our EPK 196k Tok & 19.79 & 21.72 & 21.94 & 51.43 & 56.60 & 33.00 & 47.10 & 50.87 & 37.81 \\
Our EK 196k with Crawl Tok & 17.65 & 19.80 & 24.91 & 50.50 & 56.30 & 34.00 & 56.42 & 50.79 & 38.67 \\
\bottomrule
\end{tabular}
\end{adjustbox}
\caption{Comparison of Korean downstream model performance across different tokenizers. All evaluations were conducted under a 3-shot setting.}
\label{tab:tok-downstream-kr}
\end{table}

% Table~\ref{tab:tok-downstream-en}, \ref{tab:tok-downstream-kr}는 토크나이저별로 학습한 모델의 영어/한국어 다운스트림 성능을 요약한다. 영어에서는 \textbf{EK 125k}가 가장 좋은 성능을, \textbf{EPK 125k}가 그 다음을 기록했다. 주목할 점은 영어 compression에서 상용 토크나이저가 우월함에도, 제안 토크나이저로 학습한 모델이 전반적으로 더 높은 다운스트림 성능을 보였다는 것이다. 이는 “압축 효율 ↔ 다운스트림 일반화”가 단순한 단조 관계가 아님을 시사하며 \citep{tokenizerimpact}, 토크나이저 설계가 \emph{데이터 혼합/도메인}과 함께 최적화되어야 함을 뒷받침한다. 한국어에서도 제안 토크나이저가 상위권을 기록했으며, 특히 \textbf{EPK 125k}가 가장 우수했다. 
The downstream performance results for both English and Korean are summarized as follows. In English, \textbf{EK 125K} achieved the best overall performance, followed by \textbf{EPK 125K}. Notably, despite commercial tokenizers exhibiting superior compression in English, models trained with our proposed tokenizers consistently achieved higher downstream performance. This finding suggests that the relationship between compression efficiency and downstream generalization is non-monotonic \citep{tokenizerimpact}, highlighting that tokenizer design should be jointly optimized with the data mixture and the model’s downstream performance rather than treating them as independent factors.

In Korean, the proposed tokenizers also ranked among the top performers, with \textbf{EPK 125K} demonstrating the highest results.

% 한편 vocabulary 확장은 compression 측면에서는 일관된 이득이 있었으나, 영어 다운스트림에서는(일부 EK-with-Crawl 예외) 평균 성능이 오히려 하락하는 경향이 관찰되었다. 이 결과를 바탕으로 우리는 \textbf{196k} 토크나이저를 최종 후보에서 제외하기로 결정하였다 \citep{tokenizerimpact}.
Meanwhile, vocabulary expansion yielded consistent gains in compression but tended to reduce average downstream performance in English (with a few exceptions such as EK-with-Crawl). Based on these observations, we excluded the \textbf{196K}-vocabulary tokenizers from the final set of candidates \citep{tokenizerimpact}.

\subsection{Safety of KORMo Tokenizers}
\begin{figure}[h]
    \centering
    \makebox[\textwidth][c]{%
        \includegraphics[width=0.9\textwidth]{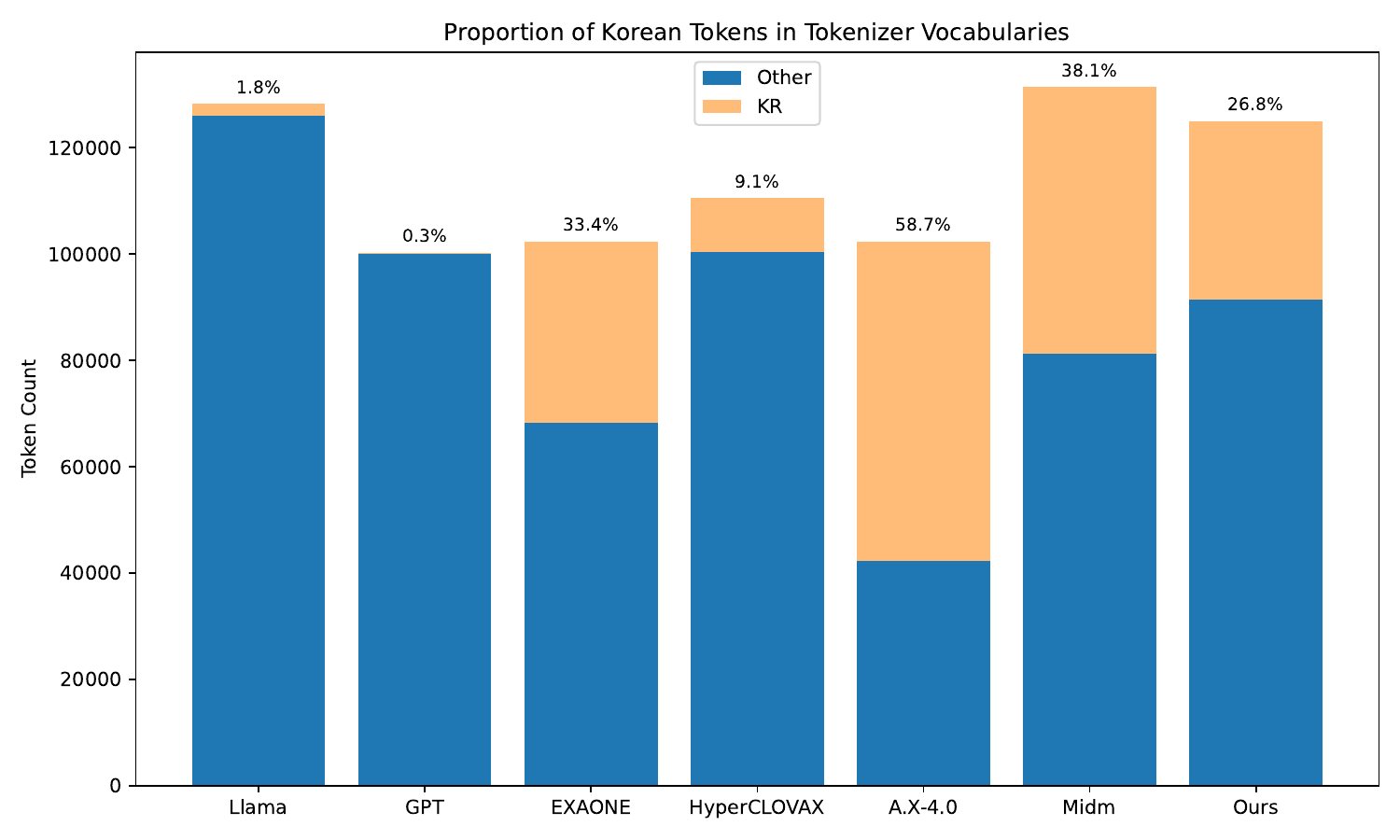}
    }
    \caption{Proportion of Korean tokens within tokenizer vocabularies across different models. Each bar represents the share of Korean (\textit{KR}) versus non-Korean (\textit{Other}) tokens. While English-centric models such as LLaMA and GPT exhibit minimal Korean coverage (1.8\% and 0.3\%, respectively), Korean-specialized models (Exaone4, HyperCLOVAX, A.X-4.0, Midm and ours) show significantly higher proportions.}
    \label{fig:proportion_of_korean}
\end{figure}

\begin{table*}[t!]
\centering
\scriptsize
\renewcommand{\arraystretch}{1.2}
\begin{tabularx}{\textwidth}{lX}
\toprule
    \rowcolor{gray!15}
    \textbf{Tokenizer} & \multicolumn{1}{c}{\textbf{Trained Token with Bias}} \\
    \midrule
    Llama & 출장안마, 마사지, 출장마사지, 콜걸 \\
    \midrule
    EXAONE & 새끼, 마사지, 시발, 몰카, 새끼야, 씨발, 개새끼, 야한, 지랄, 카지노, 시발, 토토, 젖꼭지, 좆같, 자위, 섹스, 좇, 추천인 \\
    \midrule
    HyperCLOVAX-SEED & 놀이터토토, 역출장샵추천, 출장만남, 동출장만남, 역출장, 역출장만남, 동출장맛사지후기, 사설놀이터, 토토사이트, 동출장샵, 토토, 보지, 안마후기, 출장맛사지, 출장마사지, 동출장, 출장샵추천, 출장대행, 동출장안마, 바카라, 먹튀, 동출장대행, 마사지, 카지노, 동출장마사지, 동콜걸출장마사지, 동콜걸추천, 면출장대행, 면출장만남, 면출장마사지, 영화무료보기어플, 역출장마사지, 면출장샵추천, 역출장대행, 출장안마, 동출장아가씨, 콜걸추천, 콜걸출장마사지, 동출장샵추천, 동출장만남후기\\
    \midrule
    A.X-4.0 & 토토, 자위, 좇, 병신, 카지노, 애미, 자지, 섹스, 시발, 새끼\\
    \midrule
    Midm & 모텔, 토토, 보지, 자위, 좇, 몰카, 카지노, 애미, 자지, 섹스, 시발, 새끼 \\
    \midrule
    EK-Crawl & 안출장안마, 출장미인아가씨, 예약금없는출장샵, 흥출장안마, 토토사이트, 온라인카지노, 출장업계, 토토, 양출장안마, 성출장안마, 출장샵, 코인카지노, 먹튀, 출장걸, 콜걸업소, 출장업계위, 출장만남, 지역출장마사지샵, 호텔카지노, 출장, 더킹카지노, 출장소이스홍성, 출장만족보장, 출장가격, 출장코스가격, 출장업소, 출장샵예약, 출장만족, 콜걸샵, 카지노사이트, 우리카지노, 에티오피, 출장마사지, 출장최고, 캐츠비카지노, 모텔출장, 출장부르는법, 타이마사지, 출장외국인, 출장샵안내, 출장연애인급, 콜걸후기, 출장샵콜걸, 콜걸출장마사지, 카지노하는곳, 역출장안마, 솔레어카지노, 미시출장안마, 출장서비스, 출장소, 출장최강미녀, 바카라사이트, 보지, 시출장샵, 콜걸출장안마, 출장소이스, 콜걸, 예스카지노, 콜걸강추, 바카라하는곳, 콜걸만남, 카지노, 모텔출장마사지샵, 출장샵강추, 마사지황형, 출장샵추천, 출장최고시, 주출장안마, 출장서비스보장, 츠비카지노, 콜걸추천, 외국인출장만남, 릉콜걸샵, 출장샵예약포항, 출장아가씨, 바카라, 출장색시미녀언니, 출장오쓰피걸, 출장몸매최고, 오피걸, 레어카지노, 마사지, 천출장안마, 킹카지노, 출장여대생, 출장샵후기, 동출장마사지, 새끼, 출장오피, 산출장안마, 전지역출장마사지샵, 출장마사지샵, 출장안마, 출장전화번호, 출장맛사지 \\
    \midrule
    EK & 보지, 카지노, 젖, 새끼 \\
    \midrule
    EPK & 보지, 카지노, 젖, 새끼 \\
\bottomrule
\end{tabularx}
\caption{Examples of biased or potentially harmful tokens identified in Korean-specialized tokenizers.}
\label{tab:tokenizer-bias}
\end{table*}
% 본 절에서는 후보 토크나이저 vocab 내 잠재적 유해(독성·편향) 토큰을 점검하고, 한국어 특화 공개 모델(EXAONE, HyperCLOVA X, A.X, Midm)과 비교하였다. 유해성 평가는 대표적 독성 기준/리스트에 의거하여 진행했다(예: RealToxicityPrompts, HateXplain, HateCheck) \citep{gehman2020realtoxicity,hatexplain,hatecheck}. 
This section examines the presence of potentially harmful tokens (e.g., toxic or biased expressions) within the vocabularies of the proposed tokenizer candidates and compares them with Korean-specialized commercial models (Exaone, HyperCLOVA X, A.X, and Midm). The harmfulness evaluation was conducted based on established toxicity and bias benchmarks such as RealToxicityPrompts, HateXplain, and HateCheck \citep{gehman2020realtoxicity,hatexplain,hatecheck}.

% Table~\ref{tab:tokenizer-bias}는 각 토크나이저의 vocabulary 중 유해 가능 토큰을 요약한 것이다. 다수의 한국어 특화 모델에서도 일정 수준의 유해 토큰이 존재했으며, HyperCLOVA X 계열에서 상대적으로 높은 비중이 관찰되었다. 반면 영어 중심 모델(Llama, GPT)의 vocab에서는 한국어 비율이 매우 낮아(우리의 집계 기준) 한국어 유해 토큰 검출이 제한적이었다(그 자체로 한국어 표현력 한계를 시사). Figure~\ref{fig:proportion_of_korean}에서도 두 모델의 한국어 비중이 매우 낮음을 확인할 수 있다. 
Table~\ref{tab:tokenizer-bias} summarizes the harmful or potentially sensitive tokens identified in each tokenizer’s vocabulary. We found that several Korean-specialized models also contained a non-negligible number of harmful tokens, with the HyperCLOVA X family exhibiting relatively higher proportions. In contrast, English-centric models such as LLaMA and GPT showed minimal Korean coverage in their vocabularies under our counting criteria, thereby limiting the detection of harmful Korean tokens—an observation that itself suggests constraints in their expressive capacity for Korean. Figure~\ref{fig:proportion_of_korean} further confirms the very low proportion of Korean tokens in these models.

% 종합하면, 제안 후보(EK-Crawl, EPK)는 비교적 적은 수의 유해 토큰을 포함했으나, \emph{Crawl} 데이터를 혼합하는 경우 유해 토큰 수가 증가하는 경향이 있었다. 데이터 소스의 특성이 토크나이저 학습의 편향 형성에 영향을 미칠 수 있음을 시사하며, 본 연구는 이러한 우려에 따라 토크나이저 학습에서 Crawl 데이터를 배제하였다 \citep{fineweb_2024,ultrafineweb_2025}.
Overall, the proposed candidates (EK and EPK) contained comparatively fewer harmful tokens. However, the inclusion of \emph{crawled} data tended to increase the number of such tokens. This finding indicates that data source characteristics can influence the formation of bias during tokenizer training. In light of this concern, we excluded crawled data from the final tokenizer training configuration \citep{fineweb_2024,ultrafineweb_2025}. Considering all the aforementioned factors, \textbf{EPK-125K} was selected as the final tokenizer.

\section{Pretraining Phase}
% 이번 장에서는 우리가 제안하는 사전학습 데이터의 수집 및 생성 방식, 데이터의 퀄리티와 난이도를 고려한 학습 stage 구성 방법에 대해 소개한다.
This section introduces our proposed approach for collecting and generating pre-training data, as well as the methodology for constructing training stages that account for data quality and difficulty.

%우리는 다음과 같은 관점에서 사전학습 데이터를 준비하였다:
%\begin{enumerate}
%   \item 영어 지식을 한국어로 전이할 수 있는 효율적인 증강데이터의 구축방법
%    \item 사전학습에 증강 및 synthetic 데이터를 활용해도 괜찮을까? 혹은 얼마나 사용해도 될까?
%\end{enumerate}

\subsection{Pretraining Datasets}\label{sec:pretrain_dataset}
\begin{table}[h!]
\centering
\begin{adjustbox}{max width=\textwidth}
\begin{tabular}{llrrccccc}
\toprule
\rowcolor{gray!15}
\textbf{Language} & \textbf{Dataset Name} & \textbf{\# tokens} & \textbf{\# origin tok} & \textbf{Reasoning} & \textbf{Synthetic} & \textbf{Synthesizer} & \textbf{Seed} &\textbf{Stage}\\
\midrule
\multirow{7}{*}{English}
& DCLM\tablefootnote{\url{github.com/mlfoundations/dclm}} & 1,000B    & 6,000B & X & X& - & -  & stage 1\\
& UltraFineWeb\tablefootnote{\url{huggingface.co/datasets/openbmb/Ultra-FineWeb}}  & 793B      & 1,000B & X & X & - & - & stage 2\\
& Nemotron-CC-web\tablefootnote{\url{data.commoncrawl.org/contrib/Nemotron/Nemotron-CC/index.html}} & 280B      & 4,400B & X & X & - & - & stage 2\\
& Nemotron-CC-synthetic                                                         & 1,000B    & 1,900B & X & O & Mistral-Nemo-12B-Instruct & Nemotron-CC-HQ & stage 2\\
& stack-edu\tablefootnote{\url{huggingface.co/datasets/HuggingFaceTB/stack-edu}} & 152B      & --     & X & X & - & - & stage 2 \\
& Fine-math\tablefootnote{\url{huggingface.co/datasets/HuggingFaceTB/finemath}} & 37.3B     & --     & X & X & - & - & stage 2\\
& Cosmopedia\tablefootnote{\url{huggingface.co/datasets/HuggingFaceTB/finemath}} & 25B       & --     & X & O & Mixtral-8x7B-Instruct-v0.1 & Cosmopedia seed suite\tablefootnote{Web-text, stanford.edu, UltraChat, OpenHermes2.5, OpenStax, KhanAcademy, AutoMathText} & stage 2\\
& OpenCodeReasoning\tablefootnote{\url{huggingface.co/datasets/nvidia/OpenCodeReasoning}} & 5.46B     & --     & O & O & DeepSeek-R1 & - & stage 2\\
& OpenMathReasoning\tablefootnote{\url{huggingface.co/datasets/nvidia/OpenMathReasoning}} & 24.89B    & --     & O & O & DeepSeek-R1, QwQ-32B & - & stage 2\\
\midrule
\multirow{7}{*}{Korean}
& Ko-Web Datasets & 36.3B     & 837B   & X & X & -  & -  & stage 1 \\
& Ko-CC-Dump                                                                       & 6.2B      & 251B   & X & X & -  & -  & stage 1 \\
& Korean Opensource                                                             & 5.57B     & --     & X & X & -  & -  & stage 2 \\
& \underline{Synth-FineWeb2}                               & 10.97B    & --     & X & O & Qwen3-30B-A3B & FineWeb2\tablefootnote{\url{huggingface.co/datasets/HuggingFaceFW/fineweb-2}} & stage 2\\ 
& \underline{Synth-Nemo-HQ}                                      & 32.82B    & --     & X & O & Qwen3-30B-A3B & Nemotron-HQ & stage 2\\
& \underline{Kosmopedia}                                   & 4.07B     & --     & X & O & GPT-oss (120B) & cosmopedia & stage 2\\
& \underline{Synth-Ultrafineweb}                              & 41.69B    & --     & X & O & GPT-oss (120B) & Ultrafineweb & stage 2\\
%& \underline{Kosmopedia2}\footnote{} & 45B & -- & X & O & GPT-oss (120B) & cosmopedia\\
& \underline{Ko-Reasoning}                                          & 7.05B & -- & O & O & Qwen3-235B-A22B & Nemotron-Post  & stage 2\\
\midrule
\multicolumn{9}{c}{\textbf{English + Korean total pretraining tokens: 3,462.32B}}  \\ 
\bottomrule
\end{tabular}
\end{adjustbox}
% \caption{영어/한국어 데이터셋 구성 비교.`tokens'는 실제 학습 토큰 수를 의미하며 `origin tok'는 필터링 하기 전 기존 토큰 수를 의미한다. `Reasoning'은 Reasoning trace가 포함된 dataset인지 아닌지, `Synthetic'은 synthetic data인지 아닌지를 의미한다. underline되어 있는 데이터는 직접 생성한 generated synthetic dataset을 의미한다. `Synthesizer'는 데이터를 합성할 때 활용한 언어 모델, `Seed'는 데이터 합성을 위한 시드 데이터, `Stage'는 pretraining에서 어느 stage에서 해당 데이터를 학습하였는지를 의미한다.}
\caption{Comparison of English and Korean pretraining datasets used in KORMo. Each dataset is categorized by reasoning trace inclusion, synthetic nature, synthesizer model, and training stage. The underlined datasets indicate generated synthetic data produced in-house.}
\label{tab:pretrining-datasets-main}
\end{table}

% 사전학습은 언어모델 구축 과정에서 가장 많은 자원과 노력이 소요되는 단계이다. 우리는 KORMo의 사전학습을 위해 대규모의 고품질 텍스트 데이터를 수집하거나 생성하였다. Table~\ref{tab:pretrining-datasets-main}은 주요 사전학습 데이터셋의 구성을 요약한다. 데이터는 원천 수집(crawling 및 공개 데이터셋 활용)과 synthetic generation을 통해 확보되었으며, 일반 웹 텍스트, 도메인 특화 자료, 그리고 다양한 보조 데이터셋을 포함한다.  
Pre-training is the most resource-intensive and labor-demanding stage in building a language model. For KORMo’s pre-training, we collected and generated large-scale, high-quality text corpora. Table~\ref{tab:pretrining-datasets-main} summarizes the composition of the primary pre-training datasets. The data were obtained through both direct collection (web crawling and the use of publicly available datasets) and synthetic generation, encompassing general web text, domain-specific materials, and various auxiliary datasets.

\subsubsection{Public Data Collection}

\paragraph{English Public Data.}  
% 우리는 먼저 공개적으로 제공되는 고품질 데이터셋을 수집하여 KORMo의 사전학습에 활용하였다. Table~\ref{tab:pretrining-datasets-main}에 정리된 바와 같이, 영어 데이터는 모두 공개 데이터셋에서 수집한 뒤 추가 필터링 과정을 거쳐 사용하였다.
We first collected publicly available high-quality datasets and utilized them for KORMo's pre-training. As summarized in Table~\ref{tab:pretrining-datasets-main}, all English data were sourced from open datasets and further refined through additional filtering processes before use.
% 구체적으로, DCLM~\citep{NEURIPS2024_19e4ea30}, UltraFineWeb~\citep{wang2025ultra}, Nemotron-CC~\citep{su2024nemotron}에서 제공되는 웹 데이터는 Common Crawl을 기반으로 다단계 필터링 과정을 거쳐 구축된 고품질 코퍼스로, 일반 언어 이해를 위한 핵심 기반으로 활용되었다. 또한 Nemotron-synthetic은 Nemotron-web으로부터 생성된 synthetic 데이터셋이며, Cosmopedia는 RefinedWeb~\citep{penedo2023refinedweb}과 RedPajama~\citep{weber2024redpajama}를 시드 데이터로 활용하여 구축된 synthetic 데이터셋으로, 대규모 언어모델의 일반화 성능 강화를 목적으로 포함하였다.
Specifically, the web corpora provided by DCLM~\citep{NEURIPS2024_19e4ea30}, UltraFineWeb~\citep{wang2025ultra}, and Nemotron-CC~\citep{su2024nemotron} are high-quality datasets constructed through multi-stage filtering based on Common Crawl, serving as the foundation for general language understanding. In addition, Nemotron-synthetic is a synthetic dataset generated from Nemotron-web, while Cosmopedia is a synthetic corpus built using RefinedWeb~\citep{penedo2023refinedweb} and RedPajama~\citep{weber2024redpajama} as seed data, included to enhance the generalization performance of large-scale language models.
% 추가적으로, 수학·코딩 영역의 기초 학습을 위해 Stack-Edu 및 Fine-Math 데이터셋을 활용하였으며, 고차원적 추론 능력을 강화하기 위해 OpenCodeReasoning과 OpenMathReasoning 데이터셋을 함께 수집하였다. 이와 같은 다층적 데이터 구성을 통해 KORMo는 다양한 영역의 지식을 학습할 수 있도록 설계되었다.  
Additionally, for foundational training in mathematics and coding, we utilized the Stack-Edu and Fine-Math datasets, while the OpenCodeReasoning and OpenMathReasoning datasets were incorporated to further enhance higher-order reasoning abilities. Through this multi-layered data composition, KORMo was designed to acquire diverse forms of knowledge across multiple domains.

\paragraph{Korean Public Data.}  
% 한국어 데이터의 경우 공개적으로 활용 가능한 사전학습 데이터가 극도로 제한적이기 때문에, 우리는 기존에 공개된 리소스와 Common Crawl의 raw dump를 직접 파싱한 데이터로 구분하여 수집하였다. 
For Korean data, publicly available pre-training resources are extremely limited. Therefore, we collected data from two primary sources: existing open resources and raw dumps directly parsed from Common Crawl.

\begin{itemize}
    \item \textbf{Korean Opensource:} We collected writing, news, and document summarization data from the National Institute of Korean Language’s Modu Corpus\footnote{\url{https://kli.korean.go.kr/corpus/main/requestMain.do?lang=ko}}, and included the academic paper dataset provided by KISTI\footnote{\url{https://aida.kisti.re.kr/data/}}. These corpora are released by reputable Korean research institutions, ensuring both reliability and quality.

    \item \textbf{Ko-Web Datasets:} We utilized Korean datasets released by international research communities, including community-OSCAR\footnote{\url{https://huggingface.co/datasets/oscar-corpus/community-oscar}}, CulturaX\footnote{\url{https://huggingface.co/datasets/uonlp/CulturaX}}, and FineWeb2 v2.0.0\footnote{\url{https://huggingface.co/datasets/HuggingFaceFW/fineweb-2}}. However, these datasets share a limitation in that their knowledge cutoff extends only up to 2023.

    \item \textbf{Ko-CC-Dump:} To ensure data freshness, we directly extracted Korean text from the raw dumps of Common Crawl. Specifically, we parsed WARC files from four dumps, 2025-13, 2025-18, 2025-20, and 2025-21, corresponding to web crawl data collected between March 15 and May 25, 2025. This approach follows the findings reported in DCLM, which demonstrated that directly parsing WARC files yields higher-quality data than relying on preprocessed WET files. For data extraction, we used the \texttt{datatrove} library and employed \texttt{resiliparse} to ensure efficient and stable processing.

    The language identification provided by Common Crawl (based on CLD2) exhibited low accuracy; therefore, we implemented a new language filtering process using fastText’s LID-176 model. The classification threshold was set to 0.8 to minimize noise while maintaining high recall for Korean text. Although only four subsets were utilized in this study due to time constraints, all additional subsets will be released in the future to support further research\footnote{\url{https://huggingface.co/datasets/MLP-KTLim/Kor-CC-Resili-Parsed}}.
\end{itemize}

\begin{comment}
    국립국어원의 모두의 말뭉치\footnote{\url{https://kli.korean.go.kr/corpus/main/requestMain.do?lang=ko}}에서 글쓰기, 뉴스, 문서 요약 데이터를 수집하였고, KISTI의 학술 논문 데이터셋\footnote{\url{https://aida.kisti.re.kr/data/}}을 포함하였다. 이들은 공신력 있는 한국 연구기관에서 공개한 데이터로, 신뢰성과 품질이 보장된 리소스이다.
    
    해외 연구 커뮤니티에서 공개한 한국어 데이터셋으로 community-OSCAR\footnote{\url{https://huggingface.co/datasets/oscar-corpus/community-oscar}}, CulturaX\footnote{\url{https://huggingface.co/datasets/uonlp/CulturaX}}, FineWeb2 v2.0.0\footnote{\url{https://huggingface.co/datasets/HuggingFaceFW/fineweb-2}}를 활용하였다. 다만 이들 데이터셋은 knowledge cutoff가 2023년까지라는 한계가 존재한다.

    최신성을 확보하기 위해 Common Crawl의 raw dump로부터 직접 한국어 데이터를 추출하였다. 구체적으로, 2025년 3월 15일부터 5월 25일까지의 웹 크롤링 데이터 중 네 개 subset (2025-13, 2025-18, 2025-20, 2025-21)을 대상으로 WARC 파일을 직접 파싱하여 수집하였다. 이는 DCLM에서 보고된 바와 같이 WET 파일보다 WARC 파일을 직접 파싱할 경우 더 높은 품질의 데이터를 확보할 수 있다는 결과를 반영한 것이다. 데이터 추출에는 \texttt{datatrove} 라이브러리를 사용하였으며, 효율적인 처리와 안정성을 위해 \texttt{resiliparse}를 활용하였다.

    Common Crawl에서 제공하는 언어 식별(CL2D 기반)은 정확도가 낮아, 우리는 fastText의 LID-176 모델을 활용하여 새로운 언어 판별 과정을 수행하였다. Threshold는 0.8로 설정하여 잡음은 최소화하면서 한국어 텍스트의 recall을 높였다. 본 학습에서는 시간적 제약으로 인해 네 개 subset만을 활용했지만, 추 후 연구자들을 위해 추가 subset 또한 모두 공개한다
\end{comment}

% \subsubsection{증강데이터 생성 (민경, 민준도움)}
\subsubsection{Synthetic Data Generation}
% 우리 라이선스 프리 한국어 데이터 없자나 그래서 만들어야 하는데, 시드 데이터 조차 양질이면서 라이센스 프리 없는거야. 그래서 제안하는게 Knowledge transfer 가능한 사전학습 데이터야. 엄청 간단하지만 매우 실용적이야. 영어 지식을 한국어로 옴기는거야!! (자세한설명 추가) 이렇게 증강데이터 만들었어 엄청 많이 이거는 다양성이랑 언어 벨런스 고려해서 했어. Appendix봐바 실제 데이터 양질이고 엄청 다양해 편향되지 않는다구.

% 비영어권 언어, 특히 한국어의 경우 라이선스 제약 없는 공개 사전학습 데이터가 거의 전무하다. 이는 대규모 언어모델을 완전히 공개적(fully open source)으로 구축하려는 시도를 구조적으로 제한한다. 따라서 우리는 단순히 데이터의 양을 확보하는 수준을 넘어, 영어 중심 데이터의 지식을 한국어로 효과적으로 전이(knowledge transfer)하는 방식을 실험적으로 검토하였다.

For non-English languages, particularly Korean, publicly available pretraining data with permissive licenses remains extremely scarce. This poses a fundamental limitation for building fully open-source LLMs. To address this challenge, we do not simply aim to increase the volume of training data. Instead, we empirically investigate methods for effectively transferring knowledge from English-centric corpora into Korean through controlled augmentation strategies.

% FineWeb2와 Nemo-HQ에서 seed query를 가져와서 gpt api로 번역 작업을 진행하고, 번역된 query를 Qwen3-30B-A3B에 입력해 한국어 Non-reasoning 데이터를 생성하였다. 이 때의 prompt는 nemotron-cc의 subset 별 (QA, Distill, knowledge list, knowledge extract, wiki) 영어 프롬프트를 번역/검수해 사용하였다. 결과적으로 FineWeb2와 Nemo-HQ 각각 11B, 32B의 토큰을 생성했다.
% prompts: https://www.notion.so/Kormo-synthetic-2465ecb57e5780fdb853c7cb730d8ce4

% 최근 Nemotron-CC-HQ에서 제안된 접근처럼, 제한된 시드 데이터를 질의응답(QA) 변환, 문장 리프레이징, 위키스타일 재구성 등 다양한 방식으로 증강하는 전략은 높은 실용성을 입증하였다. 본 연구에서는 이러한 아이디어를 확장하여, 양질의 영어 시드 데이터를 입력으로 삼아 한국어 데이터를 대량 생성하였다. 단, 영어 문화적 맥락이나 생활양식이 그대로 전이될 경우 한국어 모델의 활용성에 부정적 영향을 줄 수 있음을 고려하였다. 이를 최소화하기 위해 프롬프트 설계 단계에서 한국 현실과 맥락을 반영하도록 지침을 추가하였으며, 동일 영어 시드라도 유형·서술 방식·콘텐츠가 다양화되도록 다수의 시드 풀(seed pool)을 구축하였다. 본 연구에서는 다섯 가지 한국어 증강 데이터셋을 구축하였으며, 각각은 상이한 영어 시드 데이터와 프롬프트 설계를 기반으로 한다.

Recent work, such as Nemotron-CC-HQ, has demonstrated the practical utility of augmenting limited seed data through various transformations, including QA-style conversions, sentence rephrasings, and Wikipedia-style rewrites. Building on this approach, we generate large-scale Korean data using high-quality English seed corpora as input. However, we note that directly transferring Anglocentric cultural contexts or lifestyle content may limit the downstream applicability of Korean models. To mitigate this, we introduced prompt design constraints that reflect Korean-specific context and usage scenarios. We also constructed multiple seed pools to promote diversity in format, narrative style, and content, even when using the same English input. In total, we constructed five Korean augmentation datasets, each derived from a distinct combination of English seed data and prompt strategy.

% \begin{itemize}
%     \item \textbf{Synth-FineWeb2:} FineWeb2를 시드 데이터로 활용하고, 자체 제작한 한국어 프롬프트를 적용하여 한국어 응답을 생성.
%     \item \textbf{Synth-UltraFineWeb:} Nemotron-CC 기반 영어 프롬프트에 ``한국어로 답변하라''는 지시를 추가하여 한국어 데이터를 생성.
%     \item \textbf{Synth-Nemo-HQ:} Nemotron-CC-HQ를 시드로 하여, 프롬프트를 한국어로 번역한 뒤 한국어 데이터를 구축.
%     \item \textbf{Kosmopedia:} Cosmopedia를 시드로 하고, 자체 설계한 한국어 프롬프트를 활용하여 데이터 생성.
%     \item \textbf{Ko-Reasoning:} Nemotron-Post 데이터의 영어 질의를 활용하고, 한국어 reasoning trace와 답변 세트를 병렬로 구축.
% \end{itemize}

\begin{itemize}
    \item \textbf{Synth-FineWeb2:} Korean responses generated from FineWeb2 seed data using custom-designed prompts.
    \item \textbf{Synth-UltraFineWeb:} Korean outputs generated by appending the instruction “answer in Korean” to Nemotron-CC English prompts.
    \item \textbf{Synth-Nemo-HQ:} Korean data generated by translating Nemotron-CC-HQ prompts into Korean prior to synthesis.
    \item \textbf{Kosmopedia:} Korean data generated from Cosmopedia seed texts using custom-designed Korean prompts.
    \item \textbf{Ko-Reasoning:} Constructed by pairing each English query from Nemotron-Post with a corresponding Korean reasoning trace and answer.
\end{itemize}

% 우리는 다양성을 확보하기 위해 서로 다른 다섯 개의 영어 데이터 소스로부터 증강 데이터를 생성하였다. 이는 시드 데이터의 도메인, 서술 방식, 지식 수준이 서로 상이하기 때문에, 가능한 한 폭넓고 고품질의 지식 자원을 한국어로 전이하기 위함이다. 다만 생성된 증강 데이터는 RQ1. 에서 제안한 안전성 문제가 있을 수 있다. 이를 검증하기 위해 우리는 Nemotron-HQ 연구에서 제안된 방법과 유사하게, 구축한 각 증강 데이터셋의 down stream task에서 효과를 검증하였다. 구체적으로, 제2장에서 최종적으로 선정된 \textit{proxy} 모델에 제안한 토크나이저를 적용하고, 각 증강 데이터셋으로 4B 토큰 학습을 수행하였다. 실험 결과, Kosmopedia 계열 데이터가 가장 우수한 성능을 보였으며, 다른 대부분의 증강 데이터셋 또한 웹 기반 한국어 코퍼스로 학습된 모델보다 높은 성능을 기록하였다. 이러한 결과는 \textbf{RQ1에서 제기된 증강 데이터 활용의 안정성과 효과성}을 실증적으로 뒷받침하며, 사전학습 단계에서 대규모 증강 데이터를 적극적으로 활용하는 것이 비영어권 언어 모델 구축에 유효한 전략임을 시사한다.

To ensure diversity, we generated augmented datasets from five distinct English sources. These sources differ in domain, narrative style, and knowledge density, enabling broad and high-quality knowledge transfer into Korean. However, as noted in RQ1, the use of synthetic data raises concerns about its potential impacts on training stability and long-term model performance. To assess this, we adopted an evaluation methodology similar to that proposed in the Nemotron-HQ study. Specifically, we applied our tokenizer to the proxy model selected in Section 2 and conducted pretraining with 4B tokens from each augmented dataset. We then evaluated the resulting models on downstream tasks. Experimental results show that the Kosmopedia-based data yielded the strongest performance, and most other augmented datasets also outperformed models trained on standard Korean web corpora. These findings provide empirical evidence supporting the robustness and effectiveness of synthetic data, as raised in RQ1, and suggest that large-scale synthetic pretraining is a viable strategy for building open-source LLMs in low-resource languages.

%- seed: Fineweb2 prompt: 자체 제작 한국어 프롬프트\\
%- seed: Nemotron-cc HQ. prompt: Nemotron-cc의 synthesizing 프롬프트를 한국어로 번역\\
%- seed: UltraFineWeb, prompt: Nemotron-cc의 synthesizing 프롬프트(영어 프롬프트)에 한국어로 답변하라는 시그널 추가\\ 

% (현석)
\subsubsection{Data Filtering}
\label{subsec:filtering}

% 자 사전학습 데이터는 다 만들어졌어. 보통 사전학습 데이터를 양질의 데이터만 남기기 위해 필터링을 하지. 때에 따라 양질의 데이터 필터링에 따라 절반만 되는 데이터로도 더 좋은 성능을 낼수 있기에 매우매우 중요해. 우리도 해야하는데 상황이 약간달라 왜냐하면 증강데이터가 많거든. 아직 사전학습 단계에서 증강데이터에 대한 필터링? 혹은 효과를 실험한 경우는 많지않아. 그래서 우리가 했어. 최소한 두 가지를 확인해야 했거든.

% 1. 우리가 적용한 데이터 필터링이 효과적일까? 확인을 꼭해야해
% 2. 과연 증강데이터를 사전학습 단계에서 일반데이터와 섞어 학습할 때 어느정도 비율로 하면 좋을까?

% 이걸위해 생성된 데이터에서 약간을 빼서 학습했어 그랬더니 OLD-ver4가 한국어/영어 양쪽에 모두 유의미한 정도로 좋은 성능을 보였지 뭐야. 그래서 우리 결론은 사전학습 때부터 증강데이터를 50\% 정도 사용해도 충분히 좋은 성능이 나온다 라고 판단했어. 당연히 더 많은 경우를 실험하면 좋겠지만 시간과 자원은 한정적이라 여기서 멈춰야했던걸 이해해줘. 추후 연구에서 할께.

%%%% 현석 작성 %%%%
% 사전학습 데이터는 데이터 사이즈도 가장 클 뿐더러 웹 데이터가 큰 비중을 차지하기 때문에 데이터의 quality control에 따라 성능차이가 매우 심하게 난다. 필터링은 수집된 데이터로 부터 양질의 데이터를 추출하는 방법으로 다양한 필터링 방법들이 적용되고 있다. 그 중 우리는 크게 세 가지 필터링 (Heuristic Filtering, Deduplication and Quality Filtering)을 3단계에 걸쳐 진행하였으며 구체적인 내용은 아래와 같다.
Model performance is highly sensitive to the quality of the pre-training data, which is not only the largest in volume but also predominantly sourced from the web. To address this, various filtering techniques are applied to extract high-quality text from the raw collected data. In our work, we apply a three-stage filtering pipeline: Heuristic Filtering, Deduplication, and Quality Filtering. The specifics of this process are detailed below.

% \textbf{Heuristic Filtering} 우리는 수집한 모든 한국어 웹 데이터들에 대해 Heuristic Filtering을 진행하였다. 휴리스틱 필터링은 규칙 기반으로 텍스트에 존재하는 다양한 noise를 제거하는 방법이다. 예를 들면 웹 text에 크롤링 오류로 인해 길이가 극단적인 샘플들, 특수문자의 과도한 반복 등의 텍스트를 검출해 제거하는것이다. 이러한 데이터는 모델의 학습에 악영향을 끼칠 수 있으므로, 엄격한 pipeline을 통해 제거하였다. 우리는 \citep{NEURIPS2024_19e4ea30}에 기반한 heuristic filtering pipeline을 적용해 데이터를 일차적으로 필터링 했다.

\textbf{Heuristic Filtering} We applied heuristic filtering to all collected Korean web data. Heuristic filtering refers to a rule-based approach for removing various forms of noise present in text. We identify and eliminate texts of extreme length, often caused by crawling errors, and instances of excessive repetition of special characters. Since such data be detrimental to model training, we strictly removed them through a well-defined pipeline. Our heuristic filtering pipeline was primarily based on the method proposed by \citep{NEURIPS2024_19e4ea30}.

% item 쓸까 고민중
%%%% heuristic filtering detail %%%%

\begin{enumerate}
    % \item 초기 전처리 - 서로 다른 샘플에 대해 동일한 핕터링을 적용하기 위해, 정규화를 진행한다. 연속된 스페이스, tab을 단일 공백으로, 3개 이상의 개행 문자를 2개의 개행 문자로, CR/CRLF 포함 2개 이상의 연속 개행 문자를 2개의 개행 문자로 치환하였다. 해당 과정 이후 문자열이 비어있다면 drop했다.
    \item \textbf{Initial Preprocessing} –  We first performed text normalization to ensure consistent processing across all samples. Consecutive spaces or tabs were reduced to a single space, sequences of three or more newline characters were replaced with two newlines, and any CR/CRLF sequences exceeding two were also normalized to two newlines. Samples that became empty after normalization were discarded.
    
    % \item word 길이 필터 - 크롤링 오류 등으로 내용이 짧거나 긴 샘플을 word 단위로 판별하여 필터링하였다. word는 실행 시간을 고려하여 space로 구분하였으며, 최소 단어를 10, 최대 단어를 10000 단어로 제한하여 필터링했다.
    \item \textbf{Word-Count Filter} – To remove documents with extreme lengths, which typically result from crawling errors, we applied a word-count filter. For computational efficiency, words were defined as space-delimited tokens. Documents with fewer than 10 or more than 10,000 words were discarded.
  
    % \item 문자(alpha) 미포함 단어 비율 필터 - 전체 word에서, 글자(문자)를 하나도 포함하지 않고, 숫자 또는 특수문자로만 이루어진 단어의 비율이 높다면 noise로 판단하고 제거하였다. noise 단어의 비율이 0.25를 넘어가는 샘플을 drop했다.
    \item \textbf{Non-Alphabetic Word Ratio Filter} – We filtered documents containing a high proportion of non-alphabetic words (i.e., tokens composed entirely of digits or symbols). Samples with a ratio of such non-alphabetic words exceeding 0.25 were discarded.

    % \item 문자, 숫자 character 비율 필터 - 3번 필터와 반대로, 전체 text에서 문자, 숫자 등 유의미한 character가 차지하는 비율이 작을 경우 drop하였다. 이는 text에 절대적으로 존재하는 특수문자의 비율이 클 경우 광고성, 구조화된 text일 가능성이 크다고 판단하여 제거하기 위함이다. 해당 cutoff 비율은 0.25로 설정하였다.
    \item \textbf{Alphabetic–Numeric Character Ratio Filter} – Conversely, samples in which the proportion of meaningful characters like letters or digits, within the entire text was too low were removed. Texts dominated by special symbols are likely to represent advertisements or structured markup rather than natural language. The cutoff threshold was set to 0.25.
    
    % \item 심볼 비율 핕터 - 전체 text에 포함된 특정 심볼의 개수가 단어 수 대비 과도하다면 noise sample로 판단하여 drop했다. 이는 특정 template를 따르거나 광고성, 스팸성, 표 파싱 깨짐 샘플 등을 제거하기 위함이다. 이에 해당하는 symbol은 symbols = ["\#", "...", ". . .", "\u2026"] 요거고 전체 텍스트 대비 0.1을 넘기면 drop했다.
    \item \textbf{Symbol Ratio Filter} – Samples containing an excessive number of specific symbols relative to their word count were identified as noise and removed. This filter primarily targets template-like, spam, or table-corrupted text. The monitored symbols were \verb|["#", "...", ". . .", "\u2026"]|, and samples exceeding a ratio of 0.1 were dropped.
    
    % \item n-gram 기반 repetition 필터 - 연속된 n-gram이 문서 내에서 과도하게 반복된다면 drop했다. 반복이 많은 샘플은 모델에게 repitition을 학습시킬 수 있으므로, 해당 필터링으로 제거하였다. n은 8~10으로 적용, 반복 비율이 0.2를 넘으면 drop했다.
    \item \textbf{N-gram Repetition Filter} – Samples exhibiting excessive repetition of contiguous n-grams were removed, as they can induce undesirable repetition patterns during model training. We analyzed contiguous n-grams of lengths 8 to 10 and discarded any document with a repetition ratio exceeding 0.2.
    
    % \item 말줄임표 비율 기반 필터 - 특정 symbol로 끝나는 line의 비율이 높은 경우 drop했다. 웹에는 완전한 문장 구조를 갖지 않는 문장의 비율이 많아, 이를 제거하기 위한 목적이다. ellipsis set = {'...', '. . .', '\u2026'}이고, 이때의 line은 실행 시간을 고려하여 개행 문자를 기반으로 split했다. 이러한 문장이 전체 문장 대비 0.3 비율 넘으면 제거했다.
    \item \textbf{Line-Ellipsis Ratio Filter} – We removed documents where a large fraction of lines terminated with an ellipsis (\verb|{‘...’, ‘. . .’, ‘\u2026’}|). Such incomplete sentences are prevalent in web data and provide low-quality signals for language modeling. Documents exceeding a line-ellipsis ratio of 0.3 were dropped.
    
    % \item bullet 기호 비율 기반 필터 - 불릿 기호로 시작하는 line의 비율이 높을 경우, 단순한 광고글 또는 단순한 목차, index일 확률이 높으므로 이를 걸러내기 위한 필터이다. bullets = ['●', '•', '*', '-']이고 해당 문장 비율이 0.9를 넘어가면 drop했다.
    \item \textbf{Bullet Point Ratio Filter} – Documents where a high proportion of lines began with bullet symbols (\texttt{['●', '•', '*', '-']}) were removed, as they typically represent lists or navigational content rather than natural language prose. The removal threshold for this ratio was set to 0.9.

\end{enumerate}

%%%% heuristic filtering detail %%%%
% \textbf{Deduplication} 기본적으로 모든 학습과정에서 중복되거나 반복되는 데이터를 삭제하는건 매우 중요하다. 이를 위해 우리는 DCLM에서 제안한 Bloom filter 기반의 중복 제거 방식(bff)\footnote{https://github.com/mlfoundations/dclm/tree/main/dedup/bff}을 사용했으며, 모든 하이퍼파라미터 값 또한 원본 연구를 따랐다. bff에는 (document, old-both)두 가지 유형의 중복 제거방법이 존재한다. document 방식은 토큰을 더 많이 보존하는 대신 중복 제거 강도가 상대적으로 약한 반면, old-both는 더 엄격한 기준을 적용한다.
Removing duplicate or repetitive data is a crucial step in all stages of model training. To achieve this, we adopted the Bloom Filter–based deduplication method (BFF)\footnote{\url{https://github.com/mlfoundations/dclm/tree/main/dedup/bff}}
 proposed in DCLM, following the original study’s hyperparameter settings. The BFF framework provides two modes of deduplication: \textit{document} and \textit{old-both}. The \textit{document} mode preserves more tokens but applies a relatively lenient deduplication threshold, whereas \textit{old-both} enforces a stricter criterion for removing redundant samples.

%\begin{itemize}
%    \item \textbf{영어 필터링} 대부분의 웹 텍스트는 Common Crawl을 기반으로 하므로, DCLM, FineWeb2 등 다양한 출처의 데이터셋을 함께 활용할 경우 데이터 간 중복을 제거하는 과정이 필수적이었다. 
%    \item \textbf{한국어 필터링} 해외에서 공개한 한국어 데이터로 community-oscar~\footnote{https://huggingface.co/datasets/oscar-corpus/community-oscar}, CulturaX~\footnote{https://huggingface.co/datasets/uonlp/CulturaX}, FineWeb2~\footnote{https://huggingface.co/datasets/HuggingFaceFW/fineweb-2} v2.0.0을 수집했다. 다만 knowledge cut off가 2023년 까지라는 단점이 있다.
%\end{itemize}

% 한국어 데이터의 경우, FineWeb2, CulturaX, 우리가 직접 크롤링한 CC-Dump, 그리고 community-oscar 순서로 교차 중복 제거(cross-deduplication)를 진행했다. 다만 영어와 달리 한국어는 학습 데이터 양이 매우 적기 때문에 필터링 강도에 대한 의사 결정이 중요했다. 
we performed cross-deduplication sequentially across \textit{FineWeb2}, \textit{CulturaX}, our internally crawled \textit{CC-Dump}, and \textit{Community-OSCAR}. However, unlike for English, the volume of available training data in Korean is significantly more limited, making the decisions regarding the intensity of the filtering process particularly critical.

% Table~\ref{tab:pretraining-mix-korean-acc}의 deduplication 컬럼은 Document 방법과 Old-both방법을 QF (Quality Filtering) 방식과 결합해 실험한 한국어 모델 성능이다. 결과에서 명확히 드러나듯이, 엄격한 `Old-both' 방식으로 처리된 데이터로 학습한 모델이 모든 평가 항목에서 `Document' 방식보다 높은 성능을 보였다. 이는 사전 학습 데이터셋 구축에서 단순히 데이터의 양을 늘리기보다, 중복을 최소화하여 데이터의 질(quality)을 높이는 것이 최종 모델 성능에 더 결정적인 영향을 미친다는 점을 시사한다.
The \textit{Deduplication} column in Table~\ref{tab:pretraining-mix-korean-acc} reports the performance of Korean models trained using the Document and Old-both deduplication methods in combination with Quality Filtering (QF). As clearly shown in the results, models trained on data processed with the stricter Old-both method consistently outperform those trained with the Document method across all evaluation metrics. This suggests that in pre-training dataset construction, enhancing data quality by minimizing redundancy has a more decisive impact on final model performance than simply increasing the quantity of data.

% 하지만 old-both 방식을 적용했을 때, 전체 한국어 데이터의 약 70\%가 중복으로 간주되어 제거되었다. 이처럼 막대한 양의 데이터가 제거된 것은 현재 공개된 오픈소스 한국어 데이터 상당수가 서로 중복되며, 실질적으로 활용 가능한 고유한(unique) 문서가 매우 부족하다는 현실을 보여준다. 우리는 성능 향상 효과가 명확했기 때문에, 최종적으로 old-both 방식의 중복 제거 데이터셋을 채택하였다.
However, applying the Old-both method resulted in the removal of approximately 70\% of the entire Korean corpus, as those samples were identified as duplicates. This substantial reduction reveals that a large portion of publicly available open-source Korean datasets significantly overlap with one another, leaving a limited amount of truly unique and usable content. Despite the reduction, we selected the Old-both–deduplicated dataset as our final version, given its clear and consistent performance gains across all evaluation metrics.

%반면 영어 데이터의 경우, UltraFineWeb, Nemotron-CC, DCLM 순으로 동일한 중복 제거 작업을 수행했을 때 데이터 감소량이 한국어에 비해 현저히 적었다. 이러한 차이가 발생하는 이유는 두 가지로 추정한다. 첫째, 각 데이터셋이 Common Crawl 원본 데이터를 각기 다른 텍스트 추출기(text extractor)로 처리하여, 내용이 같더라도 표면적으로는 다른 데이터로 인식되었을 가능성이다. 둘째, 방대한 영어 데이터의 규모에 비해 우리의 제한된 리소스로는 Bloom filter가 모든 중복을 탐지하기에 충분히 크지 않았을 수 있다.
%영어와 한국어 모두 DCLM에서 요구하는 하이퍼파라미터 값을 기반으로 deduplication을 진행하였다.

\begin{table}[!th]
\centering
\begin{adjustbox}{max width=\textwidth}
\begin{tabular}{ll|ccccccccc|c}
\toprule
\rowcolor{gray!15}
\textbf{Deduplication} & \textbf{Q$\cdot$F} & \textbf{csatqa} & \textbf{haerae} & \textbf{kmmlu} & \textbf{kobest\_boolq} & \textbf{kobest\_copa} & \textbf{kobest\_hellaswag} & \textbf{kobest\_sentineg} & \textbf{kobest\_wic} & \textbf{MMLU\_proX\_ko} & \textbf{AVG} \\

Document & Ver1 & 18.717 & 19.157 & 26.389 & 50.000 & 61.1 & 46.6 & 63.476 & 51.508 & 10.911 & 37.740 \\
Document & Ver2 & 25.134 & 18.698 & 25.826 & 50.997 & 59.9 & 47.8 & 71.788 & 50.873 & 10.834 & \textbf{39.006} \\
Document & Ver3 & 19.786 & 18.973 & 27.436 & 52.493 & 62.5 & 49.2 & 50.882 & 50.635 & 10.707 & 37.001 \\
Document & Ver4 & 16.578 & 18.882 & 20.645 & 51.282 & 59.3 & 49.2 & 60.705 & 50.079 & 10.418 & 36.188 \\
\midrule
Old-both & Ver1 & 19.251 & 19.890 & 26.100 & 53.276 & 64.0 & 50.2 & 58.186 & 50.556 & 10.860 & 37.947 \\
Old-both & Ver2 & 17.647 & 18.882 & 28.030 & 52.279 & 64.5 & 49.2 & 77.330 & 50.952 & 10.664 & \textbf{39.809} \\
Old-both & Ver3 & 16.578 & 19.157 & 21.259 & 51.282 & 64.1 & 51.8 & 62.972 & 51.032 & 10.613 & 37.399 \\
Old-both & Ver4 & 17.112 & 17.415 & 27.976 & 53.632 & 65.5 & 51.4 & 53.904 & 52.302 & 10.375 & 37.313 \\
\bottomrule
\end{tabular}
\end{adjustbox}
% \caption{deduplication 방식과 quality-filtering 방식에 따른 한국어 벤치마크 성능(Accuracy). `Deduplication'은 bff에서 지원하는 deduplication 방식에 따른 차이, `Q$\cdot$F'는 quality filtering 방식에 따른 차이이다. 모델은 모두 proxy 1B 모델을 활용하였으며 전체 한국어 데이터 중 추출된 한국어 40B 토큰을 학습한 결과이다.}
\caption{Korean benchmark performance(Accuracy) under different deduplication and quality-filtering strategies. `Deduplication' refers to the performance difference based on the deduplication method supported by BFF\tablefootnote{\url{https://github.com/allenai/bff}}, and `Q$\cdot$F' indicates the effect of quality filtering. All experiments use a 1B proxy model trained on 40B Korean tokens extracted from the full Korean corpus.}
\label{tab:pretraining-mix-korean-acc}
\end{table}

\begin{table}[ht]
\centering
\small
\begin{tabular}{llcccc}
\toprule
\textbf{Samples} & \textbf{Category} & \textbf{Version 1} & \textbf{Version 2} & \textbf{Version 3} & \textbf{Version 4} \\
\midrule
\multirow{4}{*}{\textbf{Positive Samples (300k)}} 
  & Korean-Opensource      & 122,340 & 122,340 & 90,000 & 90,000  \\ % 열 5개로 맞춤
  & Instruction-Following & 50,000  & 50,000  & 50,000  & 50,000  \\
  & Qwen-annotated ($>$3) & 37,660  & 37,660  & 37,660  & 37,660  \\
  & Synthetic-LLM & 90,000  & 90,000  & 122,340 & 122,340 \\
\midrule
\multirow{2}{*}{\textbf{Negative Samples (300k)}} 
  & Qwen-annotated (=0)   & $\checkmark$ &        & $\checkmark$ &        \\
  & Random crawl   &     & $\checkmark$    &     & $\checkmark$    \\
\bottomrule
\end{tabular}
\caption{Comparison of dataset composition across Version 1--4. 
Versions 1 and 2 share identical positive samples but differ in negative sample selection, and likewise for Versions 3 and 4. Versions [1,2] emphasize balanced sources, while Versions [3,4] increase the proportion of synthetic datasets.}
\label{tab:pt-quality-filtering}
\end{table}

% \textbf{Quality Filtering.}  우리는 Cross-deduplication을 통해 데이터 중복 문제를 해소한 뒤, 데이터 품질 확보를 위한 filtering 단계를 수행하였다. 언어모델 학습에서 데이터 품질은 결정적인 요소이다. 웹 데이터에는 욕설, 혐오 발언, 사회적 편향을 포함한 유해 콘텐츠뿐 아니라, 스팸·상용구와 같은 무의미한 텍스트가 다수 포함되어 있기 때문이다. 따라서 이러한 저품질 데이터를 효과적으로 제거하고, 고품질 데이터를 확보하는 과정은 필수적이다.

\textbf{Quality Filtering} Following cross-corpus deduplication to mitigate redundancy, we performed a quality filtering step to ensure the reliability of the training corpus. In large-scale language model pretraining, data quality is a pivotal factor influencing downstream performance. Web-sourced corpora often contain toxic content---including profanity, hate speech, and social biases---as well as low-information text such as spam, templated phrases, and boilerplate. Accordingly, rigorous filtering is essential to exclude such low-quality samples and construct a cleaner, high-quality dataset suitable for robust model training.

% 일반적으로 quality filtering은 입력 텍스트에 대해 분류기가 0–5점 사이의 점수를 부여하고, 기준 점수 이하 데이터를 제거하는 방식으로 진행된다. DCLM, UltraFineWeb~\citep{wang2025ultra}에서는 FastText 기반 분류기가 효율적이며 우수한 성능을 보였다고 보고된 바 있다. 영어 데이터의 경우, UltraFineWeb에서 공개한 FastText 기반 필터링 모델을 활용하여 모든 웹 데이터를 정제하였다. 이는 각 데이터셋이 자체적인 정제 과정을 거쳤음에도 불구하고, UltraFineWeb 필터가 DCLM 대비 더 나은 성능을 보였다는 보고에 근거한 추가적 조치였다.

Quality filtering is typically performed by assigning a score between 0 and 5 to each input text using a classifier and removing data below a set threshold. Both DCLM and UltraFineWeb~\citep{wang2025ultra} report that fastText-based classifiers are efficient and effective for this task. For English web data, we additionally applied the fastText-based UltraFineWeb filter, despite the data having already undergone initial filtering, based on reports that it outperforms the DCLM-style classifiers used in earlier stages.

% 반면 한국어는 별도의 품질 분류기를 직접 구축해야 했다. 우리는 DCLM과 UltraFineWeb의 방식을 참고하여 FastText 기반 한국어 quality classifier를 설계하였다. 먼저 FineWeb-Edu~\citep{fineweb_2024}에서 제안한 scoring 프롬프트를 한국어로 번역하여, Qwen3-32B 모델을 활용해 community-OSCAR 한국어 데이터 약 40만 개 샘플에 대해 0–5점 스코어를 부여하였다(Figure~\ref{fig:distribution-of-communityoscar}).

\begin{figure}[ht]
    \centering
    \includegraphics[width=0.7\linewidth]{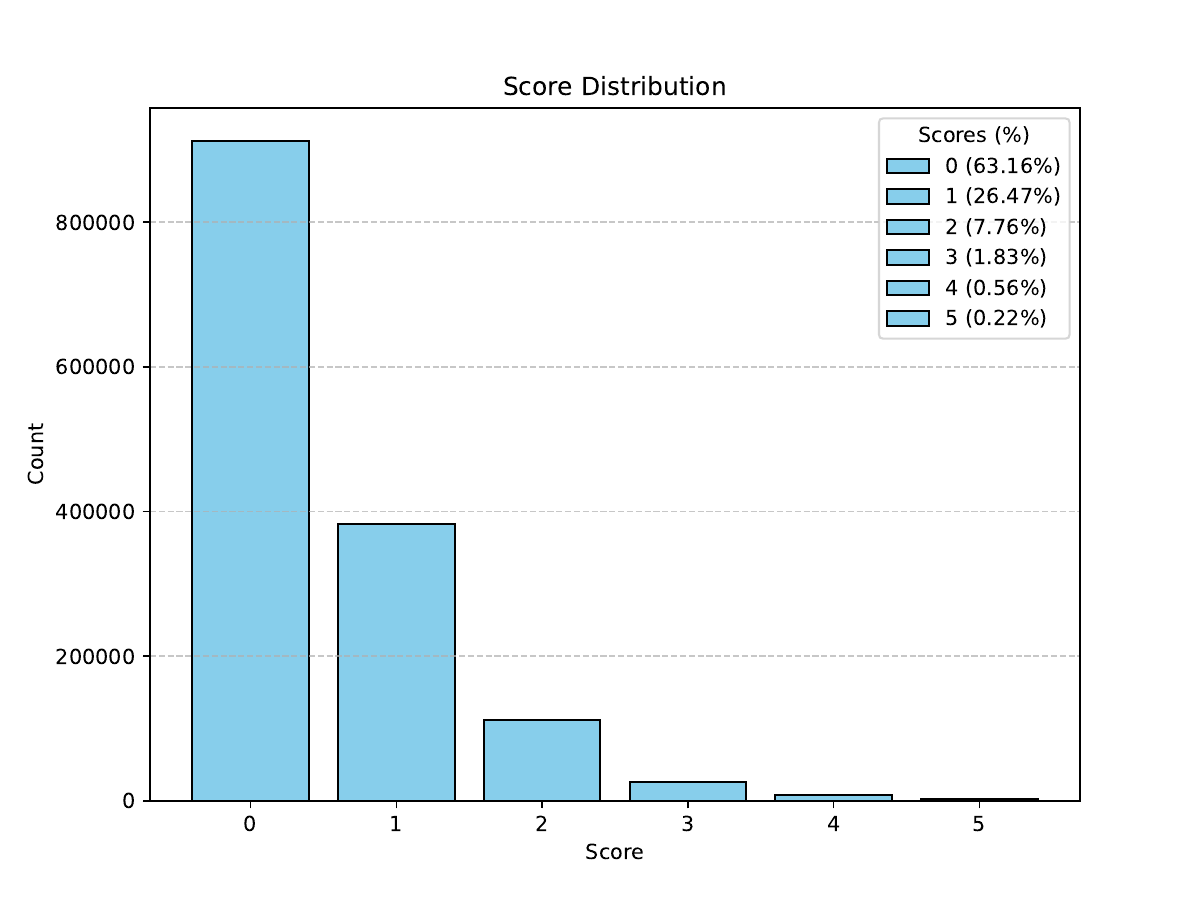}
    % \caption{Qwen3-32B 모델을 FineWeb-Edu scoring 프롬프트를 한국어로 번역한 뒤, 약 40만개의 community-oscar에 scoring한 분포 결과. 대부분의 샘플이 교육 가치가 없는 0점으로 평가됨}
    \caption{Distribution of quality scores assigned to ~400K Korean community-OSCAR samples using Qwen3-32B with translated FineWeb-Edu scoring prompts. The majority of samples were rated as 0, indicating low or no educational value.}
    \label{fig:distribution-of-communityoscar}
\end{figure}

For Korean data, we constructed a dedicated quality classifier, as no pre-existing model was available. Following the approaches of DCLM and UltraFineWeb, we designed a fastText-based classifier tailored to Korean. Using translated scoring prompts from FineWeb-Edu~\citep{fineweb_2024}, we employed Qwen3-32B to assign scores (0–5) to approximately 400K samples from the Korean portion of community-OSCAR (Figure~\ref{fig:distribution-of-communityoscar}).

% 이후 어떤 데이터를 positive/negative 샘플로 설정하느냐에 따라 분류기의 효과가 달라질 수 있으므로, 우리는 네 가지 버전의 FastText 분류기를 설계하였다(Table~\ref{tab:pt-quality-filtering}). Positive 샘플에는 (i) Korean-Opensource (공인 기관 공개 데이터), (ii) Ko-Reasoning 데이터, (iii) Qwen-annotated 데이터(3점 이상), (iv) Synth-FineWeb2 데이터를 활용하였고, Negative 샘플에는 (i) Qwen-annotated 데이터(0점), (ii) community-OSCAR에서 무작위로 샘플링한 데이터를 포함하였다. 각 fasttext quality filter 분류기는 30만 개 positive와 30만 개 negative 샘플로 학습되었다. 이후 4가지 버전의 fasttext 퀄리티 필터링 모델로 필터링된 40B 토큰을 이용해 KORMo-1B proxy을 학습한 후 필터링 효과를 평가했다.(Table~\ref{tab:pretraining-mix-korean-acc}).

To account for variations in classifier effectiveness depending on the choice of positive and negative samples, we designed four versions of fastText-based quality classifiers (Table~\ref{tab:pt-quality-filtering}). Positive samples included:
(i) Korean-Opensource (public datasets released by official institutions),
(ii) Ko-Reasoning data,
(iii) Qwen-annotated samples with scores ≥ 3, and
(iv) Synth-FineWeb2 data.
Negative samples consisted of:
(i) Qwen-annotated samples with a score of 0, and
(ii) randomly selected samples from community-OSCAR.
Each classifier was trained on 300K positive and 300K negative samples.
We then applied each classifier to filter 40B tokens of Korean data and trained a KORMo-1B proxy model to assess filtering effectiveness (Table~\ref{tab:pretraining-mix-korean-acc}).

% 분석 결과 두 가지 주요 시사점을 얻을 수 있었다.  
% \begin{enumerate}
%     \item \textbf{Negative 샘플 구성 효과:} 무작위 웹 데이터를 negative 샘플로 사용한 Version 2가, 명시적으로 0점 처리된 데이터를 사용한 Version 1보다 더 높은 평균 성능을 보였다. 이는 분류기가 명확히 `저품질’로 라벨링된 데이터만 학습하는 것보다, 웹 전반의 다양한 저품질 패턴을 학습하는 것이 모델 일반화에 더 유리함을 시사한다.  
%     \item \textbf{Positive 샘플 구성 효과:} 정제된 고급 웹 데이터를 중심으로 한 Version 1, 2가 합성 데이터 비중을 높인 Version 3, 4보다 consistently 우수한 성능을 보였다. 이는 고품질 실제 데이터의 본질적 가치를 재확인해 주는 결과이다. 다만 한국어 고급 웹 데이터는 절대적으로 부족하기 때문에, 대규모 사전학습에 필요한 수조 단위 토큰을 확보하기 위해서는 여전히 합성 데이터를 병행할 수밖에 없다.  
% \end{enumerate}

Our analysis yields two key insights:
\begin{enumerate}
\item \textbf{Effect of Negative Sample Composition:} Version 2, which utilized randomly sampled web data as negative examples, achieved higher average performance than Version 1, which used explicitly labeled low-quality (score 0) samples. This suggests that exposing the classifier to a broader spectrum of naturally occurring low-quality patterns on the web is more beneficial for generalization than training solely on narrowly defined “poor-quality” data.
\item \textbf{Effect of Positive Sample Composition:} Versions 1 and 2, primarily composed of curated, high-quality web data, consistently outperformed Versions 3 and 4, which contained a higher proportion of synthetic data. This result reaffirms the intrinsic value of high-quality, authentic data. However, due to the absolute scarcity of high-quality Korean web data, incorporating synthetic data remains necessary to acquire the trillions of tokens required for large-scale pre-training. 
\end{enumerate}

% 종합적으로 우리는 Version 2가 가장 효과적인 필터링 전략임을 확인하였다. 따라서 최종적으로 Version 2 기반 FastText 분류기를 전체 한국어 데이터에 적용하여 KORMo의 사전학습용 데이터셋을 정제하였다.  

Overall, Version 2 proved to be the most effective filtering strategy. Accordingly, we adopted the Version 2–based fastText classifier to refine the entire Korean corpus used for KORMo’s pre-training dataset.

% \input{Tables/table-datamix-en}

% (민준)
\subsection{Pretraining}
% 우리는 모든 데이터 준비가 완료된 이 후 비로소 사전학습을 진행했다. 본 장에서는 KORMo 사전학습의 핵심 요소를 다음 세 가지 관점에서 탐구한다:  
We initiated pre-training only after completing all stages of data preparation. This section explores the core components of KORMo’s pre-training from the following three perspectives:

% \begin{itemize}
%     \item \textbf{최적의 학습률 탐색:} 제안한 아키텍처와 데이터 환경에서 적합한 learning rate 설정
%     \item \textbf{언어 비율:} 학습 단계별 한국어와 영어 데이터 비율의 조정
%     \item \textbf{Synthetic 데이터 비율:} synthetic 데이터 편중 사용에 따른 잠재적 한계 검증
% \end{itemize}
\begin{itemize}
    \item \textbf{Optimal Learning Rate Search:} Determining an appropriate learning rate configuration for the proposed architecture and data environment.
    \item \textbf{Language Ratio:} Adjusting the proportion of Korean and English data across training stages.
    \item \textbf{Synthetic Data Ratio:}Examining the potential limitations arising from an overreliance on synthetic data.
\end{itemize}

\subsubsection{Optimal Learning Rate Search}
% 사전학습에서 가장 중요한 초기 결정 중 하나는 적절한 learning rate 설정이다. 학습률은 수렴 속도와 최종 성능에 큰 영향을 미치며, 값이 지나치게 크면 발산이나 불안정한 진동을 초래하고, 지나치게 작으면 학습이 과도하게 느려지거나 suboptimal 상태에 머무를 수 있다. 특히 모델 크기와 구조에 따라 최적의 학습률은 크게 달라지므로, 본 연구에서는 실제 목표 모델인 KORMo-10B에서 직접적인 탐색을 수행하였다.  
One of the most critical early decisions in pre-training is the selection of an appropriate learning rate. The learning rate significantly influences both convergence speed and final performance: excessively high values can lead to divergence or unstable oscillations, while excessively low values may result in slow learning or convergence to a suboptimal state. Since the optimal learning rate can vary substantially depending on model size and architecture, we conducted a direct search using the target model, KORMo-10B.

% KORMo-10B는 앞서 제안한 학습 디자인 Pre-LN, Intra-document masking, Next-token prediction 목표 함수, 자체 개발 bilingual 토크나이저를 기반으로 구축되었다. 학습률 탐색은 Global Batch Size 1024, Sequence Length 4096 환경에서 2,000 스텝 동안 \{1e-4, 3e-4, 5e-4, 7e-4, 9e-4, 1e-3, 1.5e-3, 3e-3\} 범위의 후보 값을 대상으로 진행했으며, 실제 상황과 같게 만들기 위해 모든 사전학습 데이터에서 일부분을 랜덤 추출하여 진행했다.  
KORMo-10B was built upon the proposed training design, incorporating Pre-LN architecture, intra-document masking, a next-token prediction objective, and a custom-developed bilingual tokenizer. The learning rate search was conducted with a global batch size of 1024 and a sequence length of 4096 over 2,000 steps, exploring candidate values in the range of \{1e-4, 3e-4, 5e-4, 7e-4, 9e-4, 1e-3, 1.5e-3, 3e-3\}. To closely approximate real training conditions, a randomly sampled subset of the entire pretraining corpus was used during this search.

\begin{figure}[!h]
    \centering
    \makebox[\textwidth][c]{%
        \includegraphics[width=\textwidth]{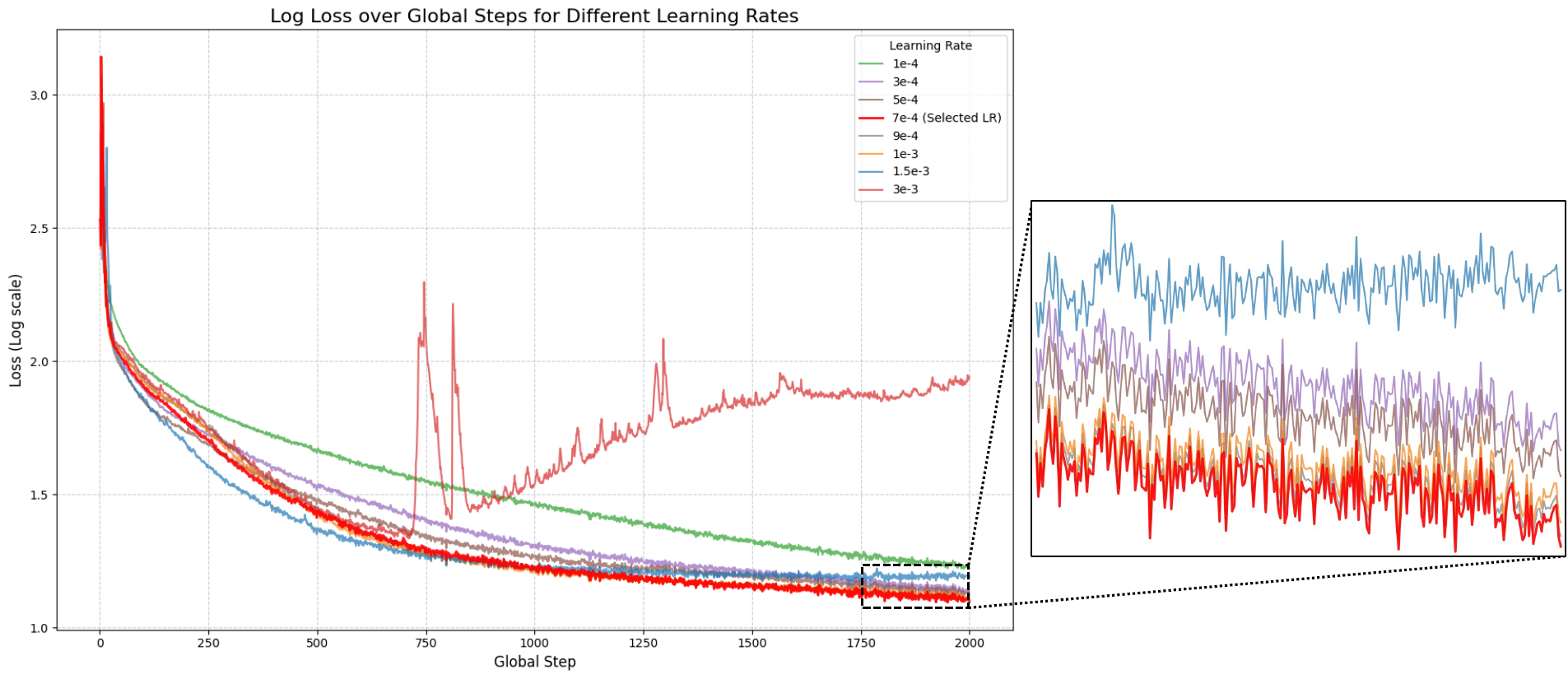}
    }
    \caption{Comparison of loss curves across different learning rates during KORMo-10B pre-training.}
    \label{fig:learning rate search}
\end{figure}

% Figure~\ref{fig:learning rate search}에 나타난 결과는 학습률에 따른 뚜렷한 최적화 양상의 차이를 보여준다.  
The results presented in Figure~\ref{fig:learning rate search} reveal distinct optimization patterns across different learning rates.
\begin{itemize}
    \item High learning rates (e.g., 3e-3) led to rapid initial loss reduction but soon caused oscillations and instability.
    \item Low learning rates (e.g., 1e-4) exhibited highly stable convergence but progressed too slowly to reach optimal performance.
    \item Moderate learning rates (7e-4, 9e-4) achieved both stability and fast convergence, with 7e-4 yielding the lowest overall loss across all steps.
\end{itemize}

% 따라서 우리는 최종 모델 학습의 최적 학습률로 \textbf{7e-4}를 채택하였다. 이 결과는 KORMo-10B의 구조와 데이터 환경에서 안정적 수렴과 효율적인 최적화가 가능한 설정임을 보여준다.  
Accordingly, we adopted a learning rate of 7e-4 as the optimal setting for the final model training. This result demonstrates that, under the architecture and data environment of KORMo-10B, this configuration enables stable convergence and efficient optimization.

%자 그러면 사전학습 때 최적의 데이터 필터링과 비율은 대략 알게되었어. 마지막으로 적용한건 커리큘럼 기반 사전학습인데 이건 요즘 누구나 다 하잖아? 그래서 그냥 기존 연구를 따라했어. 근데 이때 한 가지 트릭을 넣었는데, 앞서 데이터 필터링 실험에 왜 Instruction tuning 데이터를 소량 섞은줄 알아? 우리는 사전학습때 plaintext만 학습하는건 비효율 적이라고 생각해 왜냐하면 인간도 아기일 때 엄마와 대화하는 형식으로 하면서 학습을 하자나? 여기서 착안해 이 단계뿌터 instruction tuning 데이터를 아주 소량씩 넣어 점점 비율을 늘리면 좋지 않겠냐는 가설로 시작되었어. 그래서

\begin{figure}[htb]
    \centering
    
    \begin{subfigure}[b]{0.49\textwidth} 
        \centering
        \includegraphics[width=\linewidth]{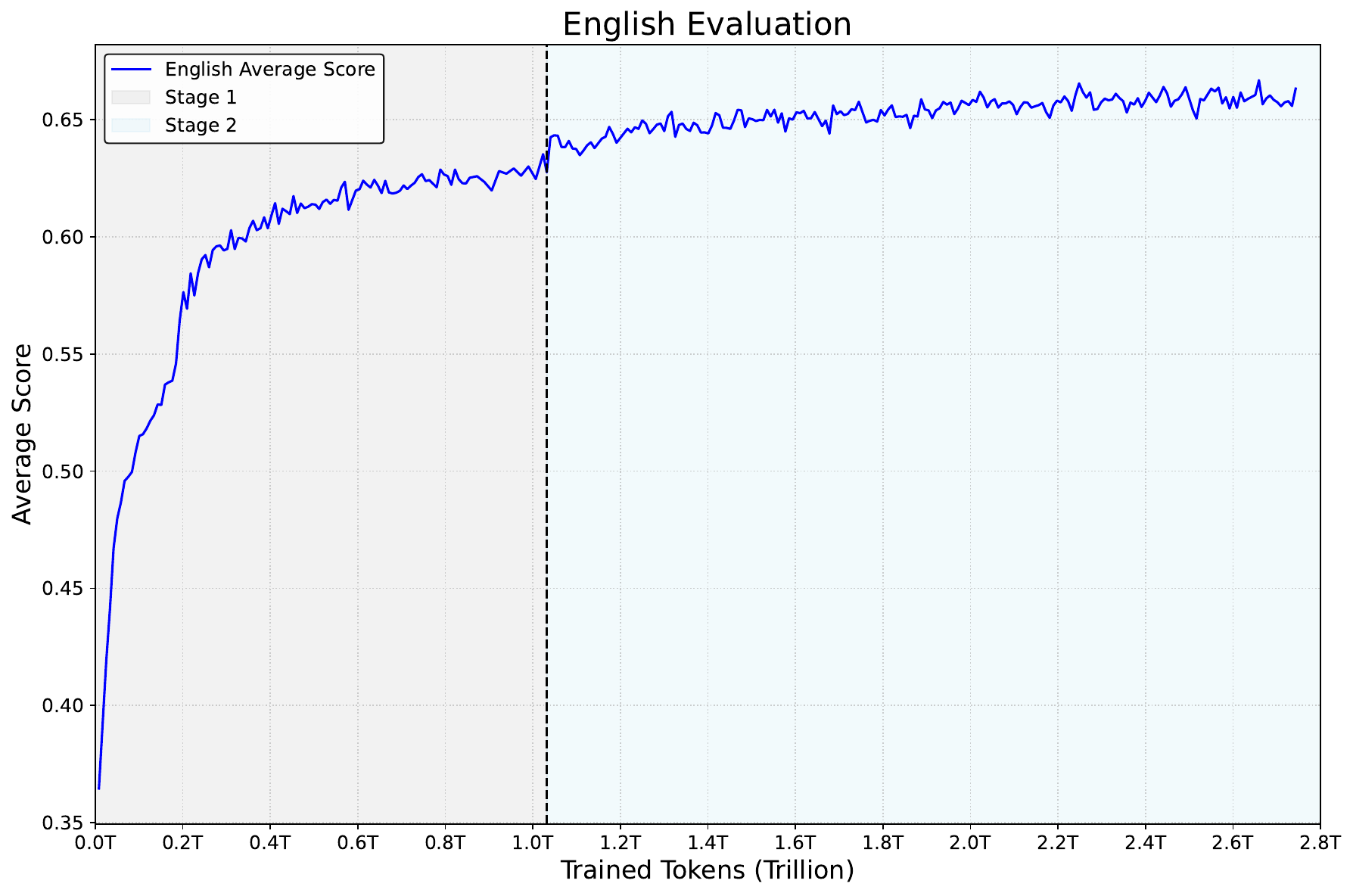} 
        \caption{English Tasks Average Score}
        \label{fig:en_plot}
    \end{subfigure}
    \hfill
    \begin{subfigure}[b]{0.49\textwidth}
        \centering
        \includegraphics[width=\linewidth]{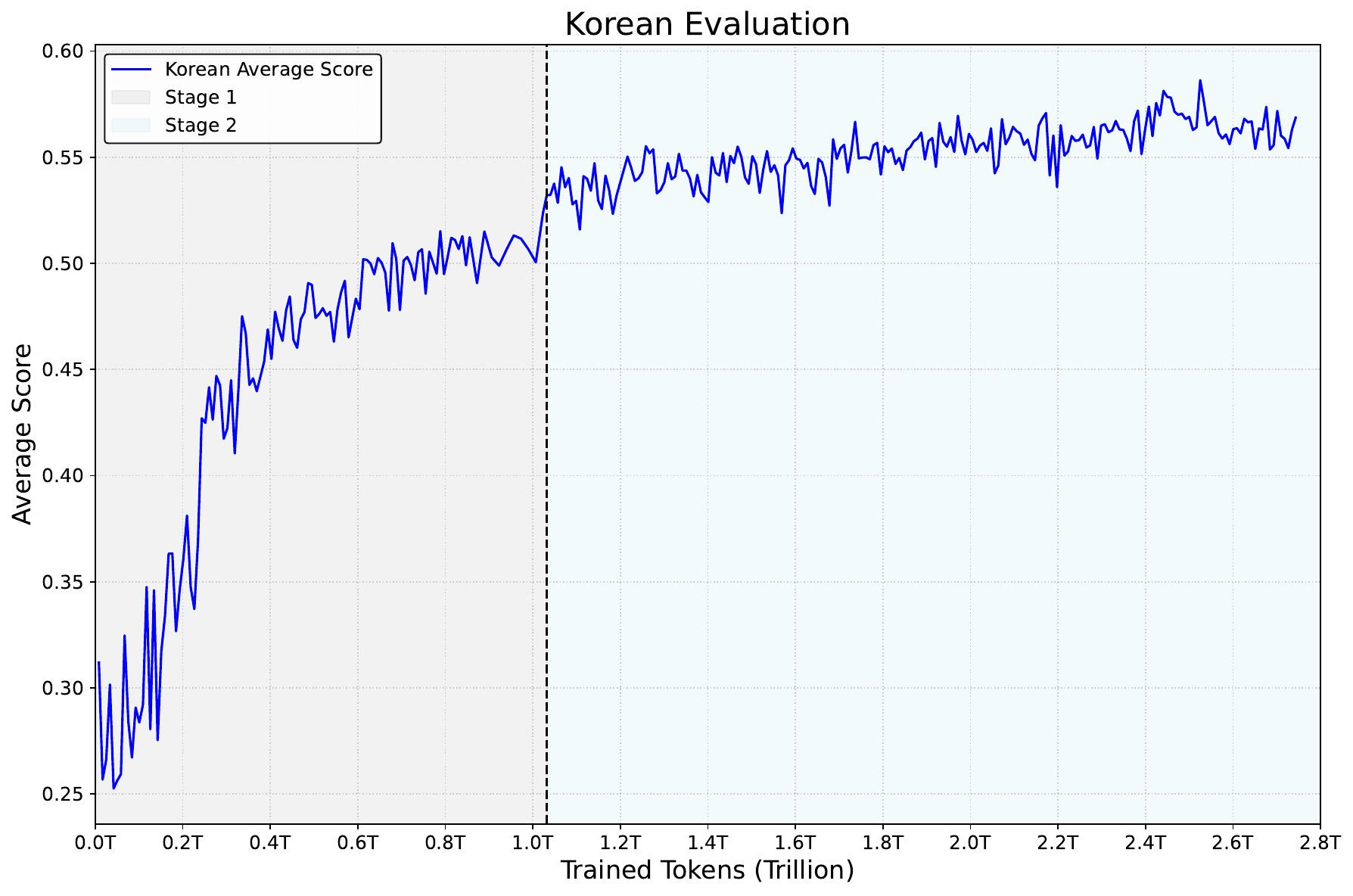}
        \caption{Korean Tasks Average Score}
        \label{fig:kr_plot}
    \end{subfigure}
    
    \caption{Comparison of Average Performance between English and Korean Evaluation Tasks During Pretraining stages.}
    \label{fig:pretrain_evals}
\end{figure}

% \subsubsection{제안하는 언어 비율 및 사전학습 단계}
\subsubsection{Proposed Language Ratio and Pre-training Stages}
\paragraph{Language Proportion.}  
% Bilingual 언어모델에서 두 언어의 데이터 비율은 모델 성능을 좌우하는 핵심 요인이다. 기존 연구에 따르면 다국어 환경에서 target 언어의 비중이 1.14\% 수준에 불과하더라도 일정 수준의 성능을 달성할 수 있음이 관찰되었다~\cite{xue-etal-2021-mt5}. 반면 LLaMA-2의 경우 한국어 데이터가 약 0.06\%에 불과하여, 사실상 한국어 성능을 평가하기 어려웠다. bilingual 환경에서 체계적으로 비율을 탐구한 사례는 드물지만, 최근 연구에서는 target 언어 비중을 1.5--5\%까지 늘려도 준수한 성능을 유지할 수 있음을 보고하였다~\cite{seto-etal-2025-training}. 그러나 이러한 결과들은 언어 간 유사성과 데이터 품질에 따라 결과가 크게 달라질 수 있음을 유념해야 한다. 결론적으로 본 연구에서는 영어와 한국어가 구조적으로 상당히 이질적임을 고려하여, 한국어 비중을 5\% 이상으로 설정하여 안정성을 도모했다.  
In bilingual language models, the data ratio between the two languages serves as a crucial determinant of model performance. Previous studies have shown that even when the target language accounts for as little as 1.14\% of the total data in a multilingual setting, a reasonable level of performance can still be achieved~\cite{xue-etal-2021-mt5}. In contrast, LLaMA-2 includes only about 0.06\% Korean data, making it practically infeasible to assess Korean performance meaningfully.
Although systematic investigations of language ratios in bilingual settings are limited, recent research has reported that increasing the target language proportion to 1.5--5\% can still maintain competitive performance~\cite{seto-etal-2025-training}. However, such results may vary substantially depending on interlingual similarity and data quality.
Taking into account the structural heterogeneity between English and Korean, this study set the Korean data proportion to over 5\% to ensure stable bilingual learning.

\paragraph{Training Phase.}  
% 최근 LLM 사전학습은 커리큘럼 러닝의 효과를 활용하기 위해 데이터 품질을 점차 높여가는 다단계 학습 방식을 채택하는 추세다. KORMo의 경우 한국어 영어 비율을 토대로 전체 학습 토큰 수가 3T를 초과하기 어려운 현실적 제약이 있어, 두 단계 학습 전략을 도입하였다.  
Recent trends in LLM pre-training adopt multi-stage training strategies that progressively increase data quality to leverage the benefits of curriculum learning. In the case of KORMo, due to practical constraints that limit the total number of training tokens to below 3T under the chosen Korean–English ratio, we adopted a two-stage pre-training strategy.

% \begin{itemize}
%     \item \textbf{Stage 1:} 비교적 저품질의 웹 데이터를 중심으로 약 1T tokens을 학습한다. 구체적으로 Table~\ref{tab:pretrining-datasets-main}의 `stage1'에 해당하는 DCLM (960B), Korean Web (36.3B), Korean-CC (6.2B)를 포함하였다. 이 단계의 목표는 모델이 대규모 웹 텍스트에 대한 기본적 언어 이해를 습득하고, noisy 데이터에 대해 robust하게 적응하는 것이다.  

%     \item \textbf{Stage 2:} 고품질 웹 텍스트, synthetic 데이터, 그리고 짧은 reasoning path를 포함한 저난이도 reasoning 데이터를 학습한다. Table~\ref{tab:pretrining-datasets-main}의 `stage2'로 표기된 총 1.8T tokens (영어 1.7T, 한국어 0.1T) 규모이며, stage1에서 습득한 기초 언어 능력을 기반으로 구조화된 데이터의 특성을 반영해 고급 언어 이해 및 추론 능력을 강화한다.  
% \end{itemize}
\begin{itemize}
    \item \textbf{Stage 1:} Approximately 1T tokens were trained primarily on relatively low-quality web data. Specifically, this stage included DCLM (960B), Korean Web (36.3B), and Korean-CC (6.2B) datasets corresponding to the “stage1” entries in Table~\ref{tab:pretrining-datasets-main}. The objective of this stage is to enable the model to acquire fundamental language understanding from large-scale web text and to develop robustness against noisy data.

    \item \textbf{Stage 2:} The model is trained on high-quality web text, synthetic data, and low-difficulty reasoning datasets that include short reasoning paths. As shown in the “stage2” entries of Table~\ref{tab:pretrining-datasets-main}, this stage comprises a total of 1.8T tokens (1.7T in English and 0.1T in Korean). Building upon the foundational language abilities acquired in Stage 1, this phase aims to strengthen advanced language understanding and reasoning capabilities by incorporating structured and higher-quality data.
\end{itemize}

\begin{table}[!h]
\centering
\label{tab:model_configs}
\begin{tabular}{ll}
\toprule
\multicolumn{2}{l}{\textbf{Architecture Details}} \\
\midrule
Number of Total Parameters & 10.75B\\
Number of Embedding Parameters & 1.02B\\
Number of Non-Embedding Parameters & 9.73B\\
Vocabulary Size & 125,184\\
Hidden Size & 4096 \\
Intermediate Size & 16,384 \\
Number of Hidden Layers & 40 \\
Number of Attention Heads & 32 \\
Number of Key/Value Heads & 8 \\
Head Dimension & 128 \\
Attention Dropout & 0.0 \\
Attention Bias & 0.0 \\
Weight tying & False \\
Hidden Activation & \texttt{SwiGLU} \\
Normalizer & RMSNorm\\
RMS Norm Epsilon & $1e-05$ \\
RoPE Theta & $5e+5$ \\
Data type & \texttt{bfloat16} \\
\bottomrule
\end{tabular}
\caption{\kormo-10B Configurations}
\label{tab:kormo_config}
\end{table}

\paragraph{Training Details.}  
% Table~\ref{tab:kormo_config}는 KORMo-10B 모델의 configuration을 요약한다. 주요 학습 설정은 다음과 같다.  
Table~\ref{tab:kormo_config} summarizes the configuration of the KORMo-10B model. The key training settings are as follows.

% \begin{itemize}
%     \item \textbf{모델 구조:} \autoref{sec:proxy-expr}에서 확인된 결과를 반영하여 Pre-LN 구조의 transformer decoder를 사용한다. 학습은 intra-document attention masking과 next-token prediction 목표 함수를 기반으로 수행하며, 효율성을 위해 flash-attention-3~\citep{shah2024flashattention} 커널을 적용하였다.  

%     \item \textbf{Context length:} 기본적으로 sequence length 4096을 기준으로 packing을 적용하였다. 다만 reasoning 데이터의 경우, packing이 연결된 추론의 절단으로 인해 부정적 영향을 줄 수 있으므로, 별도의 packing 없이 샘플 단위 padding으로 처리하였다.  

%     \item \textbf{Weight decay:} OLMo 연구~\citep{olmo20242}를 참고하여 토큰 임베딩에는 weight decay를 적용하지 않았다. 이는 임베딩 벡터 크기가 과도하게 축소되는 부작용을 방지하기 위함이다.  

%     \item \textbf{Initialize optimizer:} 각 스테이지를 시작할 때, 데이터 분포의 변화에 안정적으로 적응하기 위해 optimizer를 초기화 하고 전체 스텝 0.03\%의 learning rate warm-up 을 적용한다.
% \end{itemize}

\begin{itemize}
    \item \textbf{Model Architecture:} Reflecting the findings presented in \autoref{sec:proxy-expr}, the model employs a transformer decoder with a Pre-LN architecture. Training is conducted using intra-document attention masking and a next-token prediction objective, while the flash-attention-3 kernel~\citep{shah2024flashattention} is applied to improve computational efficiency.

    \item \textbf{Context length:} By default, sequence packing was applied based on a sequence length of 4096. However, for reasoning datasets, packing was avoided since it could negatively affect coherence by truncating connected reasoning chains. Instead, sample-level padding was applied without additional packing.

    \item \textbf{Weight decay:} Following the approach of the OLMo study~\citep{olmo20242}, weight decay was not applied to token embeddings. This decision was made to prevent undesirable side effects, such as excessive shrinkage of embedding vector magnitudes. 

    \item \textbf{Initialize optimizer:} At the beginning of each stage, the optimizer is reinitialized to ensure stable adaptation to the shift in data distribution, and a learning rate warm-up of 0.03\% of the total training steps is applied.

    \item \textbf{Training Resources:} All experiments were conducted in a multinode environment using NVIDIA H200 GPUs, with a total of 128 GPUs utilized in parallel. Considering the model size and number of parameters, the Fully Sharded Data Parallel (FSDP)~\citep{zhao2023pytorch} strategy was adopted to enable parameter sharding across nodes and efficient memory utilization.
\end{itemize}

% Figure~\ref{fig:pretrain_evals}는 사전학습 단계에서 학습 토큰 수에 따른 영어 및 한국어 평가 평균 성능을 나타낸다. 영어 평가는 MMLU, BoolQ, COPA, ARC-Challenge, ARC-Easy, AGIEval-EN, CommonsenseQA, OpenBookQA, PIQA, HellaSwag, SocialIQA, Winogrande 등 총 12개 벤치마크로 구성하였으며, 한국어 평가는 KMMLU, CSATQA, CLICK, Haerae, K2-Eval, KoBEST의 6개 벤치마크를 사용하였다.  
Figure~\ref{fig:pretrain_evals} illustrates the average English and Korean evaluation performance as a function of the number of training tokens during pre-training.
The English evaluation comprises 12 benchmarks—MMLU, BoolQ, COPA, ARC-Challenge, ARC-Easy, AGIEval-EN, CommonsenseQA, OpenBookQA, PIQA, HellaSwag, SocialIQA, and Winogrande—and the Korean evaluation consists of six benchmarks: KMMLU, CSATQA, CLICK, Haerae, K2-Eval, and KoBEST.

% 관측 결과 Stage 1에서는 두 언어 모두 급격한 성능 향상을 보였으며, 이는 모델이 기본적인 언어 지식과 구문 구조를 빠르게 습득했음을 시사한다. Stage 2에서는 성능 곡선이 완만한 상승세로 전환되었는데, 이는 고품질 데이터가 모델의 기존 능력을 정교화하고 추론 능력을 점진적으로 강화했음을 보여준다.  
The observations indicate that in Stage 1, both languages exhibited a sharp performance improvement, suggesting that the model rapidly acquired fundamental linguistic knowledge and syntactic structures. In Stage 2, the performance curve transitioned into a more gradual upward trend, implying that the high-quality data refined the model’s existing capabilities and progressively enhanced its reasoning ability.

% 다만 Figure~\ref{fig:kr_plot}에서 확인되는 한국어 성능 곡선은 영어 곡선(Figure~\ref{fig:en_plot}) 대비 진동(oscillation)이 두드러진다. 한국어 성능의 변동성은 (1) 사전학습 데이터에서 한국어의 절대 비중이 낮아 내부 표현이 불안정해질 수 있다는 점, (2) 평가 벤치마크 수가 영어(12개)보다 한국어(6개)에서 적어 개별 벤치마크의 편차가 평균에 더 크게 반영된다는 점에서 비롯된 것으로 해석된다. 그럼에도 불구하고, Stage 2 학습 종료 시 한국어 성능은 평균 0.55 이상으로 안정화되었으며, KORMo 모델이 bilingual 환경에서도 견고한 성능을 확보했음을 확인하였다.  
However, as shown in Figure~\ref{fig:kr_plot}, the Korean performance curve exhibits more pronounced oscillations compared to the English curve in Figure~\ref{fig:en_plot}. This variability in Korean performance can be attributed to two main factors:
(1) the relatively low proportion of Korean data in the pretraining corpus, which may lead to instability in internal representations; and
(2) the smaller number of evaluation benchmarks for Korean (six) compared to English (twelve), causing individual benchmark fluctuations to have a greater impact on the overall average.
Nevertheless, by the end of Stage 2, Korean performance stabilized at an average score above 0.55, confirming that the KORMo model achieved robust bilingual performance.

% 한편, 학습 토큰을 2.8T 이상으로 늘릴 경우 추가적인 성능 향상이 나타났으나, 비용 대비 효율성을 고려할 때 동일한 계산 자원을 mid-training 또는 post-training 단계에 투자하는 것이 더 합리적이라고 판단하였다.  
Meanwhile, when the total number of training tokens exceeded 2.8T, additional performance gains were observed. However, considering cost-effectiveness, we concluded that allocating the same computational resources to mid-training or post-training stages would be a more reasonable strategy.

\subsection{Impact of Augmented Data Diversity}

\begin{figure}[htb]
    \centering
    
    \begin{subfigure}[b]{0.49\textwidth} 
        \centering
        \includegraphics[width=\linewidth]{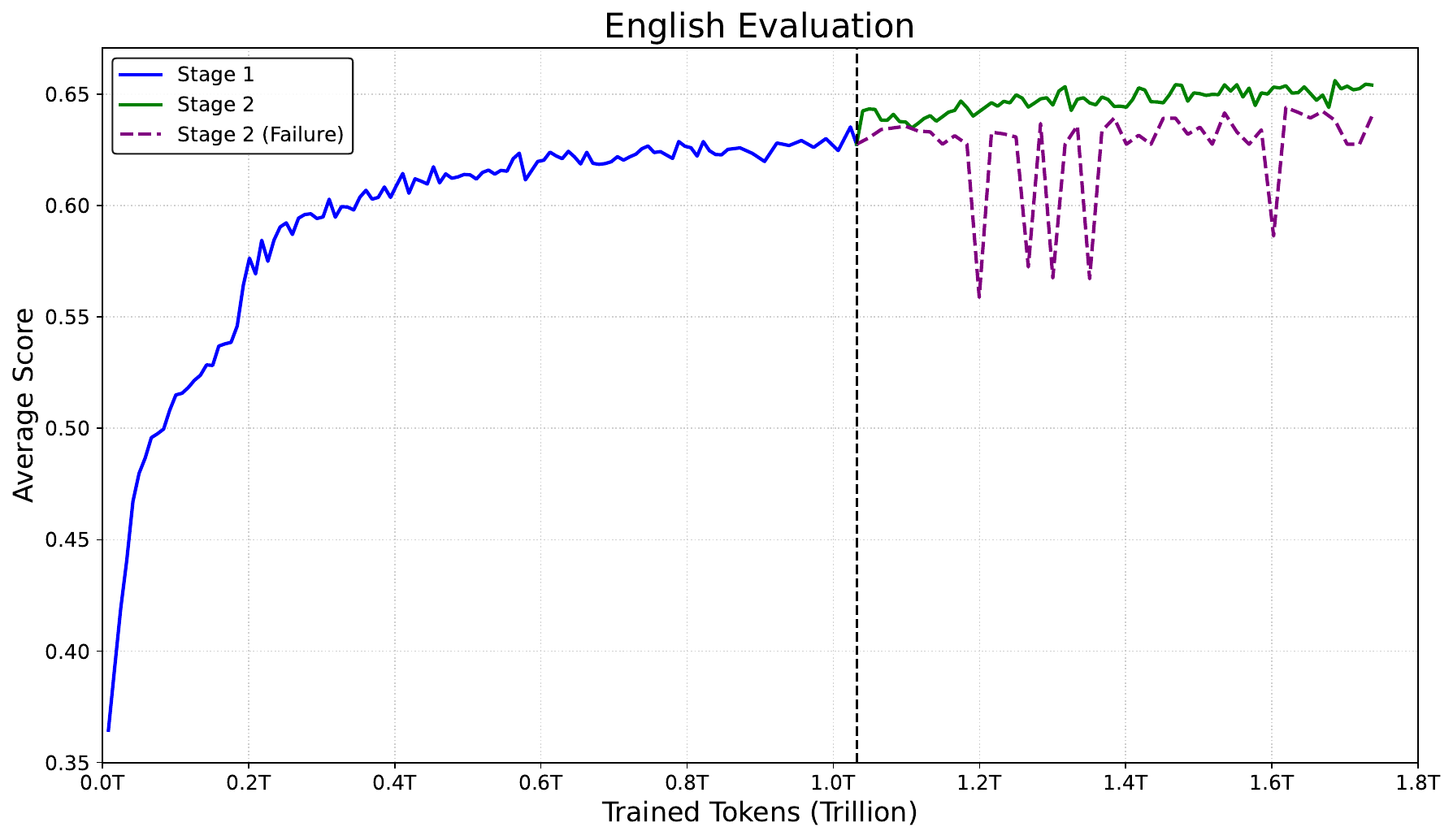} 
        \caption{English Tasks Average Score}
        \label{fig:fail_en_plot}
    \end{subfigure}
    \hfill
    \begin{subfigure}[b]{0.49\textwidth}
        \centering
        \includegraphics[width=\linewidth]{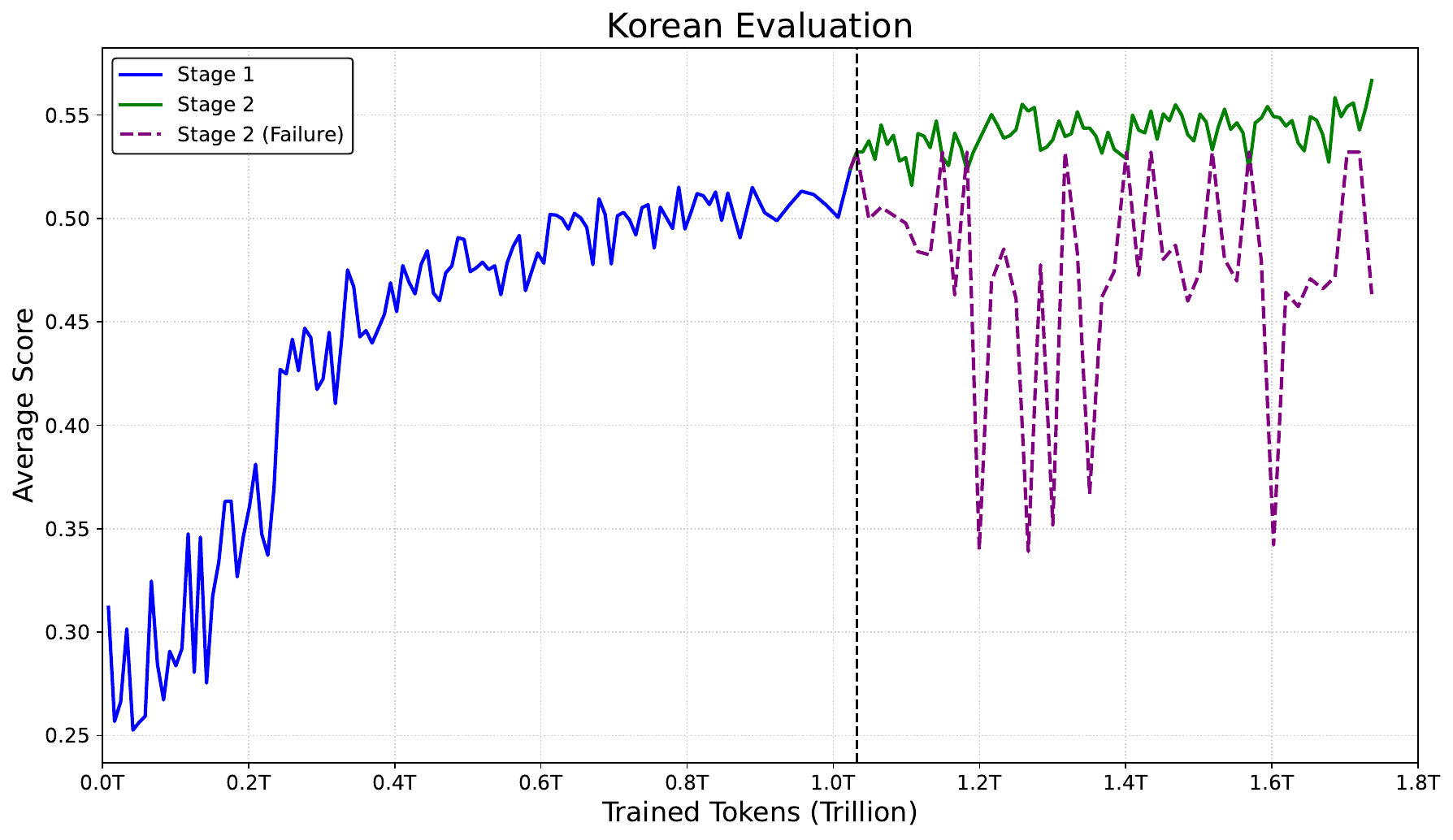}
        \caption{Korean Tasks Average Score}
        \label{fig:fail_kr_plot}
    \end{subfigure}
    
    \caption{Graph illustrating a failed pretraining case using a single synthetic dataset.}
    \label{fig:failure_evals}
\end{figure}

% KORMo는 Stage 2에서부터 본격적으로 대규모 증강데이터를 활용하였다. 그러나 해당 단계에서 증강데이터의 다양성이 부족할 경우 모델 학습에 심각한 부작용이 발생함을 관찰하였다. Figure~\ref{fig:failure_evals}의 Stage 2 (Failure) 곡선은 앞서 제시한 Stage 2와 같은 환경이나 다만 한국어 증강 데이터를 `Synth-Nemo-HQ'만 사용해 단일 synthesizer에서 생성된 데이터로 학습한 실험결과이다. 이로서 다수의 synthesizer가 사용된 증강데이터와 단일 synthesizer에서 만들어진 증강데이터가 언어모델 학습에 미치는 영향을 비교할 수 있다.
KORMo began leveraging large-scale augmented data in Stage 2. However, we observed that insufficient diversity in the augmented data at this stage led to severe adverse effects during model training. The Stage 2 (Failure) curve in Figure~\ref{fig:failure_evals} presents results obtained under the same configuration as the standard Stage 2 setup, except that the Korean augmented data consisted solely of samples generated by a single synthesizer, Synth-Nemo-HQ. This comparison allows us to examine the impact of using augmented data generated from multiple synthesizers versus that produced by a single synthesizer on language model training.

% Stage 2 (Failure) 모델의 성능을 살펴보면 두 언어 모두 Stage 1에서 달성한 최고 성능을 유지하지 못하고, 전반적으로 성능이 하락하며 큰 진동을 보였다. 구체적으로, 영어의 경우 성능이 0.55--0.63 범위에서 불안정하게 요동쳤고, 한국어는 더욱 극적인 하락을 보여 Stage 1 초기 수준으로 성능이 역전되었다. 이러한 급격하고 비가역적인 성능 저하는 \textbf{model collapse} 현상으로 해석된다.
Examining the performance of the Stage 2 (Failure) model reveals that neither language maintained the peak performance achieved in Stage 1; instead, both experienced overall degradation accompanied by pronounced oscillations. Specifically, English performance fluctuated unstably within the range of 0.55–0.63, while Korean performance declined even more sharply, reverting to levels comparable to the early Stage 1 phase. This abrupt and irreversible degradation is interpreted as a manifestation of model collapse.

% collapse의 주요 원인은 다음으로 두 가지로 유추해 볼 수 있다.  
The primary causes of this collapse can be inferred as follows:
% \begin{enumerate}
%     \item \textbf{단일 모델 기반 합성 데이터:} 모든 한국어 데이터가 단일 모델(Qwen3-30B-A3B)의 출력으로 구성되면서, 해당 모델의 내재적 편향과 오류가 반복적으로 학습되었을 수 있다 \cite{shumailov2024nature,long2024synthreview,bukharin2024diversity}. 특히 Stage 2와 같이 정제된 데이터 기반 학습 단계에서는 합성 데이터의 다양성 부족이 광범위한 지식 분포를 빠르게 소실하게 만들었다 \cite{shumailov2023curse,diversityimpact2024,gerstgrasser2024breakcollapse}.  
%     \item \textbf{Seed와 프롬프트 유형의 단일성:} seed 데이터와 프롬프트 유형이 제한될 경우 synthetic 데이터 분포가 협소해진다. 이로 인해 모델이 제한된 패턴에 과적합되며, Stage 1에서 습득한 일반적 언어 이해 및 추론 능력을 잃게 될 수 있다 \cite{bukharin2024diversity}.  
% \end{enumerate}

\begin{enumerate}
    \item \textbf{Single-Model-Based Synthetic Data:} All Korean data were generated solely from a single model (Qwen3-30B-A3B), which may have caused the repeated learning of that model’s inherent biases and errors~\cite{shumailov2024nature,long2024synthreview,bukharin2024diversity}. Particularly in refined training stages such as Stage 2, the lack of diversity in synthetic data led to a rapid loss of the model’s broad knowledge distribution~\cite{shumailov2023curse,diversityimpact2024,gerstgrasser2024breakcollapse}.

    \item \textbf{Uniformity of Seed and Prompt Types:} When the seed data and prompt types are limited, the resulting synthetic data distribution becomes narrow. This constraint can lead the model to overfit to restricted patterns, potentially causing the loss of general language understanding and reasoning abilities acquired during Stage 1~\cite{bukharin2024diversity}.
\end{enumerate}

% 종합하면, 증강데이터 활용은 고품질 학습에 필수적이지만, synthetic 데이터 생성에 있어 다양한 시드와 모델, 프롬프트를 활용하지 않을 경우 심각한 성능 붕괴를 유발할 수 있음을 확인하였다. 따라서 대규모 bilingual 언어모델 구축에서는 증강데이터의 양뿐 아니라 다양성을 확보하는 전략이 반드시 병행되어야 한다.

In summary, while the use of augmented data is essential for high-quality training, our findings demonstrate that relying on limited seeds, models, or prompt types in synthetic data generation can lead to severe performance degradation. Therefore, in large-scale bilingual language model development, ensuring not only the quantity but also the diversity of augmented data is imperative for stable and effective learning.
% (총괄: 민경)
\section{Mid-training}

% 최근 언어모델은 활용 목적에 따라 다양한 형태로 발전하고 있으며, 이에 따라 \textit{mid-training} 단계가 점차 표준화되고 있다. 특히 추론 중심 언어모델에서는 장문 맥락 이해와 복잡한 추론 능력의 내재화가 필수적이므로, 사전학습 이후 이를 보완하는 mid-training 과정이 필요하다 \citep{liu2023_lost_in_the_middle,gao2025_how_to_train_long_context}. KORMo 모델 또한 (1) 컨텍스트 길이 확장, (2) 추론 능력 강화라는 두 가지 목표를 설정하고 mid-training을 설계하였다. 
Recent language models have diversified according to their intended applications, leading to the gradual standardization of the \textit{mid-training} stage. In reasoning-oriented models, the ability to comprehend long contexts and internalize complex reasoning processes is essential, thereby necessitating an intermediate training phase following pre-training~\citep{liu2023_lost_in_the_middle, gao2025_how_to_train_long_context}.
The KORMo model was similarly designed with two primary objectives for its mid-training stage: (1) to extend context length and (2) to enhance reasoning capability.

% (한결)
\subsection{Long Context Training}

% 최근 연구에 따르면 추론 태스크는 복잡한 reasoning trace를 포함하는 장문 출력을 요구하며, 이에 따라 모델이 긴 입력과 출력을 안정적으로 처리할 수 있도록 mid-training에서 long-context 학습이 필수적이다 \citep{liu2023_lost_in_the_middle}. KORMo 역시 추론 및 장문 텍스트 처리를 위해 long-context 학습 절차를 포함하였다.  
Recent studies have shown that reasoning tasks require long-form outputs containing complex reasoning traces. Consequently, long-context training during the mid-training phase is essential to ensure that models can reliably process both long inputs and outputs~\citep{liu2023_lost_in_the_middle}.
Following recent studies, we adopted a similar long-context training procedure for KORMo to enhance its capability in complex reasoning and long-text comprehension.

\paragraph{Data Preparation.}
% 영어 데이터셋의 경우 ProLong에서 제안된 방식을 참조하여 \texttt{prolong-data-64k}를 기반으로 하되, 최대 길이를 32k 토큰으로 \textit{truncation}하여 활용하였다 \citep{prolong64k}. 한국어 데이터셋은 대규모 장문 텍스트를 책 단위로 재구성한 뒤 약 32k 토큰 길이로 \textit{packing}하여 long-context 학습에 적합하게 변환하였다.  
For the English dataset, we followed the approach proposed in ProLong~\citep{prolong64k}, utilizing the \texttt{prolong-data-64k} corpus while truncating sequences to a maximum length of 32k tokens.
For the Korean dataset, we reconstructed large-scale long-form texts into book-level documents and repacked them to approximately 32k tokens, thereby adapting them for long-context training.

% 또한, 단지 긴 문서만으로 학습할 경우 수학·코드 등 특정 영역 성능이 저하될 수 있다는 관찰이 보고되어 왔으므로 \citep{longred_2025}, 우리는 상대적으로 짧은 고품질 데이터를 병행 학습하는 전략을 도입하였다. 구체적으로, 한국어 합성 데이터셋을 약 32k 길이로 패킹하여 long-context 학습에 포함시켜 장문 맥락 이해와 단문 reasoning 데이터 학습 간 균형을 유지했다 \citep{gao2025_how_to_train_long_context}.  
In addition, prior studies have reported that training only on long documents may degrade performance in domains such as mathematics and code~\citep{longred_2025}. To mitigate this issue, we adopted a mixed training strategy that incorporates relatively short, high-quality data. Specifically, short Korean synthetic datasets were also packed to roughly 32k tokens and integrated into the long-context training corpus to balance the model’s ability between long-context comprehension and short-form reasoning~\citep{gao2025_how_to_train_long_context}.

\begin{table}[h!]
\centering
\begin{adjustbox}{max width=\textwidth}
\begin{tabular}{llcc}
\toprule
\rowcolor{gray!15}
\textbf{Language} & \textbf{Dataset Name} & \textbf{Origin} & \textbf{\# tokens}  \\
\midrule
\multirow{1}{*}{English}
& ProLong & princeton-nlp/prolong-data-64K & 7.21B   \\
\midrule
\multirow{2}{*}{Korean}
& Ko-book     & AI-Hub & 1.11B \\
& Kosmopedia & cosmopedia (English) & 0.51B \\
& Koneedle & Ko-Web Datasets (in Table~\ref{tab:pretrining-datasets-main}) & 1.44B \\

\midrule
\multicolumn{3}{c}{\textbf{English + Korean total Reasoning tokens: 10.27B}}  \\ 
\bottomrule
\end{tabular}
\end{adjustbox}
% \caption{Reasoning Mid Train에서 사용한 한국어/영어 데이터셋 정보}
% \caption{Reasoning Mid Train 단계에서 사용된 한국어 및 영어 데이터셋 정보. 각 데이터셋의 출처(Origin)와 토큰 수(# tokens)를 함께 제시함.}
\caption{Overview of Korean and English datasets used in the Reasoning Mid Train stage, including their sources and token counts.}
\label{tab:mid-train-data-dist}
\end{table}

% 또한, 한국어 장문 QA에서 모델이 한국어에 익숙하지 않아 답변을 영어로 생성하는 문제가 관찰되었다. 이를 개선하기 위해 \emph{Needle in a Haystack}과 유사한 형식의 데이터를 제작하여, 32k 토큰 문서에 무작위 삽입된 정보를 질의로 회수하는 능력을 점검했다. 최종 long-context 확장에 사용한 데이터셋은 영어 7.21B와 한국어 3.06B이다.
Furthermore, during Korean long-form QA tasks, we observed that the model produced responses in English due to limited Korean exposure. To address this issue, we constructed Needle-in-a-Haystack-style data, designed to evaluate and improve the model’s ability to retrieve specific information randomly inserted into 32k-token documents. The final datasets used for long-context extension consisted of 7.21B tokens in English and 3.06B tokens in Korean.

\paragraph{Experimental Results on Long Context Extension}

% KORMo의 장문 문맥 내 정보 회수 능력을 평가하기 위해, 영어–한국어 이중 언어 기반의 Needle In A Haystack (NIAH) 벤치마크를 활용하였다. 본 벤치마크는 최대 127K 토큰 길이의 문서 내 임의 위치에 삽입된 정보를 모델이 정확히 검색할 수 있는지를 측정한다.
To evaluate KORMo’s ability to retrieve information within long contexts, we employed a English–Korean bilingual version of the Needle In A Haystack (NIAH) benchmark. This benchmark measures a model’s capacity to accurately locate and extract specific information randomly inserted within documents of up to 127K tokens in length.

\begin{figure}[t]
\centering
\includegraphics[width=\textwidth]{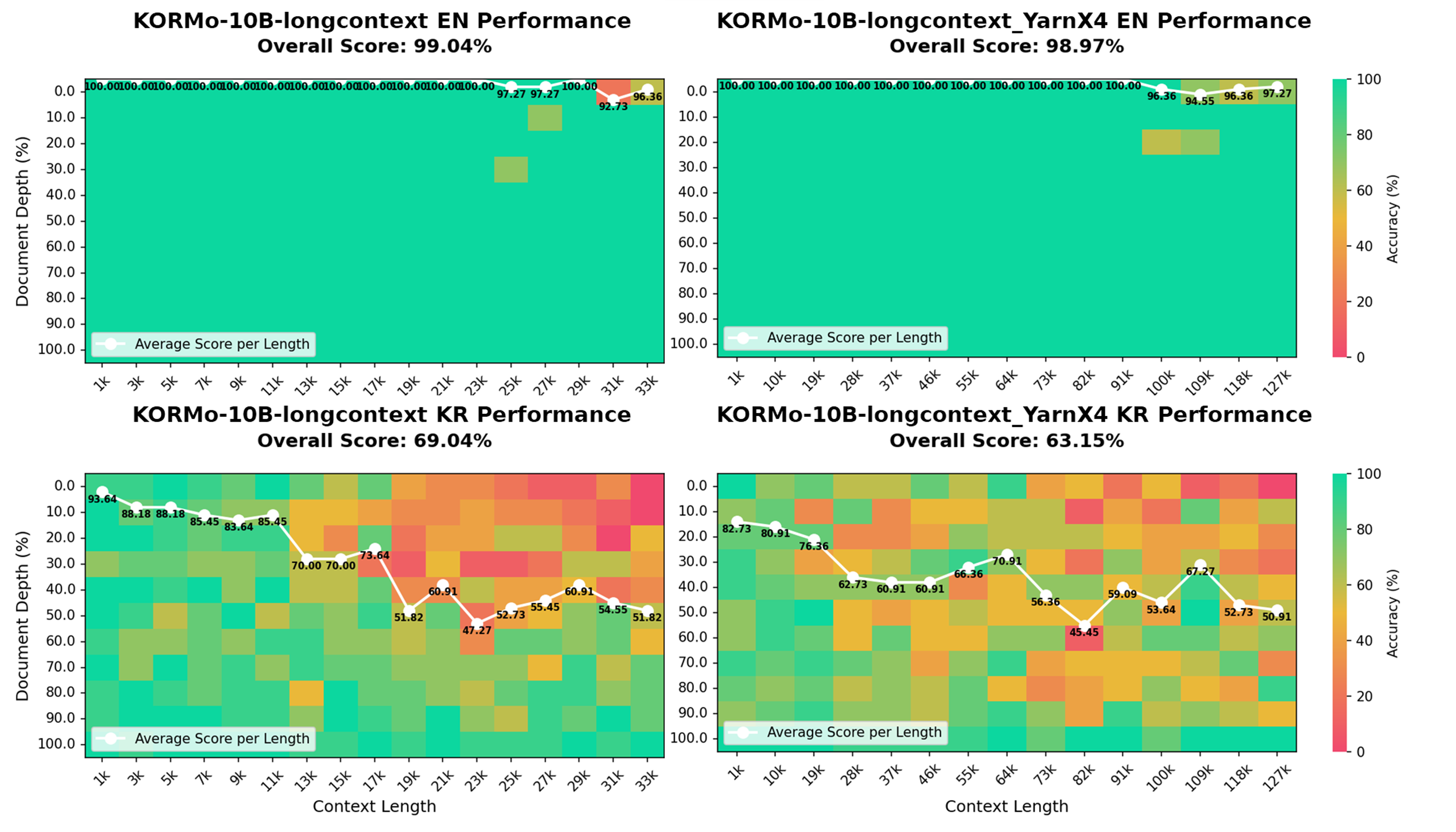}
\caption{Performance of KORMo-10B models on the Needle In A Haystack benchmark for English (top row) and Korean (bottom row). Color intensity denotes retrieval accuracy per document depth and context length.}
\label{fig:NIAH_32k}
\end{figure}

% Figure~\ref{fig:NIAH_32k}는 해당 벤치마크 결과를 나타낸다. KORMo-10B-longcontext 기본 모델은 영어에서 99.04\%, 한국어에서 69.04\%의 정확도를 기록하였으며, YarnX4 기반 컨텍스트 확장을 적용한 모델은 영어에서 98.97\%, 한국어에서 63.15를 기록하였다.
Figure~\ref{fig:NIAH_32k} presents the evaluation results. The baseline KORMo-10B-longcontext model achieved an accuracy of 99.04\% in English and 69.04\% in Korean, while the YarnX4-extended model recorded 98.97\% in English and 63.15\% in Korean.

% 영어의 경우 두 모델 모두 전 구간에서 거의 완벽에 가까운 성능을 유지하였으며, 컨텍스트 길이가 32K 이상으로 확장되어도 안정적인 정보 검색 능력을 보였다. 이는 장문 입력에 대한 높은 강건성과 안정적인 학습 효과를 입증한다.
For English, both models maintained almost perfect performance across all input lengths, demonstrating stable retrieval capability even beyond the 32K context window. This result confirms the robustness and effectiveness of the long-context training procedure.

% 반면 한국어에서는 토큰 길이가 13K를 초과하면서부터 점진적인 성능 저하가 나타났고, 21K 이후 구간에서는 정확도가 60\% 이하로 하락하였다. YarnX4 기반 모델은 일부 길이 구간에서 더 안정적인 성능을 보였으나, 82K–118K 구간에서 성능 저하가 두드러졌다.
In contrast, Korean performance gradually declined beyond the 13K token range, dropping below 60\% accuracy after 21K tokens. Although the YarnX4-based model exhibited greater stability in certain segments, a degradation appeared between 82K and 118K tokens.

% 이러한 결과는 한국어의 경우 장문 문맥에서의 안정적인 정보 검색 능력이 아직 완전히 확보되지 않았음을 시사한다. 이는 토크나이저의 불안정성, 합성 데이터의 불균형, 혹은 장문 데이터셋의 도메인 편향 등이 복합적으로 작용한 결과일 수 있다. 따라서 한국어 장문 학습에 특화된 데이터 설계와 augmentation 전략의 추가적 개선이 요구된다.
These results suggest that for Korean, stable information retrieval capabilities in long contexts have not yet been fully established. This outcome may be attributed to a complex interplay of factors, including tokenizer instability, imbalances in the synthetic data, and domain bias within the long-context dataset. Consequently, further improvements in data design and augmentation strategies specifically tailored for Korean long-context learning are required.

%  (정훈, 민경, 민준)
\subsection{Reasoning Context Training}

% Mid-training의 두 번째 단계인 \textit{Reasoning Context Training}은 대규모 reasoning trace가 포함된 데이터셋을 활용하여 모델의 추론 능력을 강화하는 것을 목표로 한다. 본 단계는 long-context 확장이 완료된 모델 위에서 수행되며, reasoning ability의 효율적 발현과 고급 추론 능력의 내재화를 지향한다~\cite{bakouch2025smollm3, qwen3technicalreport}.

The second phase of mid-training, \textit{Reasoning Context Training}, aims to improve the model’s reasoning ability by training on datasets containing large-scale reasoning traces. This stage is performed after long-context extension and is designed to efficiently elicit reasoning capabilities and help the model internalize advanced inference patterns~\cite{bakouch2025smollm3, qwen3technicalreport}.

\paragraph{Data Preparation.}  
% Table~\ref{tab:mid-train-data-dist}는 본 단계에서 활용된 데이터셋의 분포를 요약한다. 영어 데이터는 Nemotron-Post-v1 데이터셋과 OpenThoughts 1.2M을 포함하였다. 기존 연구는 STEM, Coding, Math 영역의 논리적 reasoning trace의 비중을 높이면 일반적인 모델의 성능 향상에도 기여함을 보고하였으므로(~\cite{bakouch2025smollm3, deepseekai2025deepseekv3technicalreport, qwen3technicalreport}), Nemotron 데이터셋에서 해당 도메인을 우선적으로 선별하고, 좀더 general한 지식의 OpenThoughts 전체 데이터를 반영하였다. 추가로 general reasoning 능력의 보존을 위해 Nemotron의 chat 도메인 중 reasoning이 요구되는 영어 질의-응답 샘플을 더했다.

Table~\ref{tab:mid-train-data-dist} summarizes the dataset composition used in this stage. Prior studies have shown that emphasizing reasoning traces in STEM, coding, and math domains can improve general model performance~\cite{bakouch2025smollm3, deepseekai2025deepseekv3technicalreport, qwen3technicalreport}. Based on this observation, we prioritized STEM, coding, and math domains when constructing the subset from Nemotron-Post-v1, and incorporated the full OpenThoughts dataset to broaden coverage. To further support general reasoning ability, we added QA pairs from the reasoning-relevant portion of Nemotron’s chat domain.

% 한국어의 경우 공개된 대규모 reasoning 데이터셋이 부재하므로, pretraining 단계에서 구축한 Section~\ref{sec:pretrain_dataset}의 Ko-Reasoning 데이터를 재사용하였다. 다만 Ko-Reasoning 데이터의 STEM. Code, Math 비율이 낮기에 추가 데이터가 필요하여 Nemotron 데이터셋 일부를 번역하여 포함했다. 이때 효율적인 reasoning 학습을 위해 난이도 기반 필터링을 수행했으며 이 과정에서 reasoning trace가 실제로 요구되는 고품질 샘플만을 최종적으로 선별하였다.

We reuse the Ko-Reasoning corpus constructed during pretraining (Section~\ref{sec:pretrain_dataset}), as there are no publicly available large-scale reasoning datasets for Korean. Since Ko-Reasoning contains limited coverage of STEM, coding, and math domains, we supplement it by translating selected portions of the Nemotron dataset. To ensure effective reasoning supervision, we apply difficulty-based filtering and retain only high-quality samples that demonstrably require reasoning traces.

\begin{algorithm}[!h]
\caption{The two-stage filtering algorithm for selecting high-difficulty reasoning seeds for translate. Stage 1 selects for samples incorrectly answered by two models, and Stage 2 filters for a consensus on high difficulty.}
\label{alg:mid-train-seed-filtering}
\begin{algorithmic}[1]
\State \textbf{Input:} Annotated dataset $D$ with model answers and difficulty labels
\State \textbf{Output:} Final high-difficulty subset $D'$
\State $P \gets \emptyset$ \Comment{candidate pool}
\State $D' \gets \emptyset$ \Comment{final dataset for translation}
\Statex

\State \textbf{Stage 1: Non-Reasoning Sample Filtering}
\For{$d \in D$}
  \If{\Call{IsIncorrect}{Qwen-30B} $\textbf{and}$ \Call{IsIncorrect}{Qwen-4B}}
    \State $P \gets P \cup \{d\}$ \Comment{Add only samples where both models failed}
  \EndIf
\EndFor
\Statex

\State \textbf{Stage 2: Difficulty Consensus}
\For{$d \in P$}
  \If{\Call{IsHighDifficulty}{Qwen-30B} $\textbf{and}$ \Call{IsHighDifficulty}{Qwen-4B}}
    \State $D' \gets D' \cup \{d\}$ \Comment{Add only samples where both models agree on high difficulty}
  \EndIf
\EndFor
\Statex

\State \textbf{Return:} $D'$
\end{algorithmic}
\end{algorithm}

% 기존의 난이도 판별 알고리즘 내용 작성, 비교하여 적용한 방법의 장점 서술
% 모델을 이용한 난이도 annotation의 근거?

\begin{table}[h!]
\centering
\begin{adjustbox}{max width=\textwidth}
\begin{tabular}{llccc}
\toprule
\rowcolor{gray!15}
\textbf{Language} & \textbf{Dataset Name} & \textbf{Synthesizer} & \textbf{\# tokens} & \textbf{\# num rows} \\
\midrule
\multirow{2}{*}{English}
& Nemotron-Post-Training-Dataset-v1\footnote{\url{nvidia/Nemotron-Post-Training-Dataset-v1}}  & Qwen3-235B \& DeepSeek-R1 & 144.75B  & 24.9M  \\

& OpenThoughts3-1.2M\footnote{\url{https://huggingface.co/datasets/open-thoughts/OpenThoughts3-1.2M}}       & QwQ-32B & 5.46B  & 0.4M \\

\midrule
\multirow{2}{*}{Korean}
& Ko-Reasoning~\ref{tab:pretrining-datasets-main}                                           & Qwen3-235B & 7.05B & 5.8M \\
& Nemotron-Post-Trans & GPT-oss(120B) \& Qwen3-Next-80B & 2.83B & 1.1M \\

\midrule
\multicolumn{5}{c}{\textbf{English + Korean total Reasoning tokens: 157.76B}}  \\ 
\bottomrule
\end{tabular}
\end{adjustbox}
\caption{Table 15: Korean and English Datasets Used in Reasoning Mid-Training}
\label{tab:mid-train-data-dist}
\end{table}
 \label{tab:mid-train-data-dist}

\paragraph{Data Filtering}  
% 알고리즘~\ref{alg:mid-train-seed-filtering}는 영어 Nemotron-Post-V1 데이터로부터 고품질 한국어 번역 seed를 샘플링하기 위한 절차를 정의한다. 두 단계로 구성되며, 각 단계는 reasoning 난이도를 정밀하게 반영하기 위해 설계되었다.  
Algorithm~\ref{alg:mid-train-seed-filtering} outlines the procedure for sampling high-difficulty Korean translation seeds from the English Nemotron-Post-V1 dataset. The process consists of two stages, each designed to assess the reasoning difficulty of the samples.

\begin{enumerate}
    % \item \textbf{Non-Reasoning Sample Filtering:} Qwen-30B와 Qwen-4B가 동시에 오답을 생성한 샘플만을 선별하였다. 이는 복잡한 reasoning을 요구하는 샘플일수록 instruction 모델이 정답을 맞히기 어렵다는 직관에 기반한다. 따라서 본 단계에서 확보된 샘플은 정밀한 reasoning trace를 필요로 하며, 학습에 활용할 경우 추론 능력 강화를 직접적으로 유도할 수 있다. 
    \item \textbf{Non-Reasoning Sample Filtering:} We first selected only the samples that were answered incorrectly by both Qwen-30B and Qwen-4B. This approach is based on the intuition that samples demanding complex reasoning are more challenging for instruction-tuned models to solve correctly. Consequently, the samples retained in this stage require detailed reasoning traces and can directly enhance reasoning capability when used for training.

    % \item \textbf{Difficulty Consensus Filtering:} 1단계 후보군 중에서, Qwen-30B와 Qwen-4B가 모두 \textit{high-difficulty}로 태깅한 샘플만을 최종 데이터셋에 포함하였다. 기존 연구에서는 reasoning path의 길이를 난이도의 지표로 활용하는 경우가 많았으나, deepseek 계열 모델은 self-verification 절차를 빈번히 포함해 단순 길이 기반 지표의 신뢰도가 낮다는 단점이 있다~\cite{jung2025reasoning, peng2025revisiting}. 또한 길이 편향은 불필요한 반복 양식 학습이나 비명료한 추론 구조를 야기할 수 있다. 따라서 본 연구는 모델이 직접 산출한 난이도 태그를 사용하고, 두 모델 간 consensus가 형성된 고난도 샘플만을 최종적으로 채택하였다.  
    \item \textbf{Difficulty Consensus Filtering:}
    From the Stage 1 candidates, only samples that both Qwen-30B and Qwen-4B labeled as \textit{high-difficulty} were included in the final dataset. While prior studies have often used the length of the reasoning path as a proxy for task difficulty, this metric becomes unreliable for DeepSeek-family models, which frequently incorporate self-verification steps within their reasoning traces~\cite{jung2025reasoning, peng2025revisiting}. Moreover, length-based bias can lead to undesirable learning behaviors, such as redundant reasoning patterns or unclear logical structures. To mitigate these issues, we instead utilize model-generated difficulty tags and retain only those samples where both models exhibit a consistent consensus on high difficulty, thereby ensuring the inclusion of truly challenging reasoning data.
    
\end{enumerate}

% 필터링 된 domain 별 비율 추가

\begin{figure}[htbp]
    \centering
    \makebox[\textwidth][c]{%
        \includegraphics[width=\textwidth]{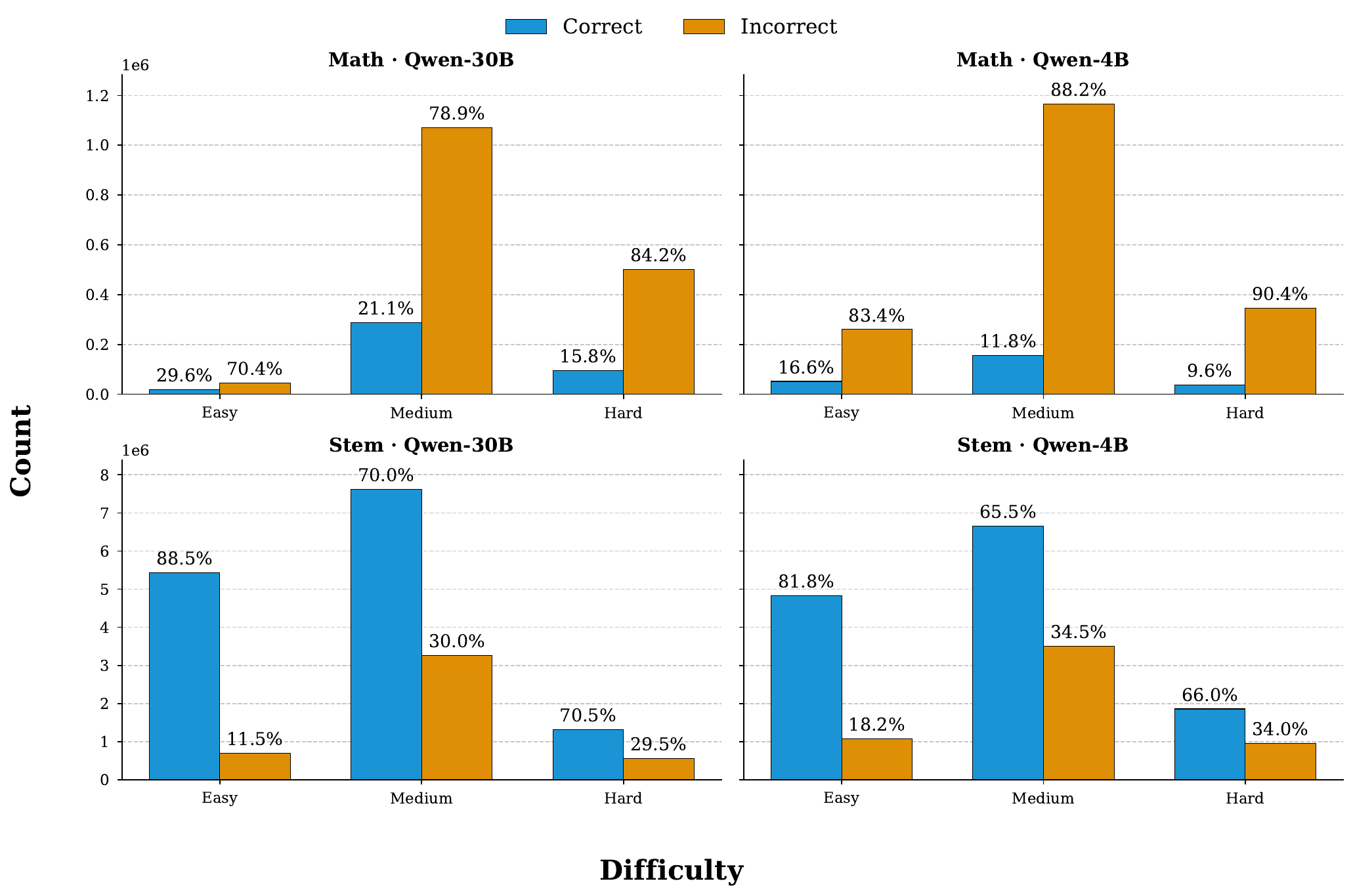}
    }
    % \caption{Comparison of difficulty-level and error-rate distributions between two models on Math and STEM Domains. 모델이 easy, medium, hard로 예측한 샘플들 사이의 정답률 비교 그래프. 파란 색은 맞춘 샘플, 주황 색은 틀린 샘플. stem은 expected answer가 단순한 형태를 띠어 exact match로 평가하였고, math는 expected answer과 모델의 답변을 qwen-4b 모델로 일치하는지 평가했다. 정답의 범위가 좁은 stem에서 정답률이 높음. 코딩, chat 도메인은 특성상 정답 판별이 어려워 stage 1 filtering을 적용하지 않았다}
    % \caption{Math 및 STEM 도메인에서 두 모델이 예측한 난이도 수준(Easy, Medium, Hard) 별 정답률 분포 비교. 각 막대는 난이도 그룹 내에서의 정답(파란색)과 오답(주황색) 비율을 나타낸다. STEM 도메인은 expected answer가 단순한 형태를 가지므로 exact match로 정답 여부를 판별했다. 반면 Math 도메인은 Qwen-4B 모델을 이용해 expected answer과 모델 출력의 일치 여부를 평가하였다. 정답의 범위가 좁은 STEM에서 정답률이 더 높게 나타난다. Code, Chat 도메인은 도메인 특성상 정답 판별이 어려워 첫 번째 단계의 필터링 대상에서 제외하였다.}
    \caption{Comparison of accuracy distributions across difficulty levels (Easy, Medium, Hard) as predicted by the two models in the Math and STEM domains. Each bar represents the proportion of correct (blue) and incorrect (orange) samples within each difficulty group. For the STEM domain, correctness was determined using exact match evaluation, given the relatively simple structure of the expected answers. In contrast, for the Math domain, correctness was assessed using Qwen-4B to verify consistency between the model’s output and the expected answer. Due to its narrower answer space, the STEM domain exhibits higher overall accuracy. The Code and Chat domains were excluded from the stage 1 filtering process, as objective correctness evaluation is inherently difficult in these domains.}
    \label{fig:mid-train-filtering-dist}
\end{figure}

% code 7\% (1.9M $\rightarrow$ 135k), math 11\% (2.04M $\rightarrow$ 226k), stem 3\% (20.7M $\rightarrow$ 596k)
% code 도메인은 190만 개 샘플 중 약 4.8만 개(2.5\%)만 남아 97.5\%가 감소하였고, math 도메인은 204만 개 중 21.7만 개(11\%)로 89\% 감소, stem 도메인은 2,066만 개 중 19.6만 개(1\%)로 99\% 감소하였다. math와 stem에서의 세부 난이도 분포는 Figure 10에 제시하였다. 반면 chat과 tool calling 도메인에서는 정답 라벨이 부재하여, 별도의 난이도 기반 샘플링을 수행하지 않았다.

In the code domain, only 48K samples (2.5\%) out of 1.90M were retained, resulting in a 97.5\% reduction. The math domain decreased from 2.04M to 217K samples (11.0\%), and the STEM domain from 20.66M to 196K samples (1.0\%), reflecting a 99.0\% reduction. Detailed difficulty distributions for math and STEM are shown in Figure 10. For the chat and tool-calling domains, difficulty-based sampling was not applied due to the absence of ground-truth labels.

% 사실 tiktoken cl100kbase로 재면 2.8B, o200kbase로 재면 2.03B 나옴 (?)
% 앞서 선정된 샘플들을 바탕으로, 한국어 데이터 생성을 위해 번역 작업을 다음과 같이 수행하였다. 우선 다양성을 확보하기 위하여 GPT-OSS-120B와 Qwen3-Next-80B-A3B-Instruct를 각각 50\% 비율로 사용하였다. 특히 chat과 code 도메인의 경우 Nemotron에서 원래 user query를 외부 소스(lmsys-chat-lm, apps, code contests, open-r1/codeforces, taco)에서 수집하였기 때문에, metadata를 활용하여 빈 query를 보완한 뒤 번역을 진행하였다.

% 번역 시 LaTeX 표현식, 변수명, 함수명 등 기술적 요소를 번역하지 않도록 하였으며, 원문의 톤(formal/casual)과 문체를 최대한 보존하였다. 또한 code 도메인의 경우 한국 프로그래밍 커뮤니티의 표준 용어를 적용하였고, tool calling 샘플에서는 단순 번역만 수행하여 AI agent 역할을 하지 않도록 유의하였다. 최종적으로 tiktoken (o200base)을 기준으로 계산했을 때, 한국어 번역 데이터는 약 2.8B 토큰 규모로 산출되었다.  

% 최종적으로 선별된 데이터셋은 다단계 추론 과정과 복합적인 논리 전개를 포함하고 있어, KORMo 모델이 reasoning ability를 보다 효율적으로 내재화하고 advanced reasoning capability로 확장할 수 있는 기반을 제공한다. 한편, 영어 데이터의 경우 난이도 분포 전체(쉬운 샘플부터 고난도 샘플까지)를 포함하여 별도의 필터링 없이 전체 데이터셋을 활용하였다.

Based on the selected samples, we constructed the Korean dataset through the following translation procedure. To ensure diversity, we used GPT-OSS-120B and Qwen3-Next-80B-A3B-Instruct in equal proportion (50\% each). For the chat and code domains, since the original user queries in Nemotron were sourced from external datasets (lmsys-chat-lm, apps, code contests, open-r1/codeforces, taco), we supplemented incomplete queries using metadata prior to translation.

During translation, we preserved LaTeX expressions, variable names, and function names without modification, and maintained the tone (formal/casual) and style of the original text as closely as possible. For the code domain, we adopted terminology commonly used in Korean programming communities. In the case of tool-calling samples, translations were conducted strictly without simulating agent behavior.

The resulting Korean dataset amounts to approximately 2.8B tokens, measured using tiktoken with the \texttt{o200k\_base} vocabulary.

The final dataset features multi-step reasoning processes and complex logical structures, providing a foundation for the KORMo model to internalize reasoning ability more effectively and to generalize toward advanced reasoning capabilities. For English data, we used the full dataset without additional filtering, covering the full spectrum of difficulty from simple to challenging samples.

\paragraph{Training Details}  
% 앞서 제안한 필터링 및 번역 과정을 거쳐 최종적으로 영어 150B, 한국어 10B 규모의 reasoning 데이터셋을 확보하였다. 해당 데이터는 모두 \texttt{<think>} 토큰 안에 reasoning trace가 포함된 형식으로 구성되었다. 단, 본 단계에서는 \texttt{<think>} 토큰을 에러 계산에서 제외하였다. 이는 추후 SFT 단계에서 \texttt{<think>} 토큰을 추론 시작 신호로 학습시킬 예정이므로, 본 단계에서 이를 모델링 대상으로 포함할 경우 instruction tuning 단계에서의 의미와 혼선을 일으킬 가능성을 방지하기 위함이다.  

Following the proposed filtering and translation procedures, we obtained a final reasoning dataset consisting of 150B English tokens and 10B Korean tokens. Each sample was formatted to enclose the reasoning trace within the \texttt{<think>} and \texttt{</think>} tokens.

During this stage, however, the \texttt{<think>} token was excluded from loss computation. This design choice prevents semantic interference in the subsequent SFT stage, where the model will be explicitly trained to treat the \texttt{<think>} token as a reasoning trigger signal.

\begin{table}[h!]
\centering
\begin{adjustbox}{max width=\textwidth}
\begin{tabular}{lrrrrrrrrr}
\toprule
Model & agieval\_en & arc\_challenge & arc\_easy & boolq & copa & winogrande & hellaswag & gpqa\_main & Avg. \\
\midrule
Stage1        & 21.78 & 54.44 & 82.53 & 77.34 & 92.00 & 72.85 & 59.05 & 25.22 & 60.15 \\
Stage2        & 28.12 & \textbf{59.90} & 85.65 & 83.88 & \textbf{95.00} & \textbf{74.98} & 59.79 & 29.46 & 64.60 \\
Midtrain(Long)       & 27.60 & 59.22 & \textbf{86.11} & \textbf{84.04} & \textbf{95.00} & 74.82 & 60.11 & 27.68 & \textbf{64.57} \\
Midtrain(Reason)    & \textbf{28.84} & 58.96 & 85.48 & 83.46 & 93.00 & 74.03 & \textbf{60.25} & \textbf{30.13} & 64.27 \\
\bottomrule
\end{tabular}
\end{adjustbox}

\caption{Benchmark performance on major English tasks across training stages. }
\label{tab:eval-midtrain-en}
\end{table}

\begin{table}[h!]
\centering
\begin{adjustbox}{max width=\textwidth}
\begin{tabular}{lrrrrrrrrr}
\toprule
Model & mmlu & mmlu\_global\_en & mmlu\_pro & mmlu\_redux & openbookqa & piqa & social\_iqa & commonsense\_qa & Avg. \\
\midrule
Stage1        & 56.89 & 52.54 & 20.92 & 58.09 & 38.60 & 80.30 & 52.15 & 66.42 & 53.74 \\
Stage2        & 65.35 & 61.71 & 31.32 & 66.68 & 38.00 & \textbf{81.23} & 51.89 & 72.15 & 58.79 \\
Midtrain(Long)       & 63.67 & 61.80 & 31.47 & 65.68 & \textbf{39.20} & 80.85 & 52.15 & \textbf{72.81} & 58.70 \\
Midtrain(Reason)    & \textbf{67.96} & \textbf{63.44} & \textbf{40.18} & \textbf{69.00} & 38.00 & 81.12 & \textbf{52.81} & 72.24 & \textbf{60.34} \\
\bottomrule
\end{tabular}
\end{adjustbox}
% \caption{각 스테이지 별 영어에 대한 QA 및 Reasoning 분야 주요 벤치마크 성능 (가로 = Benchmark, 세로 = Model). 각 열의 최고 성능을 굵게 표시했으며, 마지막 열은 평균 성능을 의미함.}
\caption{Benchmark performance on major English QA and reasoning tasks across training stages.}
\label{tab:eval-midtrain-en2}
\end{table}

\subsubsection{Experiment over Mid-training}
% \paragraph{영어 벤치마크 성능 비교.}
% Table 16–17은 KORMo 모델이 각 스테이지를 거치며 영어 벤치마크에서 어떠한 변화를 보였는지를 요약한다. 먼저, Stage2 단계에서 가장 큰 폭의 성능 향상이 관찰되었다. 평균적으로 Table 16의 종합 지표는 Stage1 대비 +4.45pt(60.15→64.60), Table 17의 Reasoning 계열은 +5.05pt(53.74→58.79) 향상되었다. 이는 ARC-Challenge, BoolQ, MMLU 계열 등 폭넓은 과제에서 동시에 개선이 나타난 결과로, 전반적 일반화 능력 확보의 효과를 시사한다.

\paragraph{Comparison of English Benchmark Performance}
Tables~\ref{tab:eval-midtrain-en} and~\ref{tab:eval-midtrain-en2} summarize how the KORMo model’s performance on English benchmarks evolved across training stages. The most significant improvement was observed at the Stage 2 phase. On average, the aggregate metrics in Table 16 improved by +4.45 pt (60.15→64.60) compared to Stage 1, and the reasoning-focused benchmarks in Table 17 saw a +5.05-point gain (53.74→58.79). These gains span a broad range of tasks, including ARC-Challenge, BoolQ, and the MMLU series, suggesting that Stage 2 contributes to enhancing the model’s generalization capabilities.

% \paragraph{긴 문맥 노출의 효과.}
% Midtrain(Long) 단계에서는 긴 맥락과 서사적 입력에 대한 노출을 통해 선택형 QA와 상식 기반 판별 과제에서 이득을 보였다. ARC-Easy(86.11), BoolQ(84.04), COPA(95.0), CommonsenseQA(72.81) 등은 열 최고 성능을 달성하였다. 그러나 ARC-Challenge, Winogrande, GPQA-main과 같이 추론적 난도가 높은 과제에서는 Stage2 대비 소폭의 성능 저하가 발생하였다. 이는 장문 문맥 신호가 상식적·표면적 패턴에는 유리하지만, 정밀한 역추론이나 편향 제어에는 다소 한계를 가짐을 보여준다.

\paragraph{Impact of Long-Context Training}
The Midtrain(Long) stage benefited tasks such as multiple-choice QA and commonsense-based classification by exposing the model to longer contexts and narrative-style inputs. It achieved the highest scores across several benchmarks, including ARC-Easy (86.11), BoolQ (84.04), COPA (95.0), and CommonsenseQA (72.81). However, slight performance drops were observed compared to Stage 2 on more reasoning-intensive tasks such as ARC-Challenge, Winogrande, and GPQA-main. This suggests that while long-context signals may be advantageous for commonsense or surface-level patterns, they have limitations in supporting fine-grained inference or bias control.

% \paragraph{추론 중심 전이 학습.}
% Midtrain(Reason) 단계에서는 난도가 높은 추론 중심 과제에서 뚜렷한 개선이 확인되었다. MMLU(67.96), Global-EN(63.44), MMLU-Pro(40.18), Redux(69.00), GPQA-main(30.13) 등은 모두 열 최고 성능을 기록했으며, Social-IQA(52.81) 또한 소폭 개선되었다. 평균적으로 Reasoning 스위트에서 Stage2 대비 +1.55pt 상승(58.79→60.34)을 보여, 추론 특화 데이터 신호가 고난도 문제 해결에 효과적으로 기여함을 검증하였다. 반면 ARC-Challenge나 Winogrande와 같은 일부 과제에서는 소폭 성능 하락이 있었으나, 전체적으로는 Stage2 수준을 유지하였다.

\paragraph{Reasoning-Oriented Transfer Learning}
The Midtrain(Reason) stage led to clear improvements on more challenging, reasoning-centric tasks. Benchmarks such as MMLU(67.96), Global-EN(63.44), MMLU-Pro(40.18), Redux(69.00), and GPQA-main(30.13) all achieved the highest scores in their respective columns, with Social-IQA(52.81) also showing a modest improvement. On average, the Reasoning benchmark suite saw a +1.55pt gain over Stage 2 (58.79→60.34), demonstrating that reasoning-intensive data signals contribute effectively to solving high-difficulty problems. While there was a slight drop in performance on certain tasks like ARC-Challenge and Winogrande, overall performance remained comparable to that of Stage 2.

% \paragraph{분석.}
% Stage2는 전반적인 기반 성능을 크게 끌어올리는 역할을 수행하였고, 이후 Long은 상식·지식 회수 능력 강화, Reason은 난도 높은 추론 능력 보강에 각각 특화된 이득을 제공하였다. 따라서 KORMo의 학습 파이프라인은 (1) Stage2를 통한 광역적 기저 능력 확보, (2) Long을 통한 상식·맥락적 회수 능력 강화, (3) Reason을 통한 정밀 추론 능력 보강이라는 2단계 mid-training 전략이 가장 효과적임을 시사한다. 또한 벤치마크 특성에 따라 최적의 학습 단계가 상이하게 나타나므로, 실제 응용에서는 타겟 도메인에 맞춘 스테이지 선택 또는 멀티-헤드 튜닝 전략이 합리적일 것이다.

\paragraph{Analysis}
Stage2 played a key role in significantly improving the overall base performance, while the subsequent Long and Reason stages provided specialized benefits: the Long stage, which exposes the model to extended contexts and narrative inputs, enhanced commonsense and contextual retrieval ability; the Reason stage, which focuses on high-difficulty reasoning tasks, strengthened precise reasoning ability. Therefore, KORMo’s mid-training pipeline suggests that the most effective approach is a two-phase strategy: (1) securing broad foundational capabilities through Stage 2, (2) enhancing commonsense and contextual retrieval ability through Long, and (3) reinforcing precise reasoning ability through Reason. Additionally, since the optimal training stage may vary depending on the target task characteristics, a stage selection strategy aligned with the task domain or a multi-head tuning approach would be more effective in practical applications.

\begin{table}[t]
\centering
\begin{adjustbox}{max width=\textwidth}
\begin{tabular}{lrrrrrrrrrrrr}
\toprule
Model & click & csatqa & haerae & k2\_eval & kobest & kobalt & kmmlu & kmmlu\_p & kmmlu\_r & clinical\_qa & mmlu\_global & Avg. \\
\midrule
Stage1        & 47.82 & 28.67 & 59.30 & 76.85 & 69.94 & 12.71 & 34.51 & 26.65 & 26.83 & 52.82 & 41.60 & 43.43 \\
Stage2        & 51.43 & 29.33 & 62.79 & 80.79 & 72.42 & 17.29 & 44.38 & 32.28 & 36.22 & 72.92 & 53.61 & 50.31 \\
Midtrain(Long)       & 55.59 & 30.00 & 66.73 & 83.80 & 71.56 & 19.00 & 42.92 & 34.69 & 34.94 & 68.80 & 53.89 & 51.08 \\
Midtrain(Reason)    & 55.29 & 38.00 & 68.29 & 84.49 & 75.05 & 22.86 & \textbf{46.48} & 34.51 & \textbf{37.88} & \textbf{77.32} & \textbf{55.16} & 54.12 \\
\bottomrule
\end{tabular}
\end{adjustbox}
% \caption{한국어 벤치마크에서의 모델별 성능 (가로 = Benchmark, 세로 = Model). 각 열의 최고 성능은 굵게 표시했으며, 마지막 열은 11개 벤치마크의 단순 평균이다.}
\caption{Model performance on Korean benchmarks.}
\label{tab:korean_benchmarks}
\end{table}

% \paragraph{한국어 벤치마크 성능 비교.} Table 18의 결과를 종합하면, KORMo는 학습 단계가 진행됨에 따라 한국어 벤치마크 전반에서 꾸준한 성능 향상을 보였다. Stage2에서 평균 성능이 Stage1 대비 +6.88pt(43.43→50.31) 상승하며 가장 큰 기저 능력 개선이 확인되었고, 이는 특히 Clinical-QA(72.92), MMLU-Global(53.61), K2-Eval(80.79) 등 지식 기반 및 독해 중심 과제에서 두드러졌다. 이어서 Midtrain(Long)은 평균 성능을 소폭 높였으며, Click(55.59), Haerae(66.73) 등 문맥 의존적 과제에서 개선이 두드러졌다. 마지막으로 Midtrain(Reason)은 평균 54.12로 전체 단계 중 최고 성능을 기록하며, KMMLU(46.48), KMMLU-R(37.88), Clinical-QA(77.32), MMLU-Global(55.16) 등 난도가 높은 추론 및 지식 집약형 과제에서 강력한 개선을 보여주었다. 요약하면, Stage2는 전반적 기반 성능을 확보하고, Long은 문맥·독해 과제를 강화하며, Reason은 고난도 추론 성능을 극대화하는 구조적 분화를 보여 한국어 모델 성능 향상 경로가 단계별로 뚜렷하게 구분됨을 확인할 수 있다.

\paragraph{Comparison of Korean Benchmark Performance}
As summarized in Table \ref{tab:korean_benchmarks}, KORMo exhibited steady improvements across Korean benchmarks as training progressed. Stage2 yielded the largest performance gain, with a +6.88pt increase in average score compared to Stage 1 (43.43→50.31), particularly improving knowledge-based and reading comprehension tasks such as Clinical-QA(72.92), MMLU-Global(53.61), and K2-Eval(80.79). Subsequently, Midtrain(Long) yielded a modest improvement in the average performance, with notable gains in context-dependent tasks such as Click(55.59) and Haerae(66.73). Lastly, Midtrain(Reasoning) achieved the highest overall performance with an average of 54.12, showing strong improvements on high-difficulty reasoning and knowledge-intensive tasks such as KMMLU(46.48), KMMLU-Redux(37.88), Clinical-QA(77.32), and MMLU-Global(55.16).
In summary, Stage 2 establishes strong foundational capabilities, Long enhances performance on context and reading comprehension tasks, and Reason maximizes high-level reasoning capabilities—demonstrating a clear functional specialization across stages in the Korean training pipeline.

% \paragraph{Cross-lingual 통합 비교.}
% 흥미롭게도 한국어와 영어 모두에서 스테이지별 성능 변화 패턴은 유사하게 나타났다. Stage2는 전반적인 기반 능력을 가장 크게 끌어올리며, Long 단계는 문맥 의존형 과제(예: ARC-Easy, BoolQ, Click, Haerae 등)에서 두드러진 이득을 제공하였고, Reason 단계는 고난도 추론 과제(MMLU-Pro, GPQA, KMMLU, Clinical-QA 등)에서 최고의 성능을 달성하였다. 그러나 세부적으로는 언어별 차이가 관찰된다. 영어에서는 Long 단계에서 상식적 선택형 QA(CommonsenseQA, OpenBookQA) 개선이 뚜렷한 반면, 한국어에서는 Clinical-QA와 같은 도메인 특화 QA에서 Reason 단계의 성능 향상이 더욱 극적이었다. 이는 한국어 모델이 상대적으로 적은 학습 자원에도 불구하고, 단계별 학습 전략을 통해 언어 독립적 보편 패턴(기저 → 문맥 → 추론)을 재현함과 동시에, 언어·도메인 특수성에 따른 차별적 이득을 발휘할 수 있음을 시사한다. 따라서 본 연구는 비영어권 언어 모델에서도 스테이지 설계가 단순 성능 향상뿐 아니라, 언어별 강·약점을 보완하는 정밀한 조정 도구로 활용될 수 있음을 보여준다.

\paragraph{Cross-lingual Stage-wise Comparison}
Interestingly, the performance trends across training stages appear similar for both Korean and English. Stage2 led to the most substantial improvement in overall foundational abilities; the Long stage provided notable gains on context-dependent tasks (e.g., ARC-Easy, BoolQ, Click, Haerae); and the Reason stage achieved the best results on challenging reasoning benchmarks (e.g., MMLU-Pro, GPQA, KMMLU, Clinical-QA).

However, language-specific differences were also observed. In English, the Long stage showed clear improvements on commonsense multiple-choice QA tasks (CommonsenseQA, OpenBookQA), whereas in Korean, the Reason stage led to more dramatic improvements on domain-specific QA tasks such as Clinical-QA.

This suggests that, despite relatively limited training resources, the Korean model was able to reproduce a universal learning progression (Core → Context → Reasoning) through staged training, while also achieving language- and domain-specific improvements.

Therefore, this study demonstrates that stage-wise design can serve not only to boost raw performance in non-English models but also as a fine-tuning tool to compensate for language-specific strengths and weaknesses.
\section{Post-training}

In this section, we present the \textit{post-training} procedure, the final stage of KORMo's training pipeline. This stage consists of two main components: \textit{Supervised Fine-tuning (SFT)} and a preference learning-based fine-tuning process.

\subsection{Supervised Fine-tuning}

The SFT stage aims to fine-tune the language model so that it can faithfully understand and execute user instructions. Instead of constructing a new SFT dataset from scratch, we assembled the training data by upsampling data generated in previous stages and incorporating publicly available open-source datasets. However, as emphasized in the LIMA study~\citep{zhou2023lima}, the performance of SFT is highly sensitive to data quality, which necessitated a rigorous filtering process to ensure the use of high-quality data. To further structure the training, we divided the SFT stage into two phases: \textbf{Base SFT}, which focuses on enhancing general reasoning and language capabilities, and \textbf{Instruction-Following SFT}, which emphasizes consistent formatting and faithful adherence to user instructions.

\begin{table}[h!]
\centering
\begin{adjustbox}{max width=\textwidth}
\begin{tabular}{l l r c l l}
\toprule
\rowcolor{gray!15}
\textbf{Language} & \textbf{Dataset Name} & \textbf{\# tokens (M)} & \textbf{Reasoning} & \textbf{Source (Seed)} & \textbf{Synthesizer}\\
\midrule
\multicolumn{6}{c}{\textbf{\emph{Base SFT} (6.49B tokens)}}  \\ 
\midrule
English      & smoltalk\_conversations           & 0.43M   & \xmark & HF-ST2~\citep{bakouch2025smollm3}  & -- \\
English      & smoltalk\_system\_chats           & 19.4M   & \xmark & HF-ST2~\citep{bakouch2025smollm3}  & -- \\
English      & smolagents\_toolcalling           & 64.6M   & \cmark & HF-ST2~\citep{bakouch2025smollm3}  & -- \\
English      & nemotron\_chat                    & 528.7M  & \cmark & NPT-v1~\citep{NemotronPostTrainingDatasetV1}   & -- \\
English      & nemotron\_code                    & 714.3M  & \cmark & NPT-v1~\citep{NemotronPostTrainingDatasetV1}   & -- \\
English      & nemotron\_math                    & 1029.1M & \cmark & NPT-v1~\citep{NemotronPostTrainingDatasetV1}   & -- \\
English      & nemotron\_stem                    & 576.3M  & \cmark & NPT-v1~\citep{NemotronPostTrainingDatasetV1}   & -- \\
Korean       & \underline{Ko-Reasoning}          & 3377.7M & \cmark & NPT-v1~\citep{NemotronPostTrainingDatasetV1}   & Qwen3-235B-A22B\\
Multilingual & smoltalk\_multilingual            & 175.1M  & \cmark & HF-ST2~\citep{bakouch2025smollm3}         & --\\
\midrule
\multicolumn{6}{c}{\textbf{\emph{Instruction Following SFT} (1.53B tokens)}}  \\ 
\midrule
English     & \underline{IF-math}     & 95.4M   & \xmark & MathInst~\citep{yue2023mammoth}    & Qwne3-Next-80B-A3B-Instruct \\
English     & \underline{IF-math-R}   & 24.2M   & \cmark & MathInst~\citep{yue2023mammoth}    & Qwne3-Next-80B-A3B-Thinking \\
English     & \underline{IF-stem}     & 18.4M   & \xmark & MoT\_science~\citep{openr1} & Qwne3-Next-80B-A3B-Instruct \\
English     & \underline{IF-stem-R}   & 7.3M    & \cmark & MoT\_science~\citep{openr1} & Qwne3-Next-80B-A3B-Thinking \\
English     & \underline{IF-med}      & 39.3M   & \xmark & PubMed-QA~\citep{jin2019pubmedqa} & Qwne3-Next-80B-A3B-Instruct \\
English     & \underline{IF-med-R}    & 12.9M & \cmark & PubMed-QA~\citep{jin2019pubmedqa} & Qwne3-Next-80B-A3B-Thinking \\
English     & WDRM~\citep{boizard2025wdrm} & 21.9M   & \xmark & NPT-v1~\citep{NemotronPostTrainingDatasetV1} & Qwen3-235B-A22B \\
English     & WDRM-R~\citep{boizard2025wdrm} & 347M & \cmark & NPT-v1~\citep{NemotronPostTrainingDatasetV1}    & Qwen3-235B-A22B \\
English     & \underline{MAGPIE}      & 474.2M  & \xmark & HF-ST2~\citep{bakouch2025smollm3}    & Qwne3-Next-80B-A3B-Instruct \\
English     & \underline{Magpie-R}    & 39.2M & \cmark & HF-ST2~\citep{bakouch2025smollm3}    & Qwne3-Next-80B-A3B-Thinking  \\
\hdashline
Korean     & \underline{IF-math}      & 31.7M   & \xmark & MathInst~\citep{yue2023mammoth}    & Qwne3-Next-80B-A3B-Instruct \\
Korean     & \underline{IF-stem}      & 6.7M    & \xmark & MoT\_science~\citep{openr1} & Qwne3-Next-80B-A3B-Instruct \\
Korean     & \underline{IF-med}       & 22.6M   & \xmark & PubMed-QA~\citep{jin2019pubmedqa}   & Qwne3-Next-80B-A3B-Instruct \\
Korean     & \underline{Magpie}       & 321.5M  & \xmark & HF-ST2~\citep{bakouch2025smollm3}      & Qwne3-Next-80B-A3B-Instruct \\
Korean     & \underline{Magpie-R}     & 36.2M   & \cmark & HF-ST2~\citep{bakouch2025smollm3}     & Qwne3-Next-80B-A3B-Thinking  \\
Korean    & \underline{Ko-Reasoning}  & 31.8M   & \cmark & NPT-v1~\citep{NemotronPostTrainingDatasetV1} & Qwen3-235B-A22B\\

\bottomrule
\end{tabular}
\end{adjustbox}
\caption{Overview of the datasets utilized in the supervised fine-tuning (SFT) stage. 
The table reports token counts, reasoning inclusion flags, and data sources for both English and Korean datasets. 
\textbf{Abbreviations.} HF-ST2 refers to the \texttt{HuggingFaceTB/smoltalk2} dataset, 
NPT-v1 denotes the \texttt{Nemotron Post-Training Dataset v1}, 
NHQ indicates the \texttt{Nemotron-HQ}, 
MoT-science corresponds to the \texttt{Mixture-of-Thought} dataset's science subset, 
and WDRM represents the \texttt{When Does Reasoning Matter} project. 
\underline{Underlined datasets} indicate those \textbf{curated as part of this work}. 
Together, these datasets comprise a total of 8.02B tokens.}
\label{tab:consolidated_token_counts}
\end{table}

\paragraph{Dataset Filtering.}
To secure a high-quality SFT dataset, we applied the following procedures:
\begin{enumerate}
\item \textbf{Deduplication:} We collected data from the Ko-Reasoning dataset and the English Nemotron-Post-Training-Dataset-v1 (chat, code, math, and STEM subsets). Duplicate samples were removed based on the unique identifier (uuid) assigned to each query.

\item \textbf{Difficulty-based Sampling:}
For Korean data, we employed the Qwen3-30B-A3B-Instruct-2507 model as an evaluator to filter out overly simple samples from the Ko-Reasoning dataset. The model solved STEM, math, and code problems, and its predictions were compared against the correct answers. Correctly answered samples were labeled as \textit{reasoning not required}, while incorrect ones were labeled as \textit{reasoning required}. The final dataset was balanced at a 1:1 ratio between the two categories.  
For English data, we sampled from the easy, medium, and hard difficulty levels defined in \autoref{fig:mid-train-filtering-dist}, maintaining a balanced distribution across levels.

\item \textbf{Length Filtering:} To comply with the model’s maximum input length, samples with total sequences exceeding 16,384 tokens were excluded.
\end{enumerate}

Additionally, to ensure broader conversational and functional coverage beyond single-turn question answering, we incorporated data related to \textit{multi-turn} dialogues, \textit{tool calling}, and linguistic diversity from \texttt{HuggingFaceTB/smoltalk2}.

\paragraph{BASE SFT Training Strategy.}
Based on the final dataset, we performed SFT for one epoch to train two model variants:
\begin{enumerate}
\item \textbf{Reasoning-Enhanced Model:}  
This variant is designed to explicitly perform reasoning for all queries. During loss computation, all subsequent tokens---including the reasoning span between the \texttt{<think>} and \texttt{</think>} tokens---were included in the training targets. This setup allows the model to internalize not only the generation of the final answer but also the logical reasoning process leading to it.

\item \textbf{Hybrid Model:}  
This variant allows users to toggle between reasoning and non-reasoning modes. In reasoning mode, the \texttt{<think>} block contains an explicit reasoning trace, while in non-reasoning mode, an empty reasoning block (\texttt{``<think>\textbackslash n\textbackslash n</think>''}) was intentionally inserted during training. Following the previously described difficulty-based sampling strategy, easy samples were assigned to non-reasoning mode and difficult ones to reasoning mode, maintaining a balanced 1:1 ratio overall.
\end{enumerate}

\paragraph{Instruction-Following SFT Training Strategy.}
In this stage, the model was further fine-tuned to enhance its ability to follow user instructions faithfully and to produce well-structured, contextually consistent responses. 
The training primarily focused on three aspects: (1) multi-turn dialogue coherence, (2) instruction compliance, and (3) formatting consistency. 
To achieve this, we constructed instruction-style prompts that required the model to maintain conversational context across turns, interpret and execute user commands accurately, and generate outputs that conformed to predefined response formats (e.g., lists, JSON, Markdown, or structured text). 
Compared to the Base SFT phase, which emphasized reasoning and general linguistic capability, this stage aimed to refine the model’s adherence to task-specific conventions and output structures, thereby improving its usability in real-world interactive scenarios.

During data construction for this phase, particular emphasis was placed on incorporating multi-turn interactions and explicit reasoning paths. 
For Korean data, English queries were first translated into Korean using the \textit{Qwen3-Next-80B-A3B} model. 
The translated Korean queries were then re-instructed to the same model to generate corresponding responses, ensuring natural linguistic alignment between queries and answers. 
To build coherent multi-turn conversations, we adopted the \textit{Magpie}~\citep{xu2024magpie} methodology, in which the model-generated responses were combined with the \texttt{<user>} turn tokens to form extended query-response sequences. 
In addition, to elicit reasoning traces in Korean, a language-specific signal token was inserted at the beginning of the reasoning span, which guided the model to generate reasoning paths entirely in Korean.
% 민준
\subsection{Optimization on Preference Learning}

% 우리는 KORMo의 수학 및 추론능력 향상을위해 APO, ORPO 두 가지 강화학습 데이터를 구축하였다. 이 중 APO 학습을 위해 Qwen3-\{0.6B, 4B, 8B, 30B-A3B, 80B-A3B\} 모델을 통해 Nemotron-Post-Train-V2 데이터의 프롬프트를 활용하여 응답을 생성하였다. 상대적으로 작은 모델을 rejected sample, 큰 모델을 chosen sample로 채택하는 SmolLM3의 방법을 따랐으며, \{chat, stem, code, math\} 도메인에 대하여 총 100K개의 프롬프트를 통해 모델 크기 별로 100K의 응답을 구축하였다. 다만 GPU리소스 부족 문제로 데이터만 제작했으며, 추후 실험 및 학습 결과를 공개하고자 한다. 이와 더불어 추론 활성화 모델도 함께 공개할 예정이다.

We constructed two preference learning datasets based on the APO and ORPO frameworks to enhance KORMo’s mathematical and reasoning capabilities. For APO, we used prompts from the Nemotron-Post-Train-V2 dataset to generate responses with Qwen3 models of various sizes (0.6B, 4B, 8B, 30B-A3B, and 80B-A3B). Following the SmolLM3 strategy, we treated smaller model outputs as rejected samples and larger model outputs as chosen. Using 100K prompts spanning four domains (chat, STEM, code, and math) we produced 100K responses per model size. Due to GPU constraints, we constructed the dataset without proceeding to model training; training and evaluation results will be released in future work.

% 마지막 단계는 강화학습을 통해 모델의 추론 능력과 수치해석 능력을 강화하는 단계이다. 이 단계에서 우리는 두 가지 강화학습 방법을 토대로 대량의 데이터를 생성했다. (1) DPO의 변형인 APO를 이용해 굉장히 쉽고 빠르게 선호도 학습을 했다. (2) ORPO를 적용해 수학능력과 전반적인 추론능력 그리고 답변 생성시 사용자 의도를 반영할 수 있도록 답변포멧을 조정하는 학습을 진행했다. 다만 GPU리소스 문제로 데이터만 제작했으며, 추후 실험 및 학습결과를 공개하고자 한다.

%The final stage aims to improve the model’s reasoning and numerical capabilities through reinforcement learning. To this end, we constructed a large-scale dataset using two RL-based approaches:
%(1) APO, a variant of DPO, was employed for efficient and lightweight preference learning.
%(2) APO was used to enhance mathematical reasoning, general inference ability, and alignment with user intent by refining the answer format.
%While we focused on dataset construction in this phase, model training and evaluation will be conducted in future work, as they were deferred due to current computational limitations.

%하지만, 모델 크기에 따라 선형적으로 답변 길이가 길어진다는 문제점이 있어 학습 과정에서 모델이 단순히 긴 답변을 prefer sample로 선택하는 reward hacking이 발생하는 것을 확인하였다. 
% (민준)
\section{Experiments}

% 앞선 섹션에서 우리는 각 단계별 실험과 성능을 평가했다. 이번장은 조금 더 종합 평가를 토대로 우리가 제안한 방법의 장 단점을 리포트 하려하며 다른 외부 모델들과 비교하려고한다.
In the preceding sections, we evaluated the experiments and performance at each stage. This chapter reports on the strengths and weaknesses of our proposed method based on a more comprehensive evaluation and compares it with other external models.

\subsection{Experiment settings}

\paragraph{Applied Models}
% KORMo는 하이브리드 모델로 추론 모드와 비 추론 모드로 모델을 변환하여 사용할 수 있다. 다만 현재 비교 대상 모델이 비 추론 모델이기 때문에 비추론 모드로 평가를 진행했다. 추론모드의 경우 강화학습을 통한 추론 능력 강화가 필수적이기 때문에 추 후 추론능력을 강화한 모델의 평가 또한 제공할 예정이다.
As a hybrid model, KORMo can function in two distinct modes: reasoning and non-reasoning. We performed the evaluation in the non-reasoning mode to maintain consistency with the benchmark models, which operate in a non-reasoning capacity. The reasoning mode's abilities must be strengthened through reinforcement learning, and accordingly, an assessment of the enhanced model will be presented in subsequent work.

\paragraph{Benchmarks}
%%KORMo 모델의 성능을 다각도로 평가하기 위해, 우리는 영어 및 한국어 각각에 대해 \textit{reasoning}, \textit{knowledge}, 그리고 \textit{domain-specific} 영역을 포괄하는 총 26여 개의 벤치마크를 사용하였다 (Table~\ref{tab:benchmarks_reference}).
To comprehensively evaluate the performance of the KORMo model, we utilized a total of over 26 benchmarks for both English and Korean, covering the domains of \textit{reasoning}, \textit{knowledge}, and \textit{domain-specific} tasks (Table~\ref{tab:benchmarks_reference}).

%먼저, \textbf{영어 일반 추론(General Reasoning)} 부문에서는 AGIEval, ARC-Challenge/Easy, BoolQ, CommonSenseQA, COPA \cite{roemmele2011copa}, HellaSwag, PIQA, Social-IQA, WinoGrande, 그리고 OpenBookQA를 포함하여 언어모델의 일반적 논리 추론과 상식적 판단 능력을 평가하였다. 
%\textbf{영어 지식 및 시험 기반(Knowledge \& Exam-based)} 영역에서는 MMLU를 중심으로, 확장 버전인 MMLU-Global, MMLU-Pro, MMLU-Redux, 그리고 과학적 난이도를 평가하는 GPQA-Main을 활용하여 전문지식과 학문적 문제 해결 능력을 검증하였다.
%한편, \textbf{한국어 일반 추론(General Reasoning in Korean)}에서는 CLICK, CSATQA, HAERAE, K2Eval, KoBEST, KoBALT등 최근 공개된 한국어 추론 중심 벤치마크를 포괄하여 언어모델의 문화적·언어적 적응 능력을 측정하였다.
%마지막으로, \textbf{한국어 지식 및 도메인 특화(Knowledge \& Domain-specific in Korean)} 평가에서는 KMMLU, KMMLU-Pro, KMMLU-Redux, KR-Clinical-QA, 그리고 MMLU-Global (KR)을 사용하였다. 이를 통해 모델이 한국어 기반의 전문지식과 도메인별 문제(의학, 법학 등)를 얼마나 잘 이해하고 응답하는지를 평가하였다. 이 중 KR-Clinical-QA는 가장 최근 나온 벤치마크로 이미 공개된 벤치마크의 경우 테스트 데이터가 학습과정에 활용 (contamination)될 가능성이 있었기에 객관적인 모델 성능 평가를 위해 특별히 추가했다. KR-Clinical-QA, 그리고 MMLU-Global (KR)을 사용하였다. 이를 통해 모델이 한국어 기반의 전문지식과 도메인별 문제(의학, 법학 등)를 얼마나 잘 이해하고 응답하는지를 평가하였다. 이 중 KR-Clinical-QA는 가장 최근 나온 벤치마크로 이미 공개된 벤치마크의 경우 테스트 데이터가 학습과정에 활용 (contamination)될 가능성이 있었기에 객관적인 모델 성능 평가를 위해 특별히 추가했다. 
%모든 벤치마크는 5-shot (GPQA-Main의 경우 4-shot) prompting 설정을 사용하였으며, 정확도(accuracy)를 공통 지표로 하여 성능을 비교하였다. 이러한 구성은 KORMo가 다언어·다도메인 환경에서 얼마나 균형 잡힌 reasoning과 knowledge competence를 확보했는지를 검증하기 위한 것이다.

%마지막으로, instruction 모델의 경우 모든 모델을 동일하게 instruction tuning이 완료된 모델을 활용했다. 실제 Instruction Following 능력을 평가하기 위해 대표적인 LLM-as-Judge 벤치마크 세 가지를 활용했으며, 이때 GPT-4o를 동일한 평가모델로 활용했다. 

First, in the \textbf{English General Reasoning} category, we evaluate logical reasoning and commonsense inference capabilities using AGIEval, ARC-Challenge/Easy, BoolQ, CommonSenseQA, COPA, HellaSwag, PIQA, Social-IQA, WinoGrande, and OpenBookQA.

The \textbf{English Knowledge \& Exam-based} category focuses on assessing domain-specific and academic problem-solving skills using MMLU and its extended variants—MMLU-Global, MMLU-Pro, and MMLU-Redux—along with GPQA-Main, which measures performance on advanced scientific questions.

For \textbf{Korean General Reasoning}, we employ recently released Korean benchmarks such as CLICK, CSATQA, HAERAE, K2Eval, KoBEST, and KoBALT to measure cultural and linguistic adaptability of the models.

Finally, the \textbf{Korean Knowledge \& Domain-specific} evaluation includes KMMLU, KMMLU-Pro, KMMLU-Redux, KR-Clinical-QA, and MMLU-Global (KR). These benchmarks assess the models’ ability to comprehend and respond to Korean-domain expert questions in fields such as medicine and law. Among them, KR-Clinical-QA is the most recently released benchmark and is included to mitigate potential data contamination that might occur with earlier public datasets, ensuring a more objective measurement of model performance.

All benchmarks are evaluated under a 5-shot prompting setup (4-shot for GPQA-Main), and accuracy is used as the unified evaluation metric. This configuration aims to verify whether KORMo achieves balanced reasoning and knowledge competence across multilingual and multi-domain environments.

Lastly, for the \textbf{instruction-tuned models}, we evaluate models that have undergone the same instruction tuning procedure. To assess instruction-following ability, we use three representative \emph{LLM-as-a-Judge} benchmarks, where GPT-4o serves as a consistent evaluation model for all systems.

\begin{table}[t]
\centering
\scriptsize
\renewcommand{\arraystretch}{1.1}
\setlength{\tabcolsep}{4pt}
\begin{tabular}{@{}lllll@{}}
\toprule
\textbf{Name} & \textbf{Language} & \textbf{Prompting} & \textbf{Metric} &  \\ 
\midrule
\rowcolor{gray!10}
\multicolumn{5}{l}{\textbf{General Reasoning (English)}} \\
\midrule
AGIEval \cite{zhong2023agieval} & English & 5-shot & accuracy \\
ARC-Challenge \cite{allenai:arc} & English & 5-shot & accuracy \\
ARC-Easy \cite{allenai:arc} & English & 5-shot & accuracy \\
BoolQ \cite{clark2019boolq} & English & 5-shot & accuracy \\
CommonSenseQA \cite{talmor2019commonsenseqa} & English & 5-shot & accuracy \\
COPA \cite{roemmele2011copa} & English & 5-shot & accuracy \\
HellaSwag \cite{zellers2019hellaswag} & English & 5-shot & accuracy \\
PIQA \cite{bisk2020piqa} & English & 5-shot & accuracy \\
Social-IQA \cite{sap2019socialiqa} & English & 5-shot & accuracy \\
WinoGrande \cite{sakaguchi2021winogrande} & English & 5-shot & accuracy \\
OpenBookQA \cite{obqa}  & English & 5-shot & accuracy \\
\midrule
\rowcolor{gray!10}
\multicolumn{5}{l}{\textbf{Knowledge \& Exam-based (English)}} \\
\midrule
MMLU \cite{hendrycks2021mmlu} & English & 5-shot & accuracy \\
MMLU-Global (EN) \cite{hendrycks2021mmlu} & English & 5-shot & accuracy \\
MMLU-Pro \cite{hendrycks2021mmlu} & English & 5-shot & accuracy \\
MMLU-Redux \cite{hendrycks2021mmlu} & English & 5-shot & accuracy \\
GPQA-Main \cite{rein2023gpqa} & English & 4-shot & accuracy \\
\midrule
\rowcolor{gray!10}
\multicolumn{5}{l}{\textbf{General Reasoning (Korean)}} \\
\midrule
CLICK \cite{kim2024clickbenchmarkdatasetcultural} & Korean & 5-shot & accuracy \\
CSATQA\footnote{\url{huggingface.co/datasets/HAERAE-HUB/csatqa}} & Korean & 5-shot & accuracy \\
HAERAE \cite{son2024haerae} & Korean & 5-shot & accuracy \\
K2Eval\footnote{\url{huggingface.co/datasets/HAERAE-HUB/K2-Eval}} & Korean & 5-shot & accuracy \\
KoBEST \cite{jang2022kobest} & Korean & 5-shot & accuracy \\
KoBALT \cite{shin2025kobaltkoreanbenchmarkadvanced} & Korean & 5-shot & accuracy \\
\midrule
\rowcolor{gray!10}
\multicolumn{5}{l}{\textbf{Knowledge \& Domain-specific (Korean)}} \\
\midrule
KMMLU \cite{son-etal-2025-kmmlu} & Korean & 5-shot & accuracy \\
KMMLU-Pro \cite{hong2025kmmlureduxkmmluproprofessionalkorean} & Korean & 5-shot & accuracy \\
KMMLU-Redux \cite{hong2025kmmlureduxkmmluproprofessionalkorean} & Korean & 5-shot & accuracy \\
KR-Clinical-QA \footnote{\url{huggingface.co/datasets/snuh/ClinicalQA}} & Korean & 5-shot & accuracy \\
MMLU-Global (KR) \cite{singh-etal-2025-global} & Korean & 5-shot & accuracy \\
\midrule
\multicolumn{5}{l}{\textbf{Instruction-Following}} \\
\midrule
%IFEval & English & 5-shot & strict-average \\
MT-Bench & English & LLM-as-Judge(GPT-4o) & LLM score \\
Ko-MT-Bench  & Korean & LLM-as-Judge(GPT-4o) & LLM score \\
LogicKor & Korean & LLM-as-Judge(GPT-4o) & LLM score \\
\bottomrule
\end{tabular}
\caption{\textbf{Evaluation benchmarks used in Table~\ref{tab:vertical_benchmarks}.}  
Each benchmark is categorized by domain and language, with its prompting setup (shot) and evaluation metric.}
\label{tab:benchmarks_reference}
\end{table}

% \paragraph{Evaluation Prompt.} 우리는 Table~\ref{tab:benchmarks_reference} 같은 olmo2에서 제안한 일반적인 평가 prompt를 활용했으며, Instruction Following 능력은 GPT-4o를 활용해 동일한 환경에서 같은 prompt를 활용해 평가했다.
\paragraph{Evaluation Prompt}
We adopted the standard evaluation prompts proposed by OLMo 2 (as shown in Table~\ref{tab:benchmarks_reference}). To assess instruction-following capabilities, we evaluated all models under the same conditions using GPT-4o and the same set of prompts.
\begin{tcolorbox}[colback=black!5!white, colframe=black, title=\textbf{prompt MMLU}, fonttitle=\bfseries]

Q: \texttt{\{question\}}

Put your answer within \texttt{\textbackslash boxed\{\}}.
\end{tcolorbox}
\noindent\begin{minipage}{\textwidth}
\captionsetup{justification=centering}

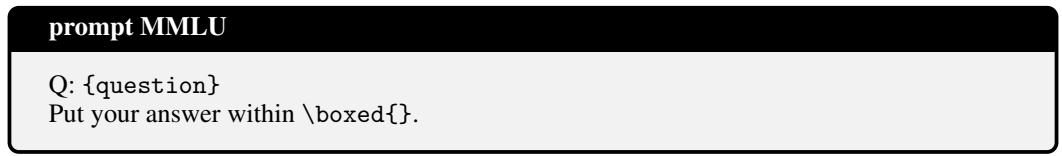
\captionof{figure}{Prompt for evaluating MMLU}
\end{minipage}

% \subsection{Base Model 평가}
% Base 모델 평가는 Instruction tuning 이전 단계에서 모델이 지닌 언어 이해, 추론, 수리 능력의 기본적 잠재력을 검증하기 위한 것이다. 이에 따라 본 연구에서는 Instruction tuning이 적용되지 않은 \textbf{base 모델}만을 대상으로 공정한 비교를 수행하였다. 현재 공개된 최신 다국어 모델 중 base 버전이 존재하는 것은 \textbf{Qwen3}, \textbf{Gemma3}, \textbf{LLaMA3.1}이며, 한국어 모델로는 \textbf{KANANA1.5}가 유일하다. 따라서 본 평가는 이들 모델을 동일한 조건에서 비교함으로써, tuning이나 RLHF 등의 후처리 효과가 배제된 순수한 사전학습 능력을 측정하도록 설계되었다. 한편, \textbf{Fully-Open Model(FOM)} 계열인 \textbf{smolLM3}와 \textbf{OLMo2} 역시 base 버전이 공개되어 있으나, 이들은 영어 및 인도 유러피언 언어로 학습되었다는 점을 함께 고려해야 한다.

\subsection{Base Model Evaluation}
The evaluation of base models aims to assess the foundational capabilities of language understanding, reasoning, and mathematical skills prior to instruction tuning. Accordingly, we compare only the base versions of models, without any instruction tuning applied, to ensure a fair evaluation. Among recently released multilingual models, only Qwen3, Gemma3, and LLaMA3.1 offer publicly available base versions. For Korean, Kanana1.5 is the sole model with a base release. Thus, this evaluation is designed to isolate the effects of pretraining, excluding any influence from instruction tuning or RLHF. Meanwhile, Fully-Open Model(FOMs) such as SmolLM3 and OLMo2 also release base versions; however, these are primarily trained on English and other Indo-European languages.

\begin{table}[t]
\centering
\begin{adjustbox}{max width=\textwidth}
\begin{tabular}{l|rrrr|rrrrr}
\toprule
&\multicolumn{4}{c}{\quad \textbf{\texttt{Fully-Open Model}}} & \multicolumn{5}{c}{\textbf{\texttt{Open-Weight Model (Multilingual)}}} \\
\midrule
\textbf{Benchmark} & kormo-10b & smolLM3-3b & olmo2-7b & olmo2-13b & kanana1.5-8b & qwen3-8b & llama3.1-8b & gemma3-4b & gemma3-12b \\
\textbf{Tokens} & (2.9T) & (10.7T) & (4T) & (5.5T) & (3.2T) & (36T) & (15T) & (4T) & (12T) \\
\midrule
\rowcolor{gray!10}
\multicolumn{10}{l}{\textbf{English Benchmarks}} \\
\midrule
arc\_challenge     & 58.96 & 55.55 & 59.13 & 61.01 & 56.48 & 63.82 & 54.61 & 53.58 & 63.82 \\
arc\_easy          & 85.48 & 83.21 & 85.06 & 86.57 & 82.74 & 87.50 & 84.01 & 82.83 & 87.37 \\
boolq              & 83.46 & 82.17 & 84.50 & 86.48 & 84.53 & 87.71 & 81.87 & 80.70 & 86.61 \\
copa               & 93.00 & 91.00 & 92.00 & 93.00 & 88.00 & 92.00 & 93.00 & 89.00 & 95.00 \\
gpqa\_main         & 30.13 & 26.79 & 26.34 & 29.24 & 29.24 & 30.13 & 23.44 & 30.13 & 35.71 \\
hellaswag          & 60.25 & 56.78 & 61.52 & 65.02 & 59.93 & 59.54 & 60.96 & 57.56 & 63.67 \\
mmlu               & 67.96 & 61.37 & 62.81 & 66.85 & 63.73 & 76.95 & 65.03 & 59.60 & 73.58 \\
mmlu\_global       & 63.44 & 57.52 & 59.88 & 63.99 & 60.21 & 75.05 & 61.30 & 57.23 & 70.23 \\
mmlu\_pro          & 40.18 & 34.94 & 27.29 & 32.50 & 34.93 & 56.58 & 36.23 & 27.79 & 37.07 \\
mmlu\_redux        & 69.00 & 62.95 & 63.53 & 68.37 & 65.88 & 78.19 & 65.86 & 60.86 & 75.25 \\
openbookqa         & 39.00 & 36.40 & 39.00 & 39.60 & 36.80 & 39.20 & 39.00 & 37.00 & 40.20 \\
piqa               & 81.12 & 78.45 & 80.79 & 82.64 & 80.30 & 79.05 & 80.90 & 79.49 & 82.59 \\
social\_iqa        & 52.81 & 50.72 & 55.89 & 57.57 & 57.01 & 56.96 & 53.12 & 51.84 & 56.45 \\
winogrande         & 74.03 & 73.32 & 77.03 & 81.69 & 73.32 & 77.03 & 77.74 & 72.93 & 80.51 \\
\midrule
\textbf{English Avg.}
& \textbf{64.20} & \textbf{60.80} & \textbf{62.48} & \textbf{65.32} & \textbf{62.36} & \textbf{68.55} & \textbf{62.65} & \textbf{60.04} & \textbf{67.72} \\
\midrule
\rowcolor{gray!10}
\multicolumn{10}{l}{\textbf{Korean Benchmarks}} \\
\midrule
click              & 55.29 & 46.97 & 37.79 & 41.80 & 62.76 & 60.70 & 49.22 & 49.62 & 62.21 \\
csatqa             & 38.00 & 26.67 & 19.33 & 24.67 & 44.67 & 52.00 & 28.67 & 28.67 & 31.33 \\
haerae             & 68.29 & 55.82 & 31.62 & 37.58 & 80.75 & 67.19 & 53.25 & 60.68 & 74.34 \\
k2\_eval           & 84.89 & 75.23 & 49.54 & 63.43 & 84.72 & 84.72 & 76.62 & 76.39 & 85.42 \\
kobest             & 75.05 & 69.13 & 57.27 & 59.02 & 81.93 & 80.05 & 70.55 & 69.33 & 77.70 \\
kobalt             & 22.86 & 15.86 & 11.43 & 13.14 & 26.29 & 26.57 & 17.43 & 15.57 & 23.86 \\
kmmlu              & 46.48 & 38.52 & 33.05 & 31.24 & 48.86 & 56.93 & 40.75 & 39.84 & 51.60 \\
mmlu\_global       & 55.16 & 44.15 & 34.00 & 36.95 & 52.65 & 61.95 & 46.34 & 46.33 & 59.68 \\
kr\_clinical\_qa   & 77.32 & 53.97 & 48.33 & 46.22 & 65.84 & 80.00 & 63.54 & 60.00 & 77.22 \\
\midrule
\textbf{Korean Avg.}
& \textbf{58.15} & \textbf{47.37} & \textbf{35.82} & \textbf{39.34} & \textbf{60.94} & \textbf{63.35} & \textbf{49.60} & \textbf{49.60} & \textbf{60.37} \\
\bottomrule
\end{tabular}
\end{adjustbox}
% \caption{모델별 벤치마크 성능 (세로 = Benchmark, 가로 = Model). 언어별 평균은 각 영역의 단순 평균.}
% \caption{Fully-Open 및 Open-Weight(다국어) 언어 모델들의 영어 및 한국어 벤치마크 성능 비교. 각 값은 정확도(\%)를 나타내며, 언어별 평균은 단순 산술 평균으로 계산하였다. 모델명 아래 괄호 안의 숫자는 학습 토큰 수(Trillion 단위)를 의미한다.}
\caption{Performance comparison of Fully-Open and Open-Weight multilingual models on English and Korean benchmarks. All scores represent accuracy, and averages are simple arithmetic means. Numbers in parentheses below model names indicate training tokens (in trillions).
}

\label{tab:vertical_benchmarks}
\end{table}

\paragraph{Overall Trends}
% KORMo-10B는 Instruction tuning 이전의 \emph{base} 모델만을 대상으로 공정 비교했을 때, 전체 평균에서 \textbf{영어 64.2점, 한국어 58.2점}으로 나타나, 동급 Fully-Open 모델 대비 안정적인 성능을 확보했다. 특히 영어에서는 OLMo2-13B 등 대형 공개모델과 유사한 수준 (64.2 vs. 65.3)을 보였으며, 한국어에서는 KANANA-8B, Qwen3-8B, Gemma3-12B 등 다국어 초거대 모델과 비교해 4\%에서 8\% 정도 낮은 성능을 보였다. 다만 이는 제한된 사전학습 토큰(2.9T) 대비 언어적 효율성이 높다는 점을 시사한다.

When evaluated in its \emph{base} form (before instruction tuning), KORMo‑10B scored 64.2 on English and 58.2 on Korean benchmarks. This indicates stable performance compared to other models in the Fully‑Open category. In English, it performed on par with large open models such as OLMo2‑13B (64.2 vs. 65.3), while in Korean, it scored approximately 4–8\% lower than multilingual LLMs including KANANA‑8B, Qwen3‑8B, and Gemma3‑12B. Given its relatively modest pretraining corpus of 2.9T tokens, these results highlight KORMo‑10B’s high language modeling efficiency.

\paragraph{English Benchmarks: Robust Reasoning Ability}
% 영어권 태스크에서는 KORMo가 일반적 추론(ARC, BoolQ, HellaSwag 등)과 상식 기반 판단(PIQA, Social-IQA, WinoGrande)에서 꾸준히 중상위 성능을 보였다. 더 큰 12B, 13B 모델들이 전반적으로 더 높은 평균을 기록했지만, 그 격차는 2\%에서 7\% 수준에 불과하며, 오히려 중형 Fully-Open 모델 대비는 안정적으로 우위를 유지한다. 이는 KORMo가 대형 다국어 모델처럼 광범위한 도메인을 학습하지 않았음에도, 영어 reasoning의 기본적 구조를 효율적으로 내재화했음을 보여준다. 반면 고난도 지식 추론(MMLU-Pro, GPQA)에서는 평균적으로 약간의 성능 저하가 관찰되어, 세밀한 지식 정밀도보다는 일반적 reasoning과 언어적 일관성에 강점을 가지는 것으로 분석된다.

KORMo consistently demonstrated strong performance in general reasoning (ARC, BoolQ, HellaSwag) and commonsense inference tasks (PIQA, Social-IQA, WinoGrande). While larger 12B and 13B models generally achieved higher scores, the margin was relatively narrow (2–7\%). KORMo also consistently outperformed other mid-sized fully open models. This suggests that KORMo was able to acquire the essential patterns of English reasoning, even without exposure to the wide domain coverage typically seen in large multilingual models. On the other hand, it showed slightly lower performance on more knowledge-intensive tasks such as MMLU-Pro and GPQA, suggesting that KORMo’s strength lies more in general reasoning and linguistic consistency than in fine-grained factual precision.

\paragraph{Korean General Reasoning: High Adaptability in Native Language}
% 한국어 일반 추론 과제인 K2-Eval 같은 종합 reasoning 평가에서는 영어-한국어 양쪽에서 모두 균형 잡힌 성능을 유지했다. 이는 한국어 데이터의 질적 구성과 균형 잡힌 사전학습 비율이 효과적으로 작동했음을 의미한다. 다만 KOBALT처럼 어휘적·의미적 미세 변별 능력을 요구하는 태스크에서는 상대적으로 낮은 성능을 보여, 언어적 세부 구문 구조를 강화하는 Post-training이 추가로 필요함을 시사한다. 

% 또한 한국어 성능의 경우 한국어 비율이 높을 수 록 좋은 성능을 보이는것을 확인할 수 있었다. kanana1.5 모델과 비교시 영어의 경우 1.84점 높으나 한국어에서 2.79포인트 뒤떨어지는 결과를 보였다. 이는 kanana의 경우 한국어 비율이 10\%정도로 보고된것으로 미루어 봤을 때 한국어 데이터가 5.6\%정도 포함된 KORMo의 한국어 성능이 더 낮다는걸 확인할 수 있다.

KORMo exhibited balanced performance across both English and Korean in comprehensive reasoning benchmarks such as K2-Eval, a general-purpose Korean reasoning task. This suggests that the quality of Korean data and its proportion in pretraining were effective. However, performance was relatively lower on tasks like KOBALT, which require fine-grained lexical and semantic discrimination. This suggests that additional post-training may be necessary to better capture subtle linguistic distinctions.

We also observed that Korean performance generally improved with a higher proportion of Korean data. Compared to the KANANA-1.5 model, KORMo scored 1.84 points higher in English, but 2.79 points lower in Korean. Considering that KANANA reportedly uses approximately 10\% Korean data, this result reflects the effect of KORMo’s lower Korean data ratio, estimated at around 5.6\%.

\paragraph{Korean Knowledge \& Domain Tasks: Specialized Strengths}
% 도메인 지식 중심의 평가 Clinical-QA에서는 KORMo가 \textbf{임상·실용 영역에서 특히 강세}를 보였다. 이는, 모델이 실제 응답형 QA 형식의 의료·상식 데이터를 학습 중 내재화했음을 보여준다. 반면 KMMLU 계열(학문·고급 지식 기반 문제)에서는 대형 다국어 모델 대비 낮은 평균을 기록해, 전공지식 세밀화보다는 일반적 reasoning 위주의 사전학습 편향이 존재함을 알 수 있다. 이는 고품질 전문데이터나 설명형(explanatory) QA를 활용한 지식이 부족한것을 의미하는데, 증강데이터로 구성된 한국어 데이터 중 고급 전문지식 분야의 생성에 한계가 있었을것으로 추축된다.

On the Clinical-QA benchmark, which focuses on domain-specific knowledge, KORMo showed notable strength in \textbf{clinical and practical} tasks. This suggests that the model effectively internalized medical and commonsense data presented in real-world QA formats during training. On the other hand, in the KMMLU suite, which focuses on academic and advanced knowledge-based tasks, KORMo showed weaker performance than large multilingual models. This indicates a pretraining bias toward general reasoning rather than fine-grained domain-specific knowledge. The result also implies a lack of exposure to high-quality expert data or explanatory QA formats, likely due to limitations in generating advanced professional content within the augmented Korean datasets.

\paragraph{Cross-lingual Patterns and Efficiency}
% 언어 간 평균 점수 차는 약 6 점(KORMo 기준)으로, 다국어 초대형 모델(Qwen, Gemma)과 유사한 수준의 균형성을 보인다. 한국어 친화 모델(KANANA)은 이 격차가 1 점 내외로 매우 작지만, 영어 성능은 다소 제한적이었다. 반면 KORMo는 \textbf{양언어 모두 고른 범용성}을 확보했다는 점에서, 언어적 효율성과 범용 reasoning의 균형적 트레이드오프가 가장 안정적으로 형성된 사례라 할 수 있다. 이는 제한된 토큰 내에서 bilingual 비율을 세밀하게 조정한 데이터 설계의 효과로 해석된다.

The average score gap between English and Korean for KORMo was around 6 points, demonstrating a level of balance comparable to large-scale multilingual models such as Qwen and Gemma. While KANANA, which is optimized for Korean, showed a much smaller gap of about 1 point, its English performance was relatively limited. In contrast, KORMo demonstrated \textbf{stable performance across both languages}, achieving a well-balanced trade-off between linguistic coverage and general reasoning ability. This outcome underscores the effectiveness of our data design, enabled by fine-grained control over bilingual proportions within a limited token budget.

% \paragraph{Insights and Future Directions}
% 요약하자면, KORMo는 (1) 영어 일반 reasoning에서의 강건함, (2) 한국어 일반 reasoning 및 실용 QA에서의 높은 적응력, (3) 학습 효율 측면에서의 뛰어난 토큰 대비 성능으로 요약된다. 반면, (a) 고난도 전문 지식 문제(MMLU-Pro, KMMLU 계열)와 (b) 어휘적 의미 변별 태스크(KOBALT)에서의 상대적 약세는 향후 보완 과제로 남는다. 이를 개선하기 위해선 (i) 고품질 전공지식 기반 mid-training, (ii) 한국어 중심 contrastive fine-tuning, (iii) 다양한 포맷(explanatory QA, chain-of-thought 등)을 포함한 post-training 전략이 효과적일 것으로 보인다.
% 결국 KORMo의 실험 결과는 “토큰 총량보다 데이터 품질과 언어적 균형이 LLM의 범용성과 효율성을 결정한다”는 점을 실증적으로 보여준다.

\paragraph{Insights and Future Directions}
In summary, KORMo demonstrates (1) robustness on English general reasoning, (2) strong adaptability on Korean general reasoning and practical QA, and (3) superior token efficiency. In contrast, we observe relative weaknesses on (a) high-difficulty specialist knowledge benchmarks (e.g., MMLU-Pro, the KMMLU family) and (b) lexical semantic discrimination tasks (e.g., KOBALT), which remain targets for improvement. To address these gaps, we plan to explore (i) high-quality, domain-knowledge–oriented mid-training, (ii) Korean-centric contrastive fine-tuning, and (iii) post-training strategies that incorporate diverse supervision formats (e.g., explanatory QA and chain-of-thought). Overall, our results provide empirical evidence that, for LLMs, data quality and linguistic balance matter more than sheer token volume in determining generality and efficiency.

%\subsection{SFT 모델 평가}
\subsection{Evaluation of SFT Models}

\begin{table}[t]
\centering
\begin{adjustbox}{max width=\textwidth}
\begin{tabular}{l|rrrr|rrrrr}
\toprule
&\multicolumn{4}{c}{\quad \textbf{\texttt{Fully-Open Model}}} & \multicolumn{5}{c}{\textbf{\texttt{Open-Weight Model (Multilingual)}}} \\
\midrule
\textbf{Benchmark} & kormo-10b  & smolLM3-3b & olmo2-7b & olmo2-13b & kanana1.5-8b & qwen3-8b & llama3.1-8b & exaone3.5-8b* & gemma3-12b \\
\textbf{Tokens} & (2.9T)  & (10.7T) & (4T) & (5.5T) & (3.2T) & (36T) & (15T) & (12T) & (12T) \\
\midrule
%\rowcolor{gray!10}
%\multicolumn{10}{l}{\textbf{English Benchmarks}} \\
%\midrule
%MMLU            & - & - & - & - & - & - & - & - & - \\
MT-Bench        & 8.32 & 7.15 & 7.32 & 7.64 & 8.45 & 8.70 & 6.32 & 8.15 & 8.70 \\
%IFEval          & - & - & - & - & - & - & - & - & - \\
%\midrule
%\textbf{English Avg.}
%& \textbf{-} & \textbf{-} & \textbf{-} & \textbf{-} & \textbf{-} & \textbf{-} & \textbf{-} & \textbf{-} & \textbf{-} \\
%\midrule
%\rowcolor{gray!10}
%\multicolumn{10}{l}{\textbf{Korean Benchmarks}} \\
%\midrule
%KMMLU              & - & - & - & - & - & - & - & - & - \\
KO-MT-Bench        & 8.54 &  - & - & - & 8.02 & 8.16 & 4.27 & 8.13 & 8.51 \\
Logickor           & 8.96 & - & - & - & 8.94 & 8.63 & 6.45 & 9.20 & 8.46 \\
\midrule
\textbf{Average}
& \textbf{8.61} & - & - & - & \textbf{8.47} & \textbf{8.50} & \textbf{5.68} & \textbf{8.49} & \textbf{8.56} \\
\bottomrule
\end{tabular}
\end{adjustbox}
% \caption{모델별 벤치마크 성능 (세로 = Benchmark, 가로 = Model). 언어별 평균은 각 영역의 단순 평균.}
\caption{Benchmark performance (MT-Bench, KO-MT-Bench, Logickor) of Fully-Open and Open-Weight multilingual language models. Scores are on a 10-point scale, with averages computed as simple arithmetic means. Numbers in parentheses indicate training tokens (in trillions). All evaluations were automatically scored using GPT-4o. *Since Exaone4 is available in 1B and 32B sizes, we conducted the comparison using the 8B model.}

\label{tab:vertical_benchmarks}
\end{table}

%KORMo는 비교 하고있는 모델들보다 가장 많은 양의 한국어 Instruction 데이터를 학습을 것으로 추측된다. 이는 KORMo가 MMLU 스타일의 멀티초이스 문제보다 실제 사람들이 느끼는 체감성능이 더 중요하다고 판단해서 다양한 한국어 instruction에 적절히 대답할 수 있는 능력을 중요시 여겼기 때문이다.
We believe KORMo was trained on a larger volume of Korean instruction data than any of the models it is compared against. This was a deliberate design decision, as we focused on the model's capability to respond suitably to various Korean instructions instead of performing well on MMLU-style multiple-choice questions. Our focus was on enhancing the practical, real-world performance as perceived by users.

%\paragraph{평가 설정 및 전제}
\paragraph{Experimental Setup and Preliminary}
%본 실험은 \textbf{Instruction tuning 단계까지만 거친 모델들}을 대상으로 진행되었다. 즉, 강화학습(RLHF, GRPO, APO 등)을 통해 추가적인 추론 강화가 이루어지지 않은 상태에서의 언어·지시 수행 능력을 비교한다. 이 설정은 모델이 지도 데이터(instruction corpus)만으로 얼마나 자연스럽고 논리적인 응답을 생성할 수 있는지를 평가하기 위한 것이다. 세 벤치마크(MT-Bench, KO-MT-Bench, Logickor)는 각각 일반 대화·한국어 대화·논리 추론 중심의 과제를 포함하고 있으며, 모델의 언어적 풍부함, task-following 정확도, 논리 일관성 등을 종합적으로 반영한다.
The present experiment was conducted with models limited to those that have completed \textbf{only the instruction tuning stage}. This means we are comparing their linguistic and instruction execution capabilities prior to any further reasoning enhancement via reinforcement learning (such as RLHF, GRPO, or APO). Such a configuration is intended to assess a model's capacity for generating natural and coherent responses based solely on supervised data. The selected benchmarks (MT-Bench, KO-MT-Bench, and Logickor) encompass tasks focused on general dialogue, Korean-specific dialogue, and logical inference, respectively, and collectively reflect the models' linguistic fluency, fidelity to instructions, and logical coherence.

%\paragraph{Instruction Following 능력(MT-Bench).}
%MT-Bench 및 LogicKor는 영어 및 한국어 기반의 다중 도메인 instruction 수행 벤치마크로, instruction-tuned 모델의 응답 품질과 일관성을 종합적으로 평가한다. 평균 점수 비교 결과, KORMo-10B는 모델 중 가장 높은 평균(8.61)을 기록하며, Gemma3-12B(8.56), Qwen3-8B(8.50) 등 상용급 모델과 거의 동등한 수준을 보였다. 이는 KORMo가 상대적으로 적은 학습 토큰(2.9T)과 자원으로도 instruction corpus의 표현력과 구조적 다양성을 잘 학습했음을 의미한다. 
\paragraph{Instruction Following Ability (MT-Bench).}
MT-Bench and LogicKor serve as multi-domain benchmarks for evaluating instruction-following capabilities in English and Korean, designed to offer a holistic assessment of response quality and consistency in instruction-tuned models. The results of our comparative analysis of mean scores reveal that KORMo-10B achieved the highest average score (8.61). This places its performance in close equivalence to commercial-grade models like Gemma3-12B (8.56) and Qwen3-8B (8.50). Such a result suggests that KORMo successfully captured the expressiveness and structural variety of the instruction corpus despite being trained with comparatively fewer tokens (2.9T) and resources.

\paragraph{The Effect of Language Distribution in Instruction Tuning Data on Model Efficiency.}
%영어(MT-Bench)와 한국어(KO-MT-Bench, Logickor)의 평균을 비교하면, KORMo는 \textbf{영어 8.32 vs 한국어 평균 8.75}로 한국어 대응력이 오히려 더 강하다. 이는 KORMo가 한국어 중심의 instruction fine-tuning 데이터를 충분히 반영하고 있음을 의미하며 제안한 모델들 보다 더 많은 양의 한국어 Instruction 데이터를 학습한 이유로 판단된다.
When comparing the mean scores for English (MT-Bench) against Korean benchmarks (KO-MT-Bench, Logickor), KORMo demonstrates superior capability in Korean, achieving an average score of 8.75, in contrast to 8.32 for English. This finding indicates that KORMo has thoroughly incorporated its Korean-focused instruction fine-tuning dataset. We infer that this is because KORMo was trained on a more substantial amount of Korean instruction data than the other models under consideration.

\paragraph{Conclusion.}
%첫째, KORMo는 instruction-only 설정에서도 대형 다국어 모델과 동등한 성능을 보였으며, 이는 \textbf{데이터 질과 instruction 다양성의 효율적 활용}으로 설명할 수 있다. 둘째, KO-MT-Bench와 Logickor의 결과는 KORMo가 \textbf{한국어 대화적 문맥 및 논리 추론}에서 독보적인 안정성을 보인다는 점을 보여준다. 셋째, 강화학습 없이도 8.6 수준의 MT-Bench 점수를 달성했다는 점은, instruction tuning 만으로도 사용자 지시 이해 및 논리 응답 생성이 충분히 가능함을 시사한다. 마지막으로, KORMo의 강점은 단순한 다국어 대응이 아니라, \textbf{고품질 bilingual instruction fine-tuning을 통한 효율적 지식 정렬(alignment)}에 있다. 향후 강화학습 단계에서는 이 기반 위에서 reasoning chain 확장 및 multi-hop consistency를 강화함으로써, instruction-tuned 성능을 reasoning-capable 모델로 확장할 예정이다.
First, KORMo achieved performance similar to large multilingual models in an instruction-only setup, which we credit to effectively using data quality and instructional diversity. Second, the KO-MT-Bench and Logickor benchmarks show that KORMo demonstrates greater robustness in Korean conversational contexts and logical inference tasks. Third, achieving an 8.6 MT-Bench score without reinforcement learning suggests that instruction tuning alone is enough for understanding user instructions and generating clear, logical responses. Finally, the main advantage of KORMo is not just its multilingual ability, but its effective knowledge alignment through high-quality bilingual instruction fine-tuning. For the subsequent reinforcement learning stage, our objective is to expand upon this base by extending reasoning chains and reinforcing multi-hop consistency, thus transitioning the model from its instruction-tuned capabilities to those of a reasoning-capable agent.

%\subsection{강화학습 모델 평가}
\subsection{Evaluation of Reinforcement Learning Models}
%우리는 최종적으로 앞서 소개한 자체제작한 APO와 GRPO 기반의 데이터를 활용해 수학, 논리추론 데이터를 활용해 모델의 추론 능력을 강화할 예정이다.
As a final step, we will further strengthen the model's reasoning capabilities by incorporating mathematics and logical reasoning data generated using our previously introduced, custom-developed APO and GRPO frameworks.
\section{Conclusion}
This paper introduced \textbf{KORMo}, the first fully open bilingual Korean-English LLM developed primarily with synthetic data. Through extensive quantitative analysis, we demonstrated that synthetic corpora can effectively replace large-scale human-curated data when designed with careful linguistic balance and stylistic diversity. Across both pretraining and instruction-tuning stages, KORMo exhibits stable convergence, strong multilingual generalization, and high reasoning consistency-comparable to leading open-weight models considering the small number of trained tokens. 

Empirical evaluation across over various benchmarks revealed several key insights. First, KORMo maintains competitive English reasoning ability while achieving superior Korean instruction-following and logical coherence. Second, task-specific results (e.g., KR-Clinical-QA, Logickor) highlight the complementary strengths of synthetic data: efficiency in generalized reasoning, yet room for improvement in fine-grained expert knowledge. Third, the language mixture and tokenizer analyses confirm that balanced bilingual compression improves token efficiency and stability, providing a replicable guideline for future FOM builders.

Importantly, KORMo validates that \textbf{synthetic data is not only viable but also scalable} as a principal resource for non-English FOMs. This finding challenges the prevailing assumption that synthetic augmentation leads to model collapse or cultural drift. Instead, with transparent preprocessing, curriculum control, and open evaluation, synthetic data can serve as a sustainable foundation for reproducible multilingual research.

In conclusion, KORMo bridges the gap between open-weight multilingual models and fully open, reproducible pipelines. By publicly releasing the entire data and training process, it enables the community to extend and audit every component of the LLM lifecycle. Future work will expand upon this foundation by (i) incorporating reasoning-oriented reinforcement learning, (ii) exploring multilingual generalization beyond Korean–English, and (iii) establishing standardized evaluation suites for synthetic-data–driven language modeling. Through this effort, KORMo aims to advance the broader movement toward \emph{transparent, reproducible, and inclusive foundation model research}.

\section*{Acknowledgments}
%\paragraph{Special Thanks.} 모델 초기 설계 경험을 공유해주신 Trillion Labs의 신재민, 한승준, 석주영, 최승택, 김규석 님께 진심으로 감사드립니다. 또한 모델 스터디에 참여하여 유익한 아이디어를 제공해주신 KAIST NLPCL 연구실의 황태호, 한승윤, 이준명, 고창건 연구원분들께 감사의 뜻을 전합니다. 아울러 본 연구를 위해 한국어 데이터를 기부해주신 LG U+의 김기현 님께 감사드립니다. 마지막으로 프로젝트의 원활한 수행을 위해 지속적으로 협력해주신 AWS 및 KETI 담당자분들의 노고에 깊이 감사드립니다.

\paragraph{Special Thanks.} We sincerely thank \textit{한승준, 석주영, 최승택, 김규석,} and \textit{신재민} from Trillion Labs for generously sharing their early experience in model design. We are also grateful to \textit{한승윤, 이준명, 고창건, 황의준} and \textit{황태호} from the KAIST NLPCL Lab and \textit{송서현} from Seoultech for their insightful discussions and valuable feedback during the model study phase. We would like to express our gratitude to \textit{김기현} from LG U+ for contributing high-quality Korean data to this project. Finally, we deeply appreciate the support and efforts of the AWS and KETI teams for their assistance in ensuring the smooth execution of this work.

\paragraph{Acknowledgments.} This work was supported by Institute of Information \& communications Technology Planning \& Evaluation (IITP) grant funded by the Korea government(MSIT) (RS-2025-02653113, High-Performance Research AI Computing Infrastructure Support at the 2 PFLOPS Scale)

\bibliography{reference}

\end{document}